\pdfoutput=1

\documentclass[11pt,onecolumn,letterpaper]{article}
\usepackage[in]{fullpage}
\usepackage{natbib}
\usepackage{amsmath}
\usepackage[standard,amsmath,thmmarks]{ntheorem}
\usepackage{amssymb}
\usepackage{amsfonts}
\usepackage{graphicx}
\usepackage{verbatim}
\usepackage{bm}
\usepackage{mathrsfs}
\usepackage[usenames,dvipsnames]{color}
\usepackage{enumerate}

\usepackage{authblk,url}

\usepackage[colorlinks,
            linkcolor=blue,
            citecolor=blue,
            urlcolor=magenta,
            linktocpage,
            plainpages=false]{hyperref}
\usepackage{nameref}

\usepackage{array}

\bibpunct{(}{)}{;}{a}{,}{,}

\usepackage{glossaries}
\DeclareMathOperator*{\argmin}{arg\,min}
\DeclareMathOperator*{\nth}{^{\text{th}}}

\DeclareMathOperator*{\support}{\mathrm{supp}}
\DeclareMathOperator*{\sign}{\mathrm{sign}}
\newcommand{\norm}{\|} 
\newcommand{\E}{\mathsf{E}}
\newcommand{\Prob}{P}

\newcommand{\ProbX}{P_\mathbf{x}}

\newcommand{\ProbZ}{P_\mathbf{z}}
\newcommand{\ProbZp}{P_\mathbf{z'}}

\newcommand{\real}{\mathbb{R}}
\newcommand{\nat}{\mathbb{N}}

\newcommand{\X}{\mathcal{X}}
\newcommand{\Y}{\mathcal{Y}}
\newcommand{\U}{\mathcal{U}}

\newcommand{\F}{\mathcal{F}}

\newcommand{\Dspace}{\mathcal{D}}
\newcommand{\DspaceEps}{\mathcal{D}_\varepsilon}
\newcommand{\Sspace}{\mathcal{S}}
\newcommand{\SspaceEps}{\mathcal{S}_\varepsilon}
\newcommand{\Uspace}{\mathcal{U}}
\newcommand{\Wspace}{\mathcal{W}}
\newcommand{\WspaceEps}{\mathcal{W}_\varepsilon}
\newcommand{\ball}[2]{#1 B_{\real^#2}}
\newcommand{\rademacher}{\mathcal{R}}
\newcommand{\gaussian}{\mathcal{G}}

\newcommand{\goodghost}{\textsc{good\,ghost}}
\newcommand{\badghost}{\textsc{bad\,ghost}}

\newcommand{\lasso}{\textsc{lasso}}
\newcommand{\loss}{l}
\newcommand{\fhatz}{\hat{f}_{\mathbf{z}}}
\newcommand{\Dhatz}{\hat{D}_{\mathbf{z}}}
\newcommand{\whatz}{\hat{w}_{\mathbf{z}}}
\newcommand{\lossof}[1]{\loss(\cdot, #1)}

\newcommand{\smargin}{\mathrm{margin}_s}
\newcommand{\booland}{\,\,\mathbf{and}\,\,}

\makeatletter
\newcommand*{\margingt}[1]{\margingt@#1\@nil}
\newcommand*{\margingt@}{}
\protected\def\margingt@#1,#2,#3\@nil{\ensuremath{ \bigl[ \mathrm{margin}(#1,#2) > #3 \bigr] }}
\makeatother

\makeatletter
\newcommand*{\smargingt}[1]{\smargingt@#1\@nil}
\newcommand*{\smargingt@}{}
\protected\def\smargingt@#1,#2,#3\@nil{\ensuremath{ \bigl[ \mathrm{margin}_s(#1,#2) > #3 \bigr] }}
\makeatother

\makeatletter
\newcommand*{\sparsephi}[1]{\sparsephi@#1\@nil}
\newcommand*{\sparsephi@}{}
\protected\def\sparsephi@#1,#2,#3\@nil{\ensuremath{ #1\text{-}\mathrm{sparse}(\varphi_{#2}(#3)) }}
\makeatother

\newglossaryentry{ball}
{
  name=$\ball{\alpha}{d}$,
  description={The ball in $\real^d$ of radius $\alpha$}
}
\newglossaryentry{Dspace}
{
  name=$\Dspace$,
  description={The space of dictionaries $\left(\ball{}{d}\right)^k$}
}
\newglossaryentry{Wspace}
{
  name=$\Wspace$,
  description={The space of linear hypotheses, equal to $\ball{r}{d}$}
}
\newglossaryentry{Dj}
{
  name=$D_j$,
  description={$j\nth$ atom (column) of dictionary (matrix) $D$}
}
\newglossaryentry{Y}
{
  name=$\Y$,
  description={Space of labels or targets}
}
\newglossaryentry{marginal}
{
  name=$\Pi$,
  description={Marginal probability measure over input space $\ball{}{d}$}
}
\newglossaryentry{P}
{
  name=$\Prob$,
  description={Joint probability measure over $\ball{}{d} \times \Y$}
}
\newglossaryentry{Pf}
{
  name=$\Prob f$,
  description={$\E_{(x,y) \sim \Prob} f(x)$}
}
\newglossaryentry{Plossf}
{
  name=$\Prob \lossof{f}$,
  description={$\E_{(x,y)} \loss(y(f(x))$}
}
\newglossaryentry{Pzf}
{
  name=$\ProbZ f$,
  description={$\sum_{i=1}^m f(x_i)$}
}
\newglossaryentry{Pzlossf}
{
  name=$\ProbZ \lossof{f}$,
  description={$\sum_{i=1}^m \loss(y_i,f(x_i))$}
}
\newglossaryentry{fhatz}
{
  name=$\fhatz$,
  description={Hypothesis returned by learner from $\mathbf{z}$}
}
\newglossaryentry{indexset}
{
  name=$[n]$,
  description={$\{1, 2, \ldots, n\}$}
}
\newglossaryentry{support}
{
  name=$\support(t)$,
  description={$\{i \in [k] : t_i \neq 0\}$}
}
\newglossaryentry{singular}
{
  name=$\varsigma_s(A)$,
  description={the $s\nth$ singular value of $A$}
}
\newglossaryentry{incoherence}
{
  name=$\mu_s(D)$,
  description={$s$-incoherence: minimum $(\varsigma_s(D))^2$ among $s$-atom subdictionaries of $D$}
}
\newglossaryentry{smargin}
{
  name={$\smargin(D,x)$},
  description={$\max_{\substack{\mathcal{I} \subseteq [k] \\ |\mathcal{I}| = k - s}} \min_{j \in \mathcal{I}} \Bigl\{ \lambda - \bigl| \langle D_j, x_i - D \varphi_D(x_i) \rangle \bigr| \Bigr\}$}
}
\newglossaryentry{smarginsample}
{
  name={$\smargin(D,\mathbf{x})$},
  description={$\min_{x_i \in \mathbf{x}} \smargin(D, x_i)$}
}
\newglossaryentry{sparse}
{
  name={$s\text{-}\mathrm{sparse}(\varphi_{D}(\mathbf{x}))$},
  description={for all $x_i \in \mathbf{x}$, $\norm \varphi_D(x_i) \norm_0 \leq s$}
}
\newglossaryentry{z}
{
  name=$\mathbf{z}$,
  description={labeled $m$-sample of training data}
}
\newglossaryentry{zp}
{
  name=$\mathbf{z'}$,
  description={second labeled $m$-sample (ghost sample)}
}
\newglossaryentry{xpp}
{
  name=$\mathbf{x''}$,
  description={unlabeled $m$-sample}
}
\newglossaryentry{Dspacemu}
{
  name=$\Dspace_\mu$,
  description={$\{D \in \Dspace: \mu_s(D) \geq \mu\}$}
}
\newglossaryentry{F}
{
  name=$\F$,
  description={$\{f_{D,w}:= x \mapsto \langle w, \varphi_D(x)\rangle : D \in \Dspace, w \in \Wspace\}$. 
}
}
\newglossaryentry{Fmu}
{
  name=$\F_\mu$,
  description={$\{f = (D,w) \in \F: D \in \Dspace_\mu\}$}
}
\newglossaryentry{Fmustar}
{
  name=$\F_{\bm{\mu^*}}$,
  description={$\{f = (D,w) \in \F: (\mu_s(D) \geq \mu^*_s) \booland (\mu_{2s}(D) \geq \mu^*_{2s})$}
}
\newglossaryentry{Fmustarx}
{
  name=$\F_{\bm{\mu^*}}(\mathbf{x})$,
  description={$\{ f \in \F_{\bm{\mu^*}}: 
    \sparsephi{s, D, \mathbf{x}} \booland \smargingt{D, \mathbf{x}, \tau} 
    \}$}
}
\newglossaryentry{Fmustaretax}
{
  name=$\F_{\bm{\mu^*},\eta}(\mathbf{x})$,
  description={$\{ f \in \F_{\bm{\mu^*}}: 
    \exists \, \mathbf{\tilde{x}} \subseteq_\eta \mathbf{x} \,\, 
    \sparsephi{s, D, \mathbf{\tilde{x}}} \booland \smargingt{D, \mathbf{\tilde{x}}, \tau} 
    \}$}
}
\newglossaryentry{allbuteta}
{
  name=$\mathbf{\tilde{x}} \subseteq_\eta \mathbf{x}$,
  description={$\mathbf{\tilde{x}}$ is a subset of $\mathbf{x}$ with at most $\eta$ elements of $\mathbf{x}$ removed}
}
\newglossaryentry{rademachervar}
{
  name=$\sigma_i$,
  description={Rademacher random variable}
}
\newglossaryentry{gaussianvar}
{
  name=$\gamma_i$,
  description={standard normal random variable}
}
\newglossaryentry{conditionalrademacher}
{
  name=$\rademacher_{m \mid \mathbf{x}}(\F)$,
  description={conditional Rademacher average of $\F$}
}
\newglossaryentry{conditionalgaussian}
{
  name=$\gaussian_{m \mid \mathbf{x}}(\F)$,
  description={conditional Gaussian average of $\F$}
}
\newglossaryentry{encoder}
{
  name=$\varphi_D$,
  description={$\varphi_D(x) := \argmin_z \norm x - D z \norm_2^2 + \lambda \norm z \norm_1$}
}
\newglossaryentry{srestricted2norm}
{
  name=$\norm S \norm_{2,s}$,
  description={The $s$-restricted 2-norm: $\sup_{\{t \in \real^n: \norm t \norm = 1, |\support(t)| \leq s\}} \norm S t \norm_2$}
}

\makeglossaries

\begin{document}

\title{On the Sample Complexity of Predictive Sparse Coding}

\author{Nishant A. Mehta$^*$}
\author{Alexander G. Gray}
\affil{
  College of Computing \\
  Georgia Institute of Technology \\
  Atlanta, GA 30332, USA
}
\date{}
\maketitle
\thispagestyle{empty}
\let\oldthefootnote\thefootnote
\renewcommand{\thefootnote}{\fnsymbol{footnote}}
\footnotetext[1]{To whom correspondence should be addressed. Email: \protect\url{niche@cc.gatech.edu}}
\let\thefootnote\oldthefootnote

\theoremstyle{plain}
\theoremheaderfont{\normalfont\bfseries}\theorembodyfont{\slshape}
\theoremseparator{}
\renewtheorem{Theorem}[theorem]{Theorem}
\renewtheorem{Lemma}[lemma]{Lemma}
\renewtheorem{Definition}[definition]{Definition}

\theoremsymbol{\rule{1ex}{1ex}}
\renewtheorem{Proposition}[proposition]{Proposition}
\renewtheorem{Corollary}[corollary]{Corollary}

\theoremheaderfont{\it}\theorembodyfont{\upshape}
\theoremstyle{nonumberplain}
\theoremseparator{:}
\theoremsymbol{\rule{1ex}{1ex}}

\renewtheorem{Proof}{Proof}

\newtheorem{Proof-sketch}{Proof sketch}

\begin{abstract}%

The goal of predictive sparse coding is to learn a representation of examples as sparse linear combinations of elements from a dictionary, such that a learned hypothesis linear in the new representation performs well on a predictive task. Predictive sparse coding algorithms recently have demonstrated impressive performance
on a variety of supervised tasks, but their generalization properties have not been studied. 
We establish the first generalization error bounds for predictive sparse coding, covering two settings: 1) the overcomplete setting, where the number of features $k$ exceeds the original dimensionality $d$; and 2) the high or infinite-dimensional setting, where only dimension-free bounds are useful. Both learning bounds intimately depend on stability properties of the learned sparse encoder, as measured on the training sample. Consequently, we first present a fundamental stability result for the LASSO, a result characterizing the stability of the sparse codes with respect to perturbations to the dictionary. In the overcomplete setting, we present an estimation error bound that decays as $\tilde{O}(\sqrt{d k/m})$ with respect to $d$ and $k$. In the high or infinite-dimensional setting, we show a dimension-free bound that is $\tilde{O}(\sqrt{k^2 s / m})$ with respect to $k$ and $s$, where $s$ is an upper bound on the number of non-zeros in the sparse code for any training data point. 
\end{abstract}

\section{Introduction}
Learning architectures such as the support vector machine and other linear predictors enjoy strong theoretical properties \citep{steinwart2008svm,kakade2009complexity}, but a learning-theoretic understanding of many more complex learning architectures is lacking. 
Predictive methods based on \emph{sparse coding} recently have emerged which simultaneously learn a data representation via a nonlinear encoding scheme and an estimator linear in that representation \citep{bradley2009differentiable,mairal2012task,mairal2008supervised}. A sparse coding representation $z \in \real^k$ of a data point $x \in \real^d$ is learned by representing $x$ as a sparse linear combination of $k$ \emph{atoms} \glsadd{Dj} $D_j \in \real^d$ of a \emph{dictionary} $D =(D_1, \ldots, D_k) \in \real^{d \times k}$. In the coding $x \approx \sum_{j=1}^k z_j D_j$, all but a few $z_j$ are zero. 

Predictive sparse coding methods such as \cite{mairal2012task}'s \emph{task-driven dictionary learning} recently have achieved state-of-the-art results on many tasks, including the MNIST digits task. 
Whereas standard sparse coding minimizes an unsupervised, reconstructive $\ell_2$ loss, predictive sparse coding seeks to minimize a supervised loss by optimizing a dictionary and a linear predictor that operates on encodings to that dictionary. 
There is much empirical evidence that sparse coding can provide good abstraction by finding higher-level representations which are useful in predictive tasks \citep{yu2009nonlinear}. 
Intuitively, the power of \emph{prediction-driven} dictionaries is that they pack more atoms in parts of the representational space where the prediction task is more difficult. However, despite the empirical successes of predictive sparse coding methods, it is unknown how well they generalize in a theoretical sense. 

In this work, we develop what to our knowledge are the first generalization error bounds for predictive sparse coding algorithms; in particular, we focus on \emph{$\ell_1$-regularized sparse coding}. 
\cite{maurer2010k}  and \cite{vainsencher2011sample} previously established generalization bounds for the classical, reconstructive sparse coding setting. 
Extending their analysis to the predictive setting introduces certain difficulties related to the richness of the class of sparse encoders. 
Whereas in the reconstructive setting, this complexity can be controlled directly by exploiting the stability of the \emph{reconstruction error} to dictionary perturbations, in the predictive setting it appears that the complexity hinges upon the stability of the \emph{sparse codes themselves} to dictionary perturbations. This latter notion of stability is much harder to prove; moreover, it can be realized only with additional assumptions which depend on the dictionary, the data, and their interaction (see Theorem \ref{thm:new-sparse-coding-stability}). Furthermore, when the assumptions hold for the learned dictionary and data, we also need to guarantee that the assumptions hold on a newly drawn sample.

\paragraph{Contributions}
We provide learning bounds for two core scenarios in predictive sparse coding: the \emph{overcomplete setting} where the dictionary size, or number of learned features, $k$ exceeds the ambient dimension $d$; and the infinite-dimensional setting where only dimension-free bounds are acceptable. Both bounds hold provided the size $m$ of the training sample is large enough, where the critical size for the bounds to kick in depends on a certain notion of stability of the learned representation.
The core contributions of this work are:
\begin{enumerate}
\item Under mild conditions, a stability bound for the LASSO \citep{tibshirani1996regression} under dictionary perturbations. (Theorem \ref{thm:new-sparse-coding-stability})
\item In the overcomplete setting, a learning bound that is essentially of order 
$\sqrt{\frac{d k}{m}} + \frac{\sqrt{s}}{\lambda \mu_s(D)}$, where each sparse code has at most $s$ non-zero coordinates. The term $\frac{1}{\mu_s(D)}$ is the inverse $s$-incoherence (see Definition \ref{def:s-incoherence}) and is roughly the worst condition number among all linear systems induced by taking $s$ columns of $D$. 
(Theorem \ref{thm:final-learning-bound-overcomplete})
\item In the infinite-dimensional setting, a learning bound that is \emph{independent} of the dimension of the data; this bound is essentially of order 
$\frac{1}{\mu_{2s}(D)} \sqrt{\frac{k^2 s}{m}}$. 
(Theorem \ref{thm:final-learning-bound-infinite})
\end{enumerate}

The stability of the sparse codes are absolutely crucial to this work. Proving that the notion of stability of contribution 1 holds is quite difficult because the LASSO is not strongly convex in general. Consequently, much of the technical difficulty of this work is owed to finding conditions under which the LASSO is stable under dictionary perturbations and proving that when these conditions hold with respect to the learned hypothesis and the training sample, they also hold with respect to a future sample.

For convenience, we have collected all of the various notation of this paper in a glossary in Appendix \ref{sec:glossary}.

\subsection{The predictive sparse coding problem}
Let \glsadd{P} $\Prob$ be a probability measure over $\ball{}{d} \times \Y$, the product of an input space \glsadd{ball} $\ball{}{d}$ (the unit ball of $\real^d$) 
and a space \glsadd{Y} $\Y$ of univariate labels; examples of $\Y$ include a bounded subset of $\real$ for regression and $\{-1,1\}$ for classification. 
Let \glsadd{z} $\mathbf{z} = (z_1, \ldots, z_m)$ be a sample of $m$ points drawn iid from $\Prob$, where each labeled point $z_i$ equals $(x_i,y_i)$ for $x_i \in \ball{}{d}$ and $y_i \in \Y$. 
In the reconstructive setting, labels are not of interest and we can just as well consider an unlabeled sample $\mathbf{x}$ of $m$ points drawn iid from the marginal probability measure \glsadd{marginal} $\Pi$ on $\ball{}{d}$. 

The sparse coding problem is to represent each point $x_i$ as a sparse linear combination of $k$ basis vectors, or \emph{atoms} $D_1, \ldots, D_k$. The atoms form the columns of a \emph{dictionary} $D$ living in a space of dictionaries \glsadd{Dspace} $\Dspace := \left(\ball{}{d}\right)^k$, for $D_i = (D_i^1, \ldots, D_i^d)^T$ in the unit $\ell_2$ ball. 
An encoder \glsadd{encoder} $\varphi_D$ can be used to frame $\ell_1$ sparse coding:
\begin{align}
\label{eqn:encoder}
\varphi_D(x) := \argmin_z \|x - D z\|_2^2 + \lambda \|z\|_1;
\end{align}
hence, encoding $x$ as $\varphi_D(x)$ amounts to solving a LASSO problem. 
The reconstructive $\ell_1$ sparse coding objective is then
\begin{align*}
\min_{D \in \Dspace} \E_{x \sim \Pi} \|x - D \varphi_D(x) \|_2^2 + \lambda \|\varphi_D(x)\|_1,
\end{align*}
Generalization bounds for the empirical risk minimization (ERM) variant of this objective have been established. 
In the infinite-dimensional setting, \cite{maurer2010k} showed\footnote{To see this, take Theorem 1.2 of \citet{maurer2010k} with $Y = \{y \in \real^k: \|y\|_1 < \frac{1}{\lambda}\}$ and \mbox{$\mathcal{T} = \{T: \real^k \rightarrow \real^d: \|T e_j\| \leq 1, j \in [k]\}$}, so that $\|\mathcal{T}\|_Y \leq \frac{1}{\lambda}$.} that with probability $1 - \delta$ over the training sample $\mathbf{x}$:
\begin{align}
\label{eqn:unsupervised-maurer-pontil-bound}
\sup_{D \in \Dspace} \Prob f_D - \ProbX f_D
\,\, \leq \,\, 
\frac{k}{\sqrt{m}} \left( \frac{14}{\lambda} + \frac{1}{2} \sqrt{\log \left( 16 m / \lambda^2 \right) } \right) + \sqrt{\frac{\log (1/\delta)}{2 m}}
\end{align}
where $f_D(x) := \min_{z \in \real^k} \|x - D z\|_2^2 + \lambda \|z\|_1$. 
This bound is \emph{independent} of the dimension $d$ and hence useful when $d \gg k$, as in general Hilbert spaces. They also showed a similar bound in the overcomplete setting where the $k$ is replaced by $\sqrt{d k}$. 
\cite{vainsencher2011sample} handled the overcomplete setting, producing a bound that is $O \bigl(\sqrt{d k / m} \bigr)$ as well as fast rates of $O(d k /m$), with only logarithmic dependence on $\frac{1}{\lambda}$. 

\emph{Predictive sparse coding}, introduced by \cite{mairal2012task}, minimizes a supervised loss with respect to a representation and an estimator linear in the representation. 
Let \glsadd{Wspace} $\Wspace$ be a space of linear hypotheses with 
$\Wspace := \ball{r}{k}$, the ball in $\real^k$ scaled to radius $r$. 
A predictive sparse coding hypothesis function $f$ is identified by $f = (D,w) \in \Dspace \times \Wspace$ and defined as $f(x) = \langle w, \varphi_D(x)\rangle$. 
The function class \glsadd{F} $\F$ is the set of such hypotheses. 
The loss will be measured via $\loss: \Y \times \real \rightarrow [0,b]$, $b > 0$, a bounded loss function that is $L$-Lipschitz in its second argument.

The predictive sparse coding objective is\footnote{While the focus of this work is \eqref{eqn:dsc-objective}, formally the predictive sparse coding framework admits swapping out the squared $\ell_2$ norm regularizer on $w$ for any other regularizer.}
\begin{align}
\label{eqn:dsc-objective}
\min_{D \in \Dspace, w \in \Wspace} \E_{(x,y) \sim \Prob} \loss(y, \langle w, \varphi_D(x) \rangle) + \frac{1}{r} \|w\|_2^2 ;
\end{align}
In this work, we analyze the ERM variant of \eqref{eqn:dsc-objective}:
\begin{align}
\label{eqn:dsc-erm-objective}
\min_{D \in \Dspace, w \in \Wspace} \frac{1}{m} \sum_{i=1}^m \loss(y_i, \langle w, \varphi_D(x_i) \rangle) + \frac{1}{r} \|w\|_2^2.
\end{align}
Because this objective is not convex and global optimizers are not known, \emph{a priori} we cannot say whether the optimal hypothesis or a nearly optimal hypothesis will be returned by any learning algorithm. However, we can and will bet on certain sparsity-related stability properties holding with respect to the learned hypothesis and the training sample. Consequently, all the presented learning bounds will hold uniformly not over the set of all hypotheses but rather potentially much smaller random subclasses of hypotheses. Additionally, the presented bounds will be algorithm-independent\footnote{Empirically we have observed that stochastic gradient approaches like the one in \cite{mairal2012task} perform very well.}, although certainly algorithm design can influence the observed stability of the learned hypothesis and hence the best learning bound that applies \emph{a posteriori}.

\paragraph{Encoder stability}
Defining the encoder \eqref{eqn:encoder} via the $\ell_1$ sparsity-inducing regularizer (sparsifier) is just one way of designing an encoder. The choice of sparsifier seems to be pivotal both from an empirical perspective and a theoretical one. \cite{bradley2009differentiable} used a differentiable \emph{approximate} sparsifier based on the Kullback-Leibler divergence (true sparsity may not result).
The $\ell_1$ sparsifier $\|\cdot\|_1$ is the most popular and notably is the tightest convex lower bound for the 
$\ell_0$ ``norm'': $\|x\|_0 := |\{i: x_i \neq 0\}|$ \citep{fazel2002matrix}. 
Regrettably, from a stability perspective the $\ell_1$ sparsifier is not well-behaved in general. Indeed, due to the lack of strict convexity, each $x$ need not have a unique image under $\varphi_D$. It also is unclear how to analyze the class of mappings $\varphi_D$, parameterized by $D$, if the map changes drastically under small perturbations to $D$. Hence, we will begin by establishing sufficient conditions under which $\varphi_D$ is stable under perturbations to $D$.

\section{Conditions and main results} 

In this section, we develop several quantities that are central to the statement of the main results. 
Throughout this paper, let \glsadd{indexset} $[n] := \{1, \ldots, n\}$ for $n \in \nat$. Also, for $t \in \real^k$, define \glsadd{support} $\support(t) := \{i \in [k] : t_i \neq 0\}$. 

\begin{Definition}[$s$-incoherence]
\label{def:s-incoherence}
For $s \in [k]$ and $D \in \Dspace$, the \emph{$s$-incoherence} \glsadd{incoherence} $\mu_s(D)$ is defined as the square of the minimum singular value among $s$-atom subdictionaries of $D$. 
Formally, 
\begin{align*}
\mu_s(D) = \bigl( \min \left\{ \varsigma_s(D_\Lambda) : \Lambda \subseteq [k], |\Lambda| = s \right\} \bigr)^2 ,
\end{align*}
where \glsadd{singular} $\varsigma_s(A)$ is the $s\nth$ singular value of $A$.
\end{Definition}
The $s$-incoherence can used to guarantee that sparse codes are stable in a certain sense. 

We now introduce some key parameter-and-data-dependent properties. 
The first property regards the sparsity of the encoder on a sample $\mathbf{x} = (x_1,\ldots, x_m)$. 
\begin{Definition}[$s$-sparsity]
If every point $x_i$ in the set of points $\mathbf{x}$ satisfies $\|\varphi_D(x_i)\|_0 \leq s$, then \emph{$\varphi_D$ is $s$-sparse on $\mathbf{x}$}. More concisely, the boolean expression \glsadd{sparse} $\sparsephi{s, D, \mathbf{x}}$ is true. 
\end{Definition}
This property is critical as the learning bounds will exploit the observed sparsity level over the training sample. 
The following collection of properties also will be useful.
\begin{Definition}[$s$-margin]
Given a dictionary $D$ and a point $x_i \in \ball{}{d}$, the \emph{$s$-margin of $D$ on $x_i$} is \glsadd{smargin}
\begin{align*}
\smargin(D, x_i) := 
\max_{\substack{\mathcal{I} \subseteq [k] \\ |\mathcal{I}| = k - s}} \min_{j \in \mathcal{I}}
\Bigl\{ \lambda - \bigl| \langle D_j, x_i - D \varphi_D(x_i) \rangle \bigr| \Bigr\} .
\end{align*}
The sample version of the $s$-margin is the maximum $s$-margin that holds for all points in $\mathbf{x}$, or the \emph{$s$-margin of $D$ on $\mathbf{x}$}: \glsadd{smarginsample}
\begin{align*}
\smargin(D, \mathbf{x}) &:= \min_{x_i \in \mathbf{x}} \smargin(D, x_i) .
\end{align*}
\end{Definition}

The importance of these $s$-margin properties flows directly from the upcoming Sparse Coding Stability Theorem (Theorem \ref{thm:new-sparse-coding-stability}). 
Intuitively, if the $s$-margin of $D$ on $x$ is high, then there is a set of $(k-s)$ inactive atoms that are poorly correlated with the optimal residual $x - D \varphi_D(x)$, and hence these atoms are far from being included in the set of active atoms. 
More formally, 
$\smargin(D, x_i)$ is equal to the $(s+1)\nth$ smallest element of the set of $k$ elements $\left\{ \lambda - \left| \langle D_j, x_i - D \varphi_D(x_i) \rangle \right| \right\}_{j \in [k]}$. 
Note that if $\|\varphi_D(x_i)\|_0 = s$, we can use the $(s + \rho)$-margin for any integer $\rho \geq 0$. Indeed, $\rho > 0$ is justified when $\varphi_D(x_i)$ has only $s$ non-zero dimensions but for precisely one index $j^*$ outside the support set $| \langle D_{j^*}, x_i - D \varphi_D(x_i) \rangle |$ is arbitrarily close to $\lambda$. In this scenario, the $s$-margin of $D$ on $x_i$ is trivially small; however, the $(s+1)$-margin is non-trivial because the $\max$ in the definition of the margin will remove $j^*$ from the $\min$'s choices $\mathcal{I}$. Empirical evidence shown in Section \ref{sec:empirical} suggests that even when $\rho$ is small, the ($s + \rho)$-margin is not too small.

\paragraph{Sparse coding stability}
The first result of this work is a fundamental stability result for the LASSO. In addition to being critical in motivating the presented conditions, the result may be of interest in its own right.

\begin{Theorem}[Sparse Coding Stability]
\label{thm:new-sparse-coding-stability}
Let dictionaries $D, \tilde{D} \in \Dspace$ satisfy \mbox{$\mu_s(D), \mu_s(\tilde{D}) \geq \mu$} and \mbox{$\|D - \tilde{D}\|_2 \leq \varepsilon$} for some $\mu > 0$, and let $x \in \ball{}{d}$. Suppose that there exists an index set $\mathcal{I} \subseteq [k]$ of $k - s$ indices such that for all $i \in \mathcal{I}$:
\begin{align}
\left| \langle D_i, x - D \varphi_D(x) \rangle \right| < \lambda - \tau 
\label{eqn:residual-corr}
\end{align}
\begin{flalign}
\text{for} && \varepsilon \leq \frac{\tau^2 \lambda}{43} \,\,\,\, . &&
\label{eqn:prp}
\end{flalign}
Then the following stability bound holds:
\begin{align*}
\| \varphi_D(x) - \varphi_{\tilde{D}}(x) \|_2 
\leq \frac{3 \varepsilon \sqrt{s}}{\lambda \mu} \,\,\,\, .
\end{align*}
Furthermore, if $\varepsilon = \frac{{\tau'}^2 \lambda}{43}$ for $\tau' < \tau$, then for all $i \in \mathcal{I}$:
\begin{align*}
\left| \langle \tilde{D}_i, x - \tilde{D} \varphi_{\tilde{D}}(x) \rangle \right| \leq \lambda - (\tau - \tau') \,\, .
\end{align*}
Thus, some margin, and hence sparsity, is retained after perturbation.
\end{Theorem}

Condition \eqref{eqn:residual-corr} means that at least $k - s$ inactive atoms in the coding $\varphi_D(x)$ do not have too high absolute correlation with the residual $x - D \varphi_D(x)$. 
We refer to the right-hand side of \eqref{eqn:prp} as the permissible radius of perturbation (PRP) because it indicates the maximum amount of perturbation for which the theorem can guarantee encoder stability. 
In short, the theorem says that if problem \eqref{eqn:encoder} admits a stable sparse solution, then a small perturbation to the dictionary will not change the fact that a certain set of $k - s$ atoms remains inactive in the new solution. The theorem further states that the perturbation to the solution will be bounded by a constant factor times the size of the perturbation, where the constant depends on the $s$-incoherence, the amount of $\ell_1$-regularization, and the sparsity level.

The proof of Theorem \ref{thm:new-sparse-coding-stability} is quite long, and so we leave all but the following high-level sketch to Appendix \ref{sec:proof-sparse-coding-stability-theorem}.
\begin{Proof-sketch} 
First, we show that the solution $\varphi_{\tilde{D}}(x)$ is $s$-sparse and, in particular, has support contained in the complement of $\mathcal{I}$. 
Second, we reframe the LASSO as a quadratic program (QP). By exploiting the convexity of the QP and the fact that both solutions have their support contained in a set of $s$ atoms, simple linear algebra yields the desired stability bound. In our view, the first step is much more difficult than the second. Our strategy for the first step has four planks:
\begin{enumerate}[(1)]
\item \textsc{optimal value stability}: 
 The two problems' optimal objective values are close; this is an easy consequence of the closeness of $D$ and $\tilde{D}$. 
\item \textsc{stability of norm of reconstructor}: 
  The \emph{norms} of the optimal reconstructors ($D \varphi_D(x)$ and $\tilde{D} \varphi_{\tilde{D}}(x)$) of the two problems are close. We show this using \textsc{optimal value stability} and
\begin{align}
\label{eqn:osborne-equality}
(x - D \varphi_D(x))^T D \varphi_D(x) = \lambda \|\varphi_D(x)\|_1,
\end{align}
the latter of which can be shown via convex duality \cite{osborne2000lasso}. 
\item \textsc{reconstructor stability}: 
  The optimal reconstructors of the two problems are close. This fact can be shown to be a consequence of \textsc{stability of norm of reconstructor}, using the $\ell_1$ norm's convexity and the equality \eqref{eqn:osborne-equality}. 
\item \textsc{preservation of sparsity}: 
  The solution to the perturbed problem also is supported on the complement of $\mathcal{I}$. To show this, it is sufficient to show that the absolute correlation of each atom $\tilde{D}_i$ ($i \in \mathcal{I}$) with the residual in the perturbed problem is less than $\lambda$. This last claim is a relatively easy consequence of \textsc{reconstructor stability}. 
\end{enumerate}
\end{Proof-sketch}

Although we do not make use of it in this work, under considerably stronger conditions, we can achieve a similar stability bound with a far smaller PRP. 
\begin{Theorem}[Restricted Stability\footnotemark] 
\label{thm:old-sparse-coding-stability}
Let dictionaries $D, \tilde{D} \in \Dspace$ satisfy \mbox{$\mu_s(D), \mu_s(\tilde{D}) \geq \mu$} and \mbox{$\|D - \tilde{D}\|_2 \leq \varepsilon$} for some $\mu > 0$, and let $x \in \ball{}{d}$. 
Suppose that there exist $\tau > 0$ and $s \in [k]$ satisfying:
\begin{align}
&(i) &&\|\varphi_D(x)\|_0 \leq s, \\
&(ii) &&\left| (\varphi_D(x))_j \right| > \tau \text{ for all } j \in \support{\varphi_D(x)}, \label{eqn:bounded-away-from-zero} \\
&(iii)  &&\left| \langle D_j, x - D \varphi_D(x) \rangle \right| < \lambda - \tau \text{ for all } j \notin \support{\varphi_D(x)}; \label{eqn:old-residual-corr}
\end{align}
\begin{flalign}
\text{for} && \varepsilon 
\leq 
\frac{\tau \mu}{\frac{s + \mu}{\lambda} + \sqrt{s} + \mu} \,\,\,\, . &&
\label{eqn:old-prp}
\end{flalign}
Then the supports of the optimal solutions are identical:
\begin{align*}
  \support(\varphi_D(x)) = \support(\varphi_{\tilde{D}}(x)) \,\, ,
\end{align*}
and the following stability bound holds:
 \begin{flalign*}
 \| \varphi_D(x) - \varphi_{\tilde{D}}(x) \|_2 
 \leq \frac{\varepsilon}{\mu} \left( \frac{\sqrt{s}}{\lambda} + 1 \right) \,\, .
 \end{flalign*}
\end{Theorem}
\footnotetext{In the previous version of this paper \citep{mehta2012sample}, the Theorem 1 of the current paper did not exist and the Theorem 2 of this paper was labeled as Theorem 1.}

Theorem \ref{thm:old-sparse-coding-stability} applies provided that every non-zero coefficient of $\varphi_D(x)$ has magnitude bounded above zero and every unused atom is far from being brought into the optimal solution in the sense of \eqref{eqn:old-residual-corr} (note this inequality is identical to \eqref{eqn:residual-corr}. The conditions of Theorem \ref{thm:old-sparse-coding-stability} are more demanding than those of Theorem \ref{thm:new-sparse-coding-stability} because in the former, there is exactly one choice of the inactive set $\mathcal{I}$, namely, $[k] \setminus \support{\varphi_D(x)}$. If this choice does not yield sufficient margin, we are out of luck. Nevertheless, when the conditions of the Theorem \ref{thm:old-sparse-coding-stability} \emph{do} hold, 
the PRP's dependence on the margin-like property $\tau$ is considerably reduced from quadratic in Theorem \ref{thm:new-sparse-coding-stability} to only linear in Theorem \ref{thm:old-sparse-coding-stability}. 
Although the PRP now depends on $s$ and $\mu$, both of these properties often are well-behaved (a small $s$ is desired and typically leads to a large $\mu$), whereas $\tau$ is a wild-card on which minimum dependence is desired. 

\begin{Proof-sketch} 
Our strategy is to show that there is a unique solution to the perturbed problem, defined in terms of the optimality conditions of the LASSO (see conditions L1 and L2 of \citep{asif2010lasso}), and this solution has the same support as the solution to the original problem. As a result, the perturbed solution's proximity to the original solution is governed in part by a condition number $\mu$ of a linear system of $s$ variables. 
\end{Proof-sketch}

\subsection{Main results}
The following notation will aid and abet the below results and the subsequent analysis. 
Recall that the loss $\loss$ is bounded by $b$ and $L$-Lipschitz in its second argument. 
Also recall that $\F$ is the set of predictive sparse coding hypothesis functions $f(x) = \langle w, \varphi_D(x)\rangle$ indexed by $D \in \Dspace$ and $w \in \Wspace$. 
For $f \in \F$, define $\lossof{f}: \Y \times \real^d \rightarrow [0,b]$ as the loss-composed function $(y,x) \mapsto \loss(y,f(x))$. Let $\loss \circ \F$ be the class of such functions induced by the choice of $\F$ and $\loss$. 
A probability measure $\Prob$ operates on functions and loss-composed functions as: \glsadd{Pf} \glsadd{Plossf}
\begin{align*}
\Prob f = \E_{(x,y) \sim \Prob} f(x) &&
\Prob \lossof{f} = \E_{(x,y) \sim \Prob} \loss(y,f(x)) .
\end{align*}
Similarly, an empirical measure $\ProbZ$ associated with sample $\mathbf{z}$ operates on functions and loss-composed functions as: \glsadd{Pzf} \glsadd{Pzlossf}
\begin{align*}
\ProbZ f = \frac{1}{m} \sum_{i=1}^m f(x_i) && 
\ProbZ \lossof{f} = \frac{1}{m} \sum_{i=1}^m \loss(y_i,f(x_i)) .
\end{align*}

Finally, when provided a training sample $\mathbf{z}$, the hypothesis returned by the learner will be referred to as \glsadd{fhatz} $\fhatz$. 
Note that $\fhatz$ is random, but $\fhatz$ becomes a fixed function upon conditioning on $\mathbf{z}$. 

Classically speaking, the overcomplete setting is the \emph{modus operandi} in sparse coding. In this setting, an overcomplete basis is learned which will be used parsimoniously in coding individual points. 
The next result bounds the generalization error in the overcomplete setting. 
The $\tilde{O}(\cdot)$ notation hides $\log(\log(\cdot))$ terms and assumes that $r \leq m^{\min\{d,k\}}$. 
\begin{Theorem}[Overcomplete Learning Bound]
\label{thm:final-learning-bound-overcomplete}
With probability at least $1 - \delta$ over \\ $\mathbf{z} \sim \Prob^m$, 
for any $s \in [k]$ and any $f = (D,w) \in \F$ satisfying 
$\sparsephi{s, D, \mathbf{x}}$ 
and 
\begin{align*}
m > \frac{387}{\smargin(D,\mathbf{x})^2 \lambda} ,
\end{align*}
the generalization error $(\Prob - \ProbZ) \lossof{f}$ is 
\begin{align}
\tilde{O} \left( 
b \sqrt{\frac{d k \log m + \log \frac{1}{\delta}}{m}} 
+ \frac{b}{m} \left( d k \log \frac{1}{\smargin^2(D,\mathbf{x}) \cdot \lambda} \right) 
+ \frac{L}{m} \left( \frac{r \sqrt{s}}{\lambda \mu_s(D)} \right) 
\right) .
\label{eqn:essential-overcomplete}
\end{align}
\end{Theorem}
Note that this bound also applies to the particular hypothesis $\fhatz = (\Dhatz, \whatz)$ learned from the training sample.

Often in learning problems, we first map the data implicitly to a space of very high dimension or even infinite dimension and use kernels for efficient computations. In these cases where $d \gg k$ or $d$ is infinite, it is unacceptable for any learning bound to exhibit dependence on $d$. It is possible to untether the analysis from $d$ by using the $s$-margin of the learned dictionary $\Dhatz$ on a second, unlabeled sample. In the infinite-dimensional setting, the following dimension-free learning bound holds. 

\begin{Theorem}[Infinite-Dimensional Learning Bound]
\label{thm:final-learning-bound-infinite}
With probability at least $1 - \delta$ over a labeled $m$-sample $\mathbf{z} \sim \Prob^m$ and a second, unlabeled sample \glsadd{xpp} $\mathbf{x''} \sim \Pi^m$, if an algorithm learns hypothesis $\fhatz = (\Dhatz, \whatz)$ such that $\varphi_{\Dhatz}$ is $s$-sparse on $(\mathbf{x} \cup \mathbf{x''})$, $\mu_{2s}(\fhatz) > 0$, and
\begin{align*}
m \geq \frac{43}{\smargin^2(\Dhatz, \mathbf{x} \cup \mathbf{x''}) \cdot \lambda}, 
\end{align*}
then the generalization error $(\Prob - \ProbZ) \lossof{\fhatz}$ is 
\begin{align}
\tilde{O} \left(
\frac{L}{\sqrt{m}} \left( \frac{r k \sqrt{s}}{\mu_{2s}(\fhatz)} \right) 
+ b \sqrt{\frac{(k^2 + \log \frac{1}{\delta}) \log m} {m}} 
+ \frac{L}{m} \left( \frac{r \sqrt{s}}{\lambda \mu_s(\fhatz)} \right)
\right) .
\label{eqn:essential-infinite}
\end{align}

\end{Theorem}

\subsection{Discussion of Theorems \ref{thm:final-learning-bound-overcomplete} and \ref{thm:final-learning-bound-infinite}}

The results highlight the central role of the stability of the sparse encoder. 
The presented bounds are data-dependent and exploit properties relating to the training sample and the learned hypothesis. 
Since $k \geq d$ in the overcomplete setting, an ideal learning bound has minimal dependence on $k$. 
The $\frac{1}{m}$ term of the learning bound for the overcomplete setting \eqref{eqn:essential-overcomplete} exhibits square root dependence on both the size of the dictionary $k$ and the ambient dimension $d$. It is unclear whether further improvement is possible, even in the reconstructive setting. The two known results in the reconstructive setting were established first by \cite{maurer2010k} and later by \cite{vainsencher2011sample}, as mentioned in the Introduction. 
The infinite-dimensional setting learning bound \eqref{eqn:essential-infinite} is dimension free, with linear dependence on $k$, square root dependence on $s$, and inverse dependence on the $2s$-incoherence $\mu_{2s}(\fhatz)$. 
While both bounds exhibit dependence on the sparsity level $s$, the sparsity level appears to be much more significant in the infinite-dimensional setting. 

Let us compare these bounds to the reconstructive setting, starting with the overcomplete regime. The first term of \eqref{eqn:essential-overcomplete} matches the slower of the rates shown by \cite{vainsencher2011sample} for the unsupervised case. Vainsencher et al. also showed fast rates of $\frac{d k}{m}$ (plus a small fraction of the observed empirical risk), but in the predictive setting it is an open question whether similar fast rates are possible. The second term of \eqref{eqn:essential-overcomplete} represents the error in approximating the estimator via an ($\varepsilon = \frac{1}{m}$)-cover of the space of dictionaries. This term reflects the stability of the sparse codes with respect to dictionary perturbations, as quantified by the Sparse Coding Stability Theorem (Theorem \ref{thm:new-sparse-coding-stability}). The reason for the lower bound on $m$ is that the $\varepsilon$-net used to approximate the space of dictionaries needs to be fine enough to satisfy the PRP condition \eqref{eqn:prp} of the Sparse Coding Stability Theorem. Hence, both this lower bound and the second term are determined primarily by the Sparse Coding Stability Theorem, and so with this proof strategy the extent to which the Sparse Coding Stability Theorem cannot be improved also indicates the extent to which Theorem \ref{thm:final-learning-bound-overcomplete} cannot be improved.

Shifting to the infinite-dimensional setting, 
\cite{maurer2010k} previously showed the generalization bound \eqref{eqn:unsupervised-maurer-pontil-bound} for unsupervised ($\ell_1$-regularized) sparse coding. 
Comparing their result to \eqref{eqn:essential-infinite} and neglecting regularization parameters, the dimension-free bound in the predictive case is larger by a factor of $\frac{\sqrt{s}}{\mu_{2s}(\hat{f}_z)}$. It is unclear whether either of the terms in this factor are avoidable in the predictive setting. At least from our analysis, it appears that the $\frac{\sqrt{s}}{\mu_{2s}(\hat{f}_z)}$ factor is the price one pays for encoder stability. 
Critically, encoder stability is not necessary in the reconstructive setting because stability in loss (reconstruction error) requires only stability in the \emph{norm of the residual} to the LASSO problem rather than stability in the \emph{value of the solution} to the problem. Stability of the norm of the residual is readily obtainable without any of the incoherence, sparsity, and margin conditions used here.

\paragraph{Remarks on conditions}
One may wonder about typical values for the various hypothesis-and-data-dependent properties in Theorems \ref{thm:final-learning-bound-overcomplete} and \ref{thm:final-learning-bound-infinite}. 
In practical applications of reconstructive and predictive sparse coding, the regularization parameter $\lambda$ is set to ensure that $s$ is small relative to the dimension $d$. As a result, both incoherences $\mu_s(D)$ and $\mu_{2s}(D)$ for the learned dictionary can be expected to be bounded away from zero. A sufficiently large $s$-incoherence certainly is necessary if one hopes for any amount of stability of the class of sparse coders with respect to dictionary perturbations. Since our path to reaching Theorems \ref{thm:final-learning-bound-overcomplete} and \ref{thm:final-learning-bound-infinite} passes through the Sparse Coding Stability Theorem (Theorem \ref{thm:new-sparse-coding-stability}), it seems that a drastically different strategy needs to be used if it is possible to avoid dependence on $\mu_s(D)$ in the learning bounds. 

A curious aspect of both learning bounds is their dependence on the $s$-margin term $\smargin(D, \mathbf{x})$. 
Suppose that a dictionary is learned which is $s$-sparse on the training sample $\mathbf{x}$, and $s$ is the lowest such integer for which this holds. It may not always be the case that the $s$-margin is bounded away from zero because for some points a small collection of inactive atoms may be very close to being brought into the optimal solution (the code); however, we can instead use the $(s+\rho)$-margin for some small positive integer $\rho$ for which the $(s+\rho)$-margin is non-trivial. 
In Section \ref{sec:empirical}, we gather empirical evidence that such a non-trivial $(s+\rho)$-margin does exist, for small $\rho$, when learning predictive sparse codes on real data. Hence, there is evidence that predictive sparse coding learns a dictionary with high $s$-incoherence $\mu_s(D)$ and non-trivial $s$-margin $\smargin(D, \mathbf{x})$ on the training sample, for low $s$. 

If one entertains a mixture of $\ell_1$ and $\ell_2$ norm regularization, 
$\lambda_1 |\cdot\|_1 + \frac{1}{2} \lambda_2 \|\cdot\|_2^2$, 
as in the elastic net \citep{zou2005regularization}, fall-back guarantees are possible in both scenarios. 
For small values of $\lambda_2$, this regularizer induces true sparsity similar to the $\ell_1$ regularizer. 
A considerably simpler, data-independent analysis is possible in the overcomplete setting with a final bound that essentially just trades $\mu_s(D)$ for the $\ell_2$ norm regularization parameter $\lambda_2$. In the infinite-dimensional setting, a simpler non-data-dependent analysis using our approach would only attain a bound of the larger order $\frac{k^{3/2}}{\lambda_2 \sqrt{m}}$.

\section{Tools}
\label{sec:tools}

As before, let $\mathbf{z}$ be a labeled sample of $m$ points (an $m$-sample) drawn iid from $\Prob$. In addition, let \glsadd{zp} $\mathbf{z'}$ be a second labeled $m$-sample drawn iid from $\Prob$. In the infinite-dimensional setting, we also will make use of an unlabeled $m$-sample $\mathbf{x''}$ drawn iid from the marginal $\Pi$. Also, an \emph{epsilon-cover} will be used to refer to the \emph{concept} of an $\varepsilon$-cover but not any specific cover. All epsilon-covers of spaces of dictionaries use the metric induced by the operator norm $\|\cdot\|_2$. 

\subsection{Symmetrization by ghost sample for random subclasses}

The next result is essentially due to \cite{mendelson2004importance}; it applies symmetrization by a ghost sample for random subclasses. Our main departure is that we allow the random subclass to depend on a second, unlabeled sample $\mathbf{x''}$. 
\begin{Lemma}[Symmetrization by Ghost Sample]
\label{lemma:symmetrization-ghost}
Let $\F(\mathbf{z},\mathbf{x''}) \subset \F$ be a random subclass which can depend on both a labeled sample $\mathbf{z}$ and an unlabeled sample $\mathbf{x''}$. Recall that $\mathbf{z'}$ is a ghost sample of $m$ points. If $m \geq \left(\frac{b}{t}\right)^2$, then
\begin{multline*}
{\Pr}_{\mathbf{z \, x''}} \left\{
\exists f \in \F(\mathbf{z},\mathbf{x''}) ,\,\, 
(\Prob - \ProbZ) \lossof{f} \geq t
\right\} \\
\leq 
2 {\Pr}_{\mathbf{z \, z' x''}} \left\{
\exists f \in \F(\mathbf{z},\mathbf{x''}) ,\,\, 
(\ProbZp - \ProbZ) \lossof{f} \geq \frac{t}{2}
\right\} .
\end{multline*}
\end{Lemma}
For completeness, this lemma is proved in Appendix \ref{sec:proof-symmetrization-ghost-lemma}. 
This symmetrization lemma will be applied in both the overcomplete and infinite-dimensional settings to shift the analysis from large deviations of the empirical risk from the expected risk to large deviations of two independent empirical risks: in the overcomplete setting the lemma will be specialized as Proposition \ref{prop:symmetrization-ghost-overcomplete}, and in the infinite-dimensional setting the lemma will be adapted to Proposition \ref{prop:symmetrization-ghost-infinite}.

\subsection{Rademacher and Gaussian averages and related results}
\label{subsec:rademacher-gaussian-appendix}

Let \glsadd{rademachervar} $\sigma_1, \ldots, \sigma_m$ be independent Rademacher random variables distributed uniformly on $\{-1,1\}$, and let \glsadd{gaussianvar} $\gamma_1, \ldots \gamma_m$ be independent Gaussian random variables distributed as $\mathcal{N}(0,1)$. Denote the collections by $\bm{\sigma} = (\sigma_1, \ldots, \sigma_m)$ and $\bm{\gamma} = (\gamma_1, \ldots, \gamma_m)$. 
Given a sample of $m$ points $\mathbf{x}$, define the conditional Rademacher and Gaussian averages of a function class as \glsadd{conditionalrademacher} \glsadd{conditionalgaussian}
\begin{align*}
\rademacher_{m \mid \mathbf{x}}(\F) 
= \frac{2}{m} \E_{\bm{\sigma}} \sup_{f \in \F} \sum_{i=1}^m \sigma_i f(x_i) 
\qquad \text{and} \qquad 
\gaussian_{m \mid \mathbf{x}}(\F) 
= \frac{2}{m} \E_{\bm{\gamma}} \sup_{f \in \F} \sum_{i=1}^m \gamma_i f(x_i) .
\end{align*}
respectively. 

Lemmas \ref{lemma:rademacher-loss} and \ref{lemma:rademacher-gaussian-bound} below are used near the end of the proof of Theorem \ref{thm:rademacher-mostly-good} of the infinite-dimensional setting, when shifting the analysis from the Gaussian complexity of a loss-composed function class to the Rademacher complexity of the original function class. 
From \citet[Theorem 7]{meir2003generalization}, the loss-composed conditional Rademacher average of a function class $\F$ is bounded by the scaled conditional Rademacher average:
\begin{Lemma}[Rademacher Loss Comparison Lemma]
\label{lemma:rademacher-loss}
For every function class $\F$, $m$-sample $\mathbf{x}$, and $\loss$ which is $L$-Lipschitz continuous in its second argument:
\begin{align*}
\rademacher_{m \mid \mathbf{z}}(\loss \circ \F)
\leq L \rademacher_{m \mid \mathbf{x}}(\F) .
\end{align*}
\end{Lemma}

Additionally, from \citet[a brief argument following Lemma 4.5]{ledoux1991probability}, the conditional Rademacher average of a function class $\F$ is bounded up to a constant by the conditional Gaussian average of $\F$:
\begin{Lemma}[Rademacher-Gaussian Average Comparison Lemma]
\label{lemma:rademacher-gaussian-bound}
For every function class $\F$ and sample of $m$ points $\mathbf{x}$:
\begin{align*}
\rademacher_{m \mid \mathbf{x}}(\F)
\leq \sqrt{\frac{\pi}{2}} \gaussian_{m \mid \mathbf{x}}(\F) .
\end{align*}
\end{Lemma}

The next relation is due to \cite{slepian1962one}:
\begin{Lemma}[Slepian's Lemma]
\label{lemma:slepian}
Let $\Omega$ and $\Gamma$ be mean zero, separable Gaussian processes\footnote{$\{\Omega_t\}_{t \in T}$ is a Gaussian process with index set $T$ if the collection is jointly Gaussian in the sense that every finite linear combination of the variables is Gaussian.} indexed by a set $T$ such that
$\E \left( \Omega_{t_1} - \Omega_{t_2} \right)^2 \leq \E \left( \Gamma_{t_1} - \Gamma_{t_2} \right)^2$for all $t_1, t_2 \in T$. Then $\E \sup_{t \in T} \Omega_t \leq \E \sup_{t \in T} \Gamma_t$.
\end{Lemma}
Slepian's Lemma essentially says that if the variance of one Gaussian process is bounded by the variance of another, then the expected maximum of the first is bounded by the expected maximum of the second. This lemma will be used in the proof of Theorem \ref{thm:gaussian-average-fixed-S-w} to bound the Gaussian complexity of an analytically difficult function class via a bound on the Gaussian complexity of a related but analytically easier function class. 

We also will make use of the following bounded differences inequality due to \cite{mcdiarmid1989method}, 
in order to shift the analysis in the proof of Theorem \ref{thm:rademacher-mostly-good}  to the Rademacher complexity of a certain function class:
\begin{Theorem}[McDiarmid's Inequality]
\label{thm:mcdiarmid}
Let $X_1, \ldots, X_m$ be random variables drawn iid according to a probability measure $\mu$ over a space $\X$. Suppose that a function $f: \X^m \rightarrow \real$ satisfies
\begin{align*}
\sup_{x_1, \ldots, x_m, x_i' \in \X} \left| f(x_1, \ldots, x_n) - f(x_1, \ldots, x_{i-1}, x_i', x_{i+1}, \ldots, x_m \right| \leq c_i
\end{align*}
for any $i \in [m]$. 
Then
\begin{align*}
{\Pr}_{X_1, \ldots, X_n} \bigl\{ f(X_1, \ldots, X_n) - \E f(X_1, \ldots, X_n) \geq t \bigr\} \leq \exp \left( -2t^2 / \sum_{i=1}^m c_i^2 \right).
\end{align*}
\end{Theorem}

\section{Overcomplete setting}
\label{sec:overcomplete}

\begin{figure}[t]
\fbox{
\includegraphics[width=\linewidth, clip=true, trim = 0mm 0mm 0mm 50mm]{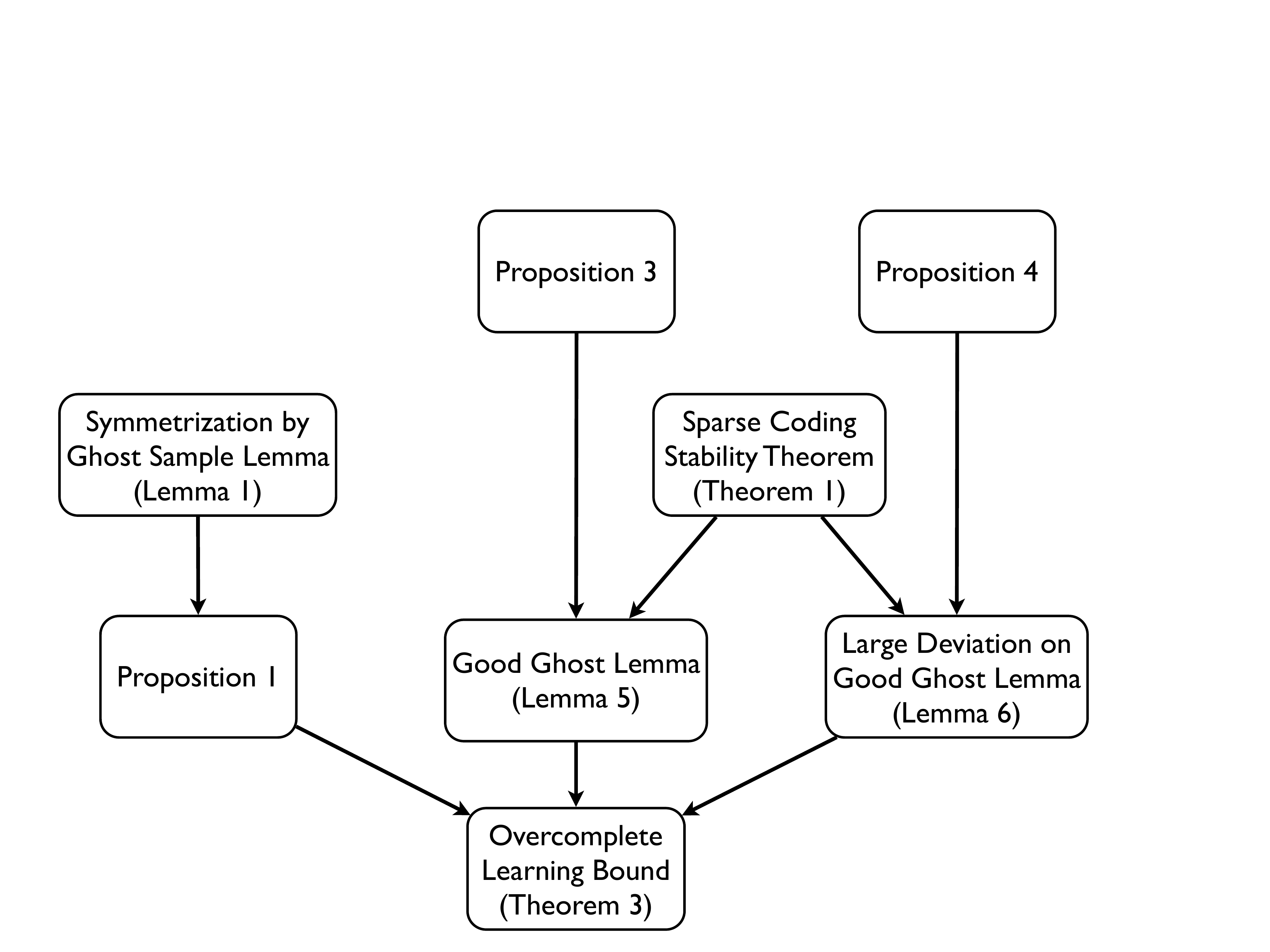}
}
\caption{
\label{fig:overcomplete-proof-flowchart}
Proof flowchart for the Overcomplete Learning Bound (Theorem \ref{thm:final-learning-bound-overcomplete}). 
}
\end{figure}

The overcomplete setting is classically the more popular regime, and in this setting useful learning bounds may exhibit dependence on both the dimension $d$ and the dictionary size $k$. 
At a high level, our strategy for the overcomplete case learning bound is to construct an epsilon-cover over a subclass of the space of functions 
$\F := \{f = (D,w): D \in \Dspace, w \in \Wspace\}$ 
and to show that the metric entropy of this subclass is of order $d k$. The main difficulty is that an epsilon-cover over $\Dspace$ need not approximate $\F$ to any degree, \emph{unless} one has a notion of encoder stability. 
Our analysis effectively will be concerned only with a training sample and a ghost sample, and it is similar in style to the luckiness framework of  \cite{shawe1998structural}. If we observe that the sufficient conditions for encoder stability hold true on the training sample, then it is enough to guarantee that most points in a ghost sample also satisfy these conditions (at a weaker level). Figure \ref{fig:overcomplete-proof-flowchart} exhibits the high-level flow of the proof of Theorem \ref{thm:final-learning-bound-overcomplete}.

\subsection{Useful conditions and subclasses}

Let \glsadd{allbuteta} $\mathbf{\tilde{x}} \subseteq_\eta \mathbf{x}$ indicate that $\mathbf{\tilde{x}}$ is a subset of $\mathbf{x}$ with at most $\eta$ elements of $\mathbf{x}$ removed. This notation is identical to
\cite{shawe1998structural}'s notation from the luckiness framework.

Our bounds will require a crucial PRP-based condition that depends on both the learned dictionary and the training sample:
\begin{align*}
\smargin(D,\mathbf{x}) \geq \iota(\lambda,\mu,\varepsilon) && \text{for} \,\,\,\, \iota(\lambda,\mu,\varepsilon) = \sqrt{\frac{387 \varepsilon}{\lambda}} .
\end{align*}
For brevity we will refer to $\iota$ with its parameters implicit; the dependence on $\varepsilon$, $\lambda$, and $\mu$ will not be an issue because we first develop bounds with these quantities fixed \emph{a priori}. 
Lastly, for $\mu > 0$ define \glsadd{Dspacemu} $\Dspace_\mu := \{D \in \Dspace: \mu_s(D) \geq \mu\}$
and \glsadd{Fmu} $\F_\mu := \{f = (D,w) \in \F: D \in \Dspace_\mu\}$.

\subsection{Learning bound}

The following proposition is simply a specialization of Lemma \ref{lemma:symmetrization-ghost} with $\mathbf{x''}$ taken as the empty set and 
$\F(\mathbf{z}, \mathbf{x''}) := \{ f \in \F_\mu : \smargingt{D, \mathbf{x}, \iota} \}$. 
\begin{Proposition}
\label{prop:symmetrization-ghost-overcomplete}
If $m \geq \left(\frac{b}{t}\right)^2$, then
\begin{align*}
&{\Pr}_{\mathbf{z}} \left\{\exists f \in \F_\mu ,\,\, 
\smargingt{D, \mathbf{x}), \iota} 
\booland
\left( (\Prob - \ProbZ) \lossof{f} > t \right) \right\} \\
&\leq 2 {\Pr}_{\mathbf{z \, z'}} \left\{\exists f \in \F_\mu ,\,\, 
\smargingt{D, \mathbf{x}, \iota} \booland 
\left( (\ProbZp - \ProbZ) \lossof{f} > t/2 \right) \right\}. 
\end{align*}

\end{Proposition}

In the RHS of the above, let the event whose probability is being measured be
\[
J := \left\{ \mathbf{z \, z'}: \exists f \in \F_\mu ,\,\, 
\smargingt{D, \mathbf{x}, \iota} \booland 
\left( \ProbZp - \ProbZ) \lossof{f} > t/2 \right) \right\}.
\]

Define $Z$ as the event that there exists a hypothesis with stable codes on the original sample, in the sense of the Sparse Coding Stability Theorem (Theorem \ref{thm:new-sparse-coding-stability}), but more than 
$\eta = \eta(m,d,k,D, \mathbf{x},\delta)$ 
points\footnote{We use the shorthand $\eta = \eta(m,d,k,D,\mathbf{x},\delta)$ for conciseness.} of the ghost sample whose codes are not guaranteed stable by the Sparse Coding Stability Theorem:
\begin{align*}
Z:= \left\{
\mathbf{z \, z'}: 
\begin{array}{l}
\exists f \in \F_\mu ,\,\, \smargingt{D, \mathbf{x}, \iota} \\
\qquad \booland
\left( \nexists \, \mathbf{\tilde{x}} \subseteq_\eta \mathbf{x'} \,\, \smargingt{D, \mathbf{\tilde{x}}, \frac{1}{3} \smargin(D, \mathbf{x})} \right)
\end{array}
\right\} .
\end{align*}
Our strategy will be to show that $\Pr(J)$ is small by use of the fact that
\[
\Pr(J) = \Pr(J \cap \bar{Z}) +  \Pr(J \cap Z) \leq \Pr(J \cap \bar{Z}) + \Pr(Z)
 ,
\]
a strategy which thus far is similar to the beginning of Shawe-Taylor et al.'s proof of the main luckiness framework learning bound \cite[see][Theorem 5.22]{shawe1998structural}. We now show that each of $\Pr(Z)$ and $\Pr(J \cap \bar{Z})$ is small in turn.

The imminent Good Ghost Lemma shadows \cite{shawe1998structural}'s notion of probable smoothness and provides a bound on $\Pr(Z)$.
\begin{Lemma}[Good Ghost]
\label{lemma:good-ghost}
Fix $\mu, \lambda > 0$ and $s \in [k]$. 
With probability at least $1 - \delta$ over an $m$-sample $\mathbf{x} \sim \Prob^m$ and a second $m$-sample $\mathbf{x'} \sim \Prob^m$, for any $D \in \Dspace_\mu$ for which $\varphi_D$ is $s\text{-}\mathrm{sparse}$ on $\mathbf{x}$, at least $m - \eta(m,d,k,D,\mathbf{x},\delta)$ points $\mathbf{\tilde{x}} \subseteq \mathbf{x'}$ satisfy $\smargingt{D, \mathbf{\tilde{x}}, \frac{1}{3} \smargin(D,\mathbf{x})}$, 
for
\begin{align*}
\eta(m,d,k,D,\mathbf{x},\delta) := 
d k \log \frac{3096}{\smargin^2(D,\mathbf{x}) \cdot \lambda} + \log (2 m + 1) + \log \frac{1}{\delta} .
\end{align*}
\end{Lemma}

\begin{figure}[h]
\fbox{
\includegraphics[width=\linewidth, clip=true, trim = 0mm 12mm 0mm 0mm]{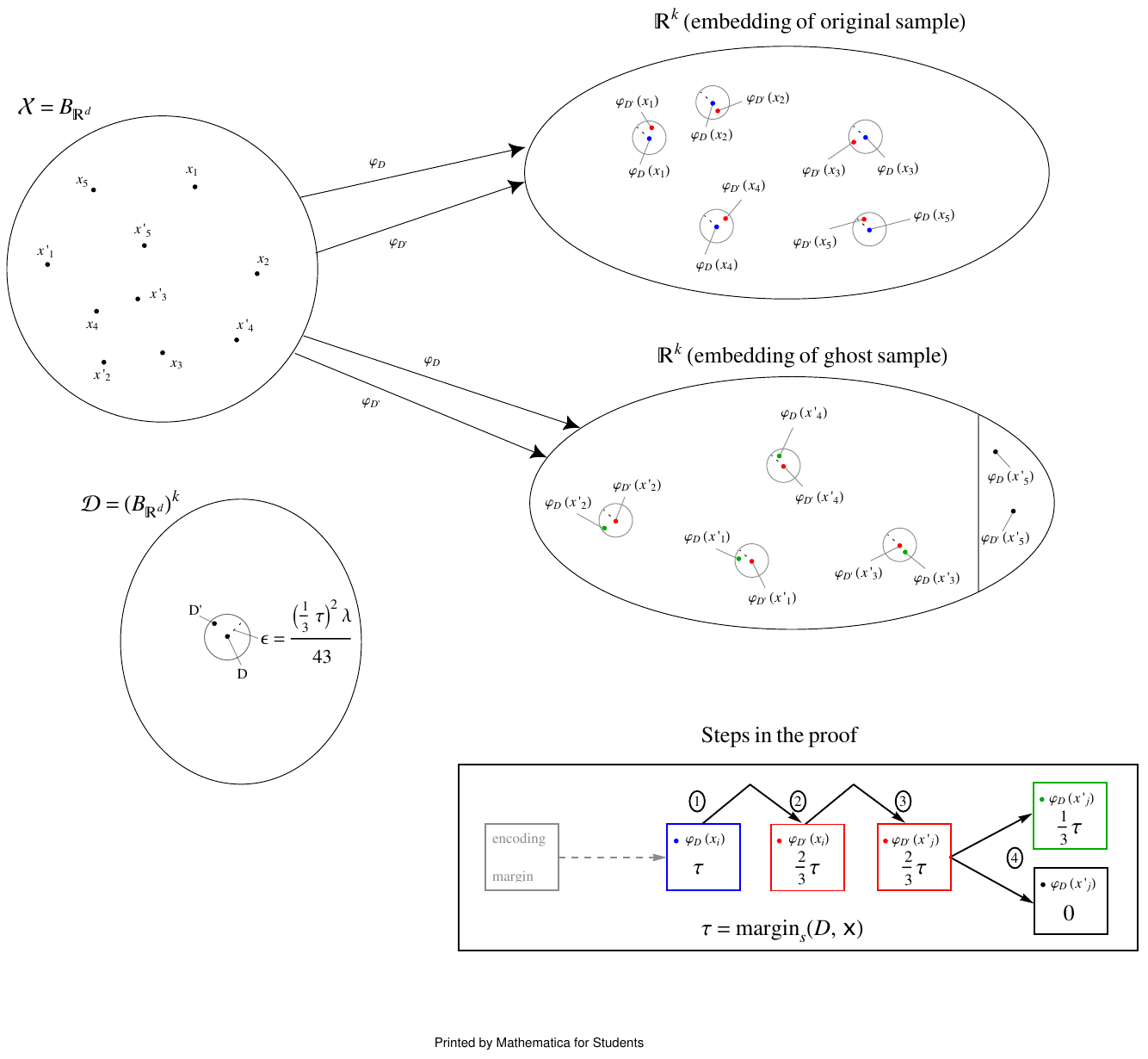}
}
\caption{
\label{fig:good-ghost}
Visualization of the proof of the Good Ghost Lemma (Lemma \ref{lemma:good-ghost}). Best seen in color.
}
\end{figure}
\begin{Proof}
Figure \ref{fig:good-ghost} illustrates the proof. By the assumptions of the lemma, consider an arbitrary dictionary $D$ satisfying $\mu_s(D) \geq \mu$ and $\sparsephi{s, D, \mathbf{x}}$. 
The goal is to guarantee with high probability that all but $\eta$ points of the ghost sample are coded by $\varphi_D$ with $s$-margin of at least $\frac{1}{3} \smargin(D,\mathbf{x})$. 

Let $\varepsilon = \frac{(\frac{1}{3} \smargin(D,\mathbf{x}))^2 \cdot \lambda}{43}$, and consider a minimum-cardinality $\varepsilon$-proper cover $\Dspace'$ of $\Dspace_\mu$. Let $D'$ be a candidate element of $\Dspace'$ satisfying $\|D - D'\|_2 \leq  \varepsilon$. 
Then the Sparse Coding Stability Theorem (Theorem \ref{thm:new-sparse-coding-stability}) implies that the coding margin of $D'$ on $\mathbf{x}$ retains over two-thirds the coding margin of $D$ on $\mathbf{x}$; that is, $\smargingt{D', \mathbf{x}, \frac{2}{3} \smargin(D,\mathbf{x})}$. 

Furthermore, we can and will show that most points from the \emph{ghost} sample satisfy \smargingt{D', \cdot, \frac{2}{3} \smargin(D,\mathbf{x})}. 
Let $\F^{\mathrm{marg}}_D := \{f^{\mathrm{marg}}_{D,\tau} | \tau \in \real_+\}$ be the class of threshold functions defined via
\begin{align*}
f^{\mathrm{marg}}_{D,\tau}(x) :=
\begin{cases}
1; & \text{if } \smargin(D, x) > \tau, \\
0; & \text{otherwise.}
\end{cases}
\end{align*}
Since the VC dimension of the one-dimensional threshold functions is 1, it follows that the $\mathrm{VC}(\F^{\mathrm{marg}}_D) = 1$. 
By using the VC dimension of $\F^{\mathrm{marg}}_D$ and the standard permutation argument of \citet[Proof of Theorem 2]{vapnik1968uniform}, it follows that for a single, \emph{fixed} element of $\Dspace'$, with probability at least $1 - \delta$ at most $\log (2 m + 1) + \log \frac{1}{\delta}$ points from a ghost sample will violate the margin inequality in question. Hence, 
by the bound on the proper covering numbers provided by Proposition \ref{prop:proper-covering-cardinality} (see Appendix \ref{sec:covering-numbers-appendix}), we can we can guarantee for all candidate members $D' \in \Dspace'$ that with probability $1 - \delta$ at most
\begin{align*}
\eta(m,d,k,D,\mathbf{x},\delta) = d k \log \frac{3096}{\smargin^2(D,\mathbf{x}) \cdot \lambda} + \log (2 m + 1) + \log \frac{1}{\delta} 
\end{align*}
points from the ghost sample violate the $s$-margin inequality. 
Thus, for arbitrary $D' \in \Dspace'$ satisfying the conditions of the lemma, with probability $1 - \delta$ at most
$\eta(m,d,k,D, \mathbf{x},\delta)$ points from the ghost sample violate $\smargingt{D', \cdot, \frac{2}{3} \smargin(D,\mathbf{x})}$. 

Finally, consider the at least $m - \eta(m,d,k,D,\mathbf{x},\delta)$ points in the ghost sample that satisfy $\smargingt{D', \cdot, \frac{2}{3} \smargin(D,\mathbf{x})}$. 
Since $\|D' - D\|_2 \leq \frac{(\frac{1}{3} \smargin(D,\mathbf{x}))^2 \cdot \lambda}{43}$, the Sparse Coding Stability Theorem (Theorem \ref{thm:new-sparse-coding-stability}) implies that these points satisfy $\smargingt{D, \cdot, \frac{1}{3} \smargin(D,\mathbf{x})}$. 
\end{Proof}

It remains to bound $\Pr(J \cap \bar{Z})$. 
\begin{Lemma}[Large Deviation on Good Ghost]
\label{lemma:large-deviation-good-ghost}
Define $\varpi := t / 2 - \left( 2 L \beta + \frac{b \eta}{m} \right)$ 
and $\beta := \frac{\varepsilon}{\lambda} \left( 1 + \frac{3 r \sqrt{s}}{\mu} \right)$. 
Then
\[
\Pr(J \cap \bar{Z})
\leq \left( \frac{8 (r/2)^{1/(d+1)}}{\varepsilon} \right)^{(d+1) k}
     \exp(-m \varpi^2 / ( 2 b^2)).
\]

Equivalently, 
the difference between the loss on $\mathbf{z}$ and the loss on $\mathbf{z'}$ is greater than $\varpi + 2 L \beta + \frac{b \eta}{m}$ with probability at most 
$\left( \frac{8 (r/2)^{1/(d+1)}}{\varepsilon} \right)^{(d+1) k} \exp(-m \varpi^2 / (2 b^2))$. 
\end{Lemma}

\begin{Proof}
Let $\goodghost$ represent the condition that all but at most $\eta$ points in the ghost sample $\mathbf{x'}$ satisfy 
$\smargingt{D, \cdot, \frac{1}{3} \smargin(D, \mathbf{x})}$ 
(and hence are ``good''), and let $\badghost$ be true if and only if $\goodghost$ is false. 

First, note that the event $J \cap \bar{Z}$ is a subset of the event
\begin{align*}
R := \left\{ \mathbf{z z'}: 
\begin{array}{l}
\exists f \in \F_\mu ,\,\, 
\smargingt{D, \mathbf{x}, \iota} \booland \\ 
\qquad \left( \exists \mathbf{\tilde{x}} \subseteq_\eta \mathbf{x'}, \,\, \smargingt{D, \mathbf{\tilde{x}}, \frac{1}{3} \smargin(D,\mathbf{x})} \right) \\
\qquad \booland \left( (\ProbZp - \ProbZ) \lossof{f} > t/2 \right)
\end{array}
\right\} .
\end{align*}
To see this, suppose that $\mathbf{z z'} \in J \cap \bar{Z}$. Since $\mathbf{z z'} \in J$, there is a particular $f = (D,w)$ in $\F_\mu$ satisfying $\smargingt{D, \mathbf{x}, \iota}$ and $(\ProbZp - \ProbZ) \lossof{f} > t/2$. Since $\mathbf{z z'} \in \bar{Z}$, there is no function in $\F_\mu$ satisfying both $\smargingt{D, \mathbf{x}, \iota}$ and $\badghost$. But since the particular $f$ satisfies $\smargingt{D, \mathbf{x}, \iota}$, it cannot also satisfy $\badghost$; thus, the particular $f$ satisfies $\goodghost$. 

Bounding the probability of the event $R$ is equivalent to bounding the probability of a large deviation (i.e. $\left( (\ProbZp - \ProbZ) \lossof{f} > t/2 \right)$) for the random subclass:
\begin{align*}
\tilde{\F}(\mathbf{x},\mathbf{x'}) := 
\left\{
\begin{array}{l}
f  \in \F_\mu: 
\smargingt{D, \mathbf{x}, \iota} \booland \\ 
\qquad 
\left( \exists \mathbf{\tilde{x}} \subseteq_\eta \mathbf{x'}, \,\, \smargingt{D, \mathbf{\tilde{x}}, \frac{1}{3} \smargin(D,\mathbf{x})} \right)
\end{array}
\right\} .
\end{align*}

Let $\F_\varepsilon = \DspaceEps \times \WspaceEps$, where $\DspaceEps$ is a minimum-cardinality proper $\varepsilon$-cover of $\Dspace_\mu$ and $\WspaceEps$ is a minimum-cardinality $\varepsilon$-cover of $\Wspace$. 
It is sufficient to bound the probability of a large deviation for all of $\F_\varepsilon$ and to then consider the maximum difference between an element of $\tilde{\F}(\mathbf{x},\mathbf{x'})$ and its closest representative in $\F_\varepsilon$. 
Clearly, for each $f = (D,w) \in \tilde{\F}(\mathbf{x}, \mathbf{x'})$, there is a $f' = (D',w') \in \F_\varepsilon$ satisfying $\|D - D'\|_2 \leq \varepsilon$ and $\|w - w'\|_2 \leq \varepsilon$. 
If $\varepsilon$ is sufficiently small, then for all but $\eta$ of the points $x_i$ in the ghost sample (and for all points $x_i$ of the original sample) it is guaranteed that
\begin{align*}
|  \langle w, \varphi_D(x_i) \rangle - \langle w', \varphi_{D'}(x_i) \rangle |
&\leq \left| \langle w - w', \varphi_D(x_i) \rangle \right| 
      + \left| \langle w', \varphi_D(x_i) - \varphi_{D'}(x_i) \rangle \right| \\
&\leq \frac{\varepsilon}{\lambda} 
      + r \frac{3 \varepsilon \sqrt{s}}{\lambda \mu} \\
&= \frac{\varepsilon}{\lambda} \left( 1 + \frac{3 r \sqrt{s}}{\mu} \right) = \beta ,
\end{align*}
where the second inequality follows from the Sparse Coding Stability Theorem (Theorem \ref{thm:new-sparse-coding-stability}). 
Trivially, for the rest of the points $x_i$ in the ghost sample there is a coarse guarantee that $\|\varphi_D(x_i) - \varphi_{D'}(x_i)\|_2 \leq \frac{2}{\lambda}$. Hence, on the original sample:
\begin{align*}
\frac{1}{m} \sum_{i=1}^m \left| \loss(y_i,\langle w, \varphi_D(x_i) \rangle) - \loss(y_i, \langle w', \varphi_{D'}(x_i) \rangle) \right| 
&\leq L \beta ,
\end{align*}
and on the ghost sample:
\begin{align*}
\frac{1}{m} & \sum_{i=1}^m \bigl| \loss(y'_i, \langle w, \varphi_D(x'_i) \rangle) - \loss(y'_i, \langle w', \varphi_{D'}(x'_i) \rangle) \bigr| \\
&\leq \frac{L}{m} \sum_{i \textsc{ good}} \left| \langle w, \varphi_D(x_i) \rangle - \langle w', \varphi_{D'}(x_i) \rangle \right| 
+ \frac{1}{m} \sum_{i \textsc{ bad}} 
\bigl| \loss(y'_i, \langle w, \varphi_D(x'_i) \rangle) - \loss(y'_i, \langle w', \varphi_{D'}(x'_i) \rangle) \bigr| \\
&\leq L \beta + \frac{b \eta}{m} ,
\end{align*}
where \textsc{good} denotes the at least $m - \eta$ points of the ghost sample for which the Sparse Coding Stability Theorem (Theorem \ref{thm:new-sparse-coding-stability}) applies, and \textsc{bad} denotes the complement thereof. 
To conclude the above argument, the difference between the losses of $f$ and $f'$ on the double sample will be at most $2 L \beta + \frac{b \eta}{m}$.

Now, if $\nu$ is the absolute deviation between the loss of $f$ on the original sample versus its loss on the ghost sample, then the absolute deviation between the loss of $f' = (D',w')$ on the original sample and the loss of $f'$ on the ghost sample must be at least
\[
\nu - \left( 2 L \beta + \frac{b \eta}{m} \right) .
\]
Consequently, if $\nu > t/2$, then the absolute deviation between the loss of $f'$ on the original sample and the loss of $f'$ on the ghost sample must be at least $t/2  - \left( 2 L \beta + \frac{b \eta}{m} \right)$. 
To bound the probability of $R$ it therefore is sufficient to control
\begin{align*}
{\Pr}_{\mathbf{z \, z'}} \left\{ \exists f = (D',w') \in \DspaceEps \times \WspaceEps ,\,\, (\ProbZp - \ProbZ) \lossof{f} > t/2 - \left( 2 L \beta + \frac{b \eta}{m} \right) \right\} .
\end{align*}

We first handle the case of a fixed $f = (D',w') \in \DspaceEps \times \WspaceEps$. Applying Hoeffding's inequality to the random variable $\loss(y_i,f(x_i)) - \loss(y'_i,f(x'_i))$, with range in $[-b,b]$, yields:
\begin{align*}
{\Pr}_{\mathbf{z \, z'}} \left\{ (\ProbZp - \ProbZ) \lossof{f} > \varpi \right\} \leq \exp(-m \varpi^2 / (2 b^2)),
\end{align*}
for $\varpi := t / 2 - \left( 2 L \beta + \frac{b \eta}{m} \right)$. By way of a proper covering number bound of $\DspaceEps \times \WspaceEps$ (see Proposition \ref{prop:hypothesis-covering-cardinality}) and the union bound, this result can be extended over all of $\DspaceEps \times \WspaceEps$:
\begin{multline*}
{\Pr}_{\mathbf{z \, z'}} \left\{ \exists f = (D',w') \in \DspaceEps \times \WspaceEps ,\,\, (\ProbZp - \ProbZ) \lossof{f} > \varpi \right\} \\
\leq \left( \frac{8 (r/2)^{1/(d+1)}}{\varepsilon} \right)^{(d+1) k}
       \exp(-m \varpi^2 / ( 2 b^2)).
\end{multline*}
The bound on $\Pr(J \cap \bar{Z})$ now follows.
\end{Proof}

The stage is now set to prove Theorem \ref{thm:final-learning-bound-overcomplete}; the full proof is in Appendix \ref{sec:overcomplete-proofs}. 
\begin{Proof-sketch}[of Theorem \ref{thm:final-learning-bound-overcomplete}] 
Proposition \ref{prop:symmetrization-ghost-overcomplete} and Lemmas \ref{lemma:good-ghost} and \ref{lemma:large-deviation-good-ghost} imply that
\begin{align*}
{\Pr}_{\mathbf{z}} & \left\{ \exists f \in \F_\mu ,\,\, 
\smargingt{D, \mathbf{x}, \iota} \booland
 \left( (\Prob - \ProbZ) \lossof{f} > t \right) \right\} \\
&\leq 2 \left( \left( \frac{8 (r/2)^{1/(d+1)}}{\varepsilon} \right)^{(d+1) k} \exp(-m \varpi^2 / ( 2 b^2)) + \delta \right) .
\end{align*}
Let $s \in [k]$ and $\mu > 0$ be fixed \emph{a priori}. Setting $\varepsilon = \frac{1}{m}$ in the above, elementary manipulations can show that provided $m > \frac{387}{\smargin(D,\mathbf{x})^2 \lambda}$,  with probability at least $1 - \delta$ over $\mathbf{z} \sim \Prob^m$, 
for any $f = (D,w) \in \F$ satisfying 
$\mu_s(D) \geq \mu$ and
$\smargingt{D, \mathbf{x}, \iota}$, 
the generalization error $(\Prob - \ProbZ) \lossof{f}$ is bounded by:
\begin{align*}
& \, 2 b \sqrt{\frac{2 ( (d+1) k \log (8 m) + k \log \frac{r}{2} + \log \frac{4}{\delta})}{m}} \\
& \, + \frac{4 L}{m} \left( \frac{1}{\lambda} \left( 1 + \frac{3 r \sqrt{s}}{\mu} \right) \right)
+ \frac{2 b}{m}
\left( d k \log \frac{3096}{\smargin^2(D,\mathbf{x}) \cdot \lambda} + \log (2 m + 1) + \log \frac{4}{\delta}
\right) .
\end{align*}

It remains to distribute a prior across the bounds for each choice of $s$ and $\mu$. To each choice of $s \in [k]$ assign prior probability $\frac{1}{k}$. To each choice of $i \in \nat \cup \{0\}$ for $2^{-i} \leq \mu$ assign prior probability $(i+1)^{-2}$. For a given choice of $s \in [k]$ and $2^{-i} \leq \mu$ we use
$\delta(s,i) := \frac{6}{\pi^2} \frac{1}{(i+1)^2} \frac{1}{k} \delta$ (since $\sum_{i=1}^\infty \frac{1}{i^2} = \frac{\pi^2}{6}$). The theorem now follows. 
\end{Proof-sketch}

\section{Infinite-dimensional setting}
\label{sec:infinite}

In the infinite-dimensional setting learning bounds with dependence on $d$ are useless. 
Unfortunately, the strategy of the previous section breaks down in the infinite-dimensional setting because the straightforward construction of any epsilon-cover over the space of dictionaries had cardinality that depends on $d$. Even worse, epsilon-covers actually were used both to approximate the function class $\F$ in $\|\cdot\|_\infty$ norm and to guarantee that most points of the ghost sample are good provided that all points of the training sample were good (the Good Ghost Lemma (Lemma \ref{lemma:good-ghost})). 

\begin{figure}[t]
\fbox{
\includegraphics[width=\linewidth, clip=true, trim = 0mm -5mm 0mm -5mm]{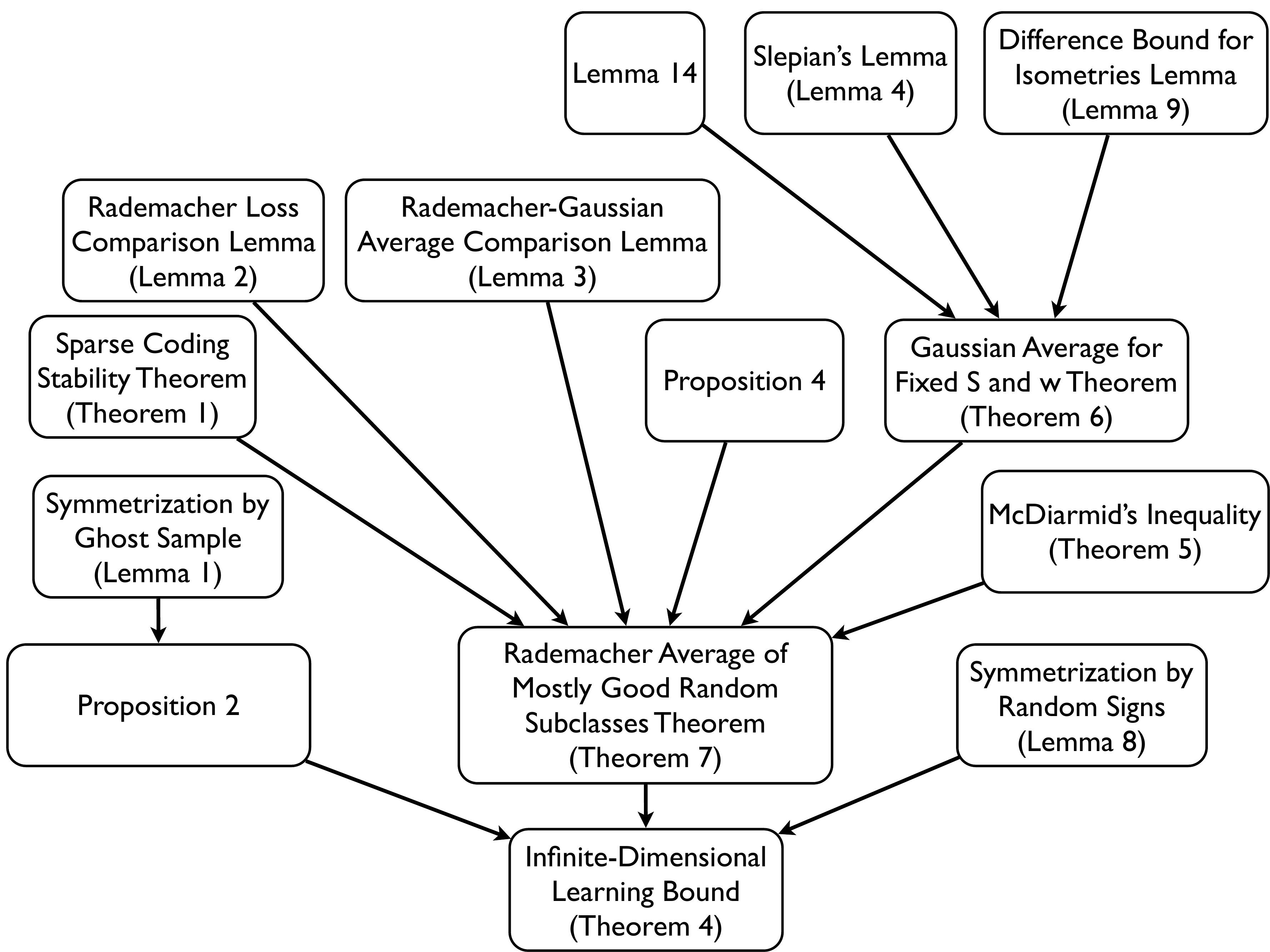}
}
\caption{
\label{fig:infinite-proof-flowchart}
Proof flowchart for the Infinite-Dimensional Learning Bound (Theorem \ref{thm:final-learning-bound-infinite}). 
}
\end{figure}

These issues can be overcome by requiring an additional, \emph{unlabeled} sample --- a device often justified in supervised learning problems because unlabeled data may be inexpensive and yet quite helpful --- and by switching to more sophisticated techniques based on conditional Rademacher and Gaussian averages. 
After learning a hypothesis $\fhatz$ from a predictive sparse coding algorithm, the sparsity level and coding margin are measured on a second, unlabeled sample $\mathbf{x''}$ of $m$ points\footnote{The cardinality matches the size of the training sample $\mathbf{z}$ purely for simplicity.}. Since this sample is independent of the choice of $\fhatz$, it is possible to guarantee that all but a very small fraction ($\frac{\eta}{m} = \frac{\log \frac{1}{\delta}}{m}$) of points of a ghost sample $\mathbf{z}$ are good with probability $1 - \delta$. In the likely case of this good event, and for a fixed sample, we then consider all possible choices of a set of $\eta$ bad indices in the ghost sample; each of the $m \choose \eta$ cases corresponds to a subclass of functions. We then approximate each subclass by a special $\varepsilon$-cover that is a disjoint union of a finite number of special subclasses; for each of these smaller subclasses, we bound the conditional Rademacher average by exploiting a sparsity property. 
The proof flowchart in Figure \ref{fig:infinite-proof-flowchart} shows the structure of the proof of Theorem \ref{thm:final-learning-bound-infinite}.

\subsection{Symmetrization and decomposition}

The proof of the infinite-dimensional setting learning bound Theorem \ref{thm:final-learning-bound-infinite} depends critically on Lemma \ref{lemma:sc-phi-difference-bound-WRT-U}, a lemma which is non-trivial only for dictionaries with non-zero $2s$-incoherence. The $s$-incoherence also will continue to play an important role, as it did in the overcomplete setting. Therefore, rather than wielding the deterministic subclass $\F_\mu$ of the previous section, we will work with a deterministic subclass with lower bounded $s$-incoherence \emph{and} lower bounded $2s$-incoherence.

Let $\bm{\mu^*} = (\mu^*_s, \mu^*_{2s}) \in \real_+^2$ and define the deterministic subclass \glsadd{Fmustar} 
\begin{align*}
\F_{\bm{\mu^*}} = \bigl\{ f = (D,w) \in \F: (\mu_s(D) \geq \mu^*_s) \booland (\mu_{2s}(D) \geq \mu^*_{2s}) \bigr\} .
\end{align*}
The next result is immediate from Lemma \ref{lemma:symmetrization-ghost}, taking the random subclass $\F(\cdot)$ to be
\begin{align*}
\F(\mathbf{z}, \mathbf{x''}) := \Bigl\{ \{\fhatz\} \cap \{ f \in \F_{\bm{\mu^*}} : \smargingt{D, \mathbf{x} \cup \mathbf{x''}, \tau} \Bigr\} .
\end{align*} 

\begin{Proposition}
\label{prop:symmetrization-ghost-infinite}
If $m \geq \left(\frac{b}{t}\right)^2$, then
\begin{align}
{\Pr}_{\mathbf{z \, x''}} &\left\{
\begin{array}{l}
\fhatz  \in \F_{\bm{\mu^*}} \,\,
\smargingt{\Dhatz, \mathbf{x \cup x''}, \tau} 
\booland 
\left( (\Prob - \ProbZ) \lossof{\fhatz} \geq t \right) 
\end{array}
\right\} \label{eqn:full-probability}\\
&\leq
2 {\Pr}_{\mathbf{z \, z' x''}} \left\{
\begin{array}{l}
\fhatz \in \F_{\bm{\mu^*}} \,\,
\smargingt{\Dhatz, (\mathbf{x} \cup \mathbf{x''}), \tau} 
\booland 
\left( (\ProbZp - \ProbZ) \lossof{\fhatz} \geq \frac{t}{2} \right)
\end{array}
\right\} . \nonumber
\end{align}
\end{Proposition}

Now, observe that the probability of interest can be split into the probability of a large deviation happening under a ``good'' event and the probability of a ``bad'' event occurring:  
\begin{align*}
{\Pr}_{\mathbf{z \, z' x''}} &\left\{ 
\begin{array}{l}
\fhatz \in \F_{\bm{\mu^*}} \,\,
\smargingt{\Dhatz, \mathbf{x} \cup \mathbf{x''}, \tau} 
\booland 
\left( (\ProbZp - \ProbZ) \lossof{\fhatz} \geq \frac{t}{2} \right) 
\end{array}
\right\} \\
= \quad &{\Pr}_{\mathbf{z \, z' x''}} \left\{ 
\begin{array}{l}
\fhatz \in \F_{\bm{\mu^*}} \,\,
\smargingt{\Dhatz, \mathbf{x} \cup \mathbf{x''}, \tau} 
\booland 
\left( \exists \, \mathbf{\tilde{x}} \subseteq_\eta \mathbf{x'} \,\, \smargingt{\Dhatz, \mathbf{\tilde{x}}, \tau} \right) \\
\qquad \booland 
\left( (\ProbZp - \ProbZ) \lossof{\fhatz} \geq \frac{t}{2} \right)
\end{array}
\right\} \\
+ \, &{\Pr}_{\mathbf{z \, z' x''}} \left\{ 
\begin{array}{l}
\fhatz \in \F_{\bm{\mu^*}} \,\,
\smargingt{\Dhatz, \mathbf{x} \cup \mathbf{x''}, \tau} 
\booland 
\left( \nexists \, \mathbf{\tilde{x}} \subseteq_\eta \mathbf{x'} \,\, \smargingt{\Dhatz, \mathbf{\tilde{x}}, \tau} \right) \\
\qquad \booland 
\left( (\ProbZp - \ProbZ) \lossof{\fhatz} \geq \frac{t}{2} \right)
\end{array}
\right\} \\
\leq \quad &{\Pr}_{\mathbf{z \, z'}} \left\{
\begin{array}{l}
\exists f \in \F_{\bm{\mu^*}} ,\,\,
\smargingt{\Dhatz, \mathbf{x}, \tau} 
\booland 
\left( \exists \, \mathbf{\tilde{x}} \subseteq_\eta \mathbf{x'} \,\, \smargingt{\Dhatz, \mathbf{\tilde{x}}, \tau} \right) \\
\qquad \booland 
\left( (\ProbZp - \ProbZ) \lossof{f} \geq \frac{t}{2} \right)
\end{array}
\right\} \\
+ \, &{\Pr}_{\mathbf{x' x''}} \left\{
\begin{array}{l}
\fhatz \in \F_{\bm{\mu^*}} \,\,
\smargingt{\Dhatz, \mathbf{x''}, \tau} 
\booland 
\left( \nexists \, \mathbf{\tilde{x}} \subseteq_\eta \mathbf{x'} \,\, \smargingt{\Dhatz, \mathbf{\tilde{x}}, \tau} \right) 
\end{array}
\right\} .
\end{align*}

Of the two probabilities summed in the last line, we treat the first in the next subsection. To bound the second one, note that for each choice of $\mathbf{x}$, $\fhatz$ is a fixed function. Hence, it is sufficient to select $\eta$ such that, for \emph{any} fixed function $f = (D,w) \in \F$, this second probability is bounded by $\delta$. The next lemma accomplishes this bound:
\begin{Lemma}[Unlikely Bad Ghost]
\label{lemma:unlikely-bad-ghost}
Let $f = (D,w) \in \F$ be fixed. 
If $\eta = \log \frac{1}{\delta}$, then 
\begin{align*}
{\Pr}_{\mathbf{x' x''}} \left\{
\begin{array}{l}
\smargingt{D, \mathbf{x''}, \tau} 
\booland 
\Bigl( \nexists \, \mathbf{\tilde{x}} \subseteq_\eta \mathbf{x'} \,\, \smargingt{D, \mathbf{\tilde{x}}, \tau} \Bigr) 
\end{array}
\right\} \leq \delta .
\end{align*}
\end{Lemma}
\begin{Proof-sketch} 
The proof just uses the same standard permutation argument as in the proof of the Good Ghost Lemma (Lemma \ref{lemma:good-ghost}). 
\end{Proof-sketch}

\subsection{Rademacher bound in the case of the good event}

We now bound the probability of a large deviation in the (likely) case of the good event. Denote by \glsadd{Fmustarx} $\F_{\bm{\mu^*}}(\mathbf{x})$ the intersection of the deterministic subclass $\F_{\bm{\mu^*}}$ with the random subclass of functions for which the Sparse Coding Stability Theorem (Theorem \ref{thm:new-sparse-coding-stability}) kicks in with constants ($\mu^*_s, s, \tau$):
\begin{align*}
\F_{\bm{\mu^*}}(\mathbf{x}) := 
\bigl\{ f \in \F_{\bm{\mu^*}}: 
\smargingt{D, \mathbf{x}, \tau} 
\bigr\}. 
\end{align*}
This is the ``good'' random subclass. 
Similarly, let \glsadd{Fmustaretax} $\F_{\bm{\mu^*},\eta}(\mathbf{x})$ denote the ``mostly good'' (or ``all-but-$\eta$-good'') random subclass:
\begin{align*}
\F_{\bm{\mu^*},\eta}(\mathbf{x}) := 
\Bigl\{ f \in \F_{\bm{\mu^*}}: 
\exists \, \mathbf{\tilde{x}} \subseteq_\eta \mathbf{x} \,\, 
\smargingt{D, \mathbf{\tilde{x}}, \tau} 
 \Bigr\} .
\end{align*} 

Recall that $\sigma_1, \ldots, \sigma_m$ are independent Rademacher random variables. 
\begin{Lemma}[Symmetrization by Random Signs]
\label{lemma:symmetrization-random-signs}
\begin{align*}
&{\Pr}_{\mathbf{z \, z'}} \left\{
\begin{array}{l}
\exists f \in \F_{\bm{\mu^*}} ,\,\, 
\smargingt{D, \mathbf{x}, \tau} 
\booland 
\left( \exists \, \mathbf{\tilde{x}} \subseteq_\eta \mathbf{x'} \,\, \smargingt{D, \mathbf{\tilde{x}}, \tau} \right) \\
\qquad \booland 
\left( (\ProbZp - \ProbZ) \lossof{f} \geq \frac{t}{2} \right)
\end{array}
\right\} \\
&\leq {\Pr}_{\mathbf{z},\bm{\sigma}} 
\left\{
\sup_{f \in \F_{\bm{\mu^*}}(\mathbf{x})} \frac{1}{m} \sum_{i=1}^m \sigma_i \loss(y_i,f(x_i)) \geq \frac{t}{4}
\right\}
+ {\Pr}_{\mathbf{z},\bm{\sigma}} 
\left\{
\sup_{f \in \F_{\bm{\mu^*},\eta}(\mathbf{x})} \frac{1}{m} \sum_{i=1}^m \sigma_i \loss(y_i,f(x_i)) \geq \frac{t}{4}
\right\} .
\end{align*}
\end{Lemma}

\begin{Proof}
From the definitions of the random subclasses $\F_{\bm{\mu^*}}(\cdot)$ and $\F_{\bm{\mu^*}, \eta}(\cdot)$, the left hand side in the lemma is equal to
\begin{align*}
 {\Pr}_{\mathbf{z \, z'}} \left\{
\sup_{f \in \F_{\bm{\mu^*}}(\mathbf{x}) \cap \F_{\bm{\mu^*},\eta}(\mathbf{x'})} 
\frac{1}{m} \sum_{i=1}^m \left( \loss(y'_i,f(x'_i)) - \loss(y_i,f(x_i)) \right) \geq \frac{t}{2}
\right\} .
\end{align*}
Now, by a routine application of symmetrization by random signs this is equal to
\begin{align*}
& {\Pr}_{\mathbf{z z'},\bm{\sigma}} 
\left\{
\sup_{f \in \F_{\bm{\mu^*}}(\mathbf{x}) \cap \F_{\bm{\mu^*},\eta}(\mathbf{x'})} \frac{1}{m} \sum_{i=1}^m \sigma_i (\loss(y'_i,f(x'_i)) - \loss(y_i,f(x_i))) \geq \frac{t}{2}
\right\} \\
&\leq {\Pr}_{\mathbf{z},\bm{\sigma}} 
\left\{
\sup_{f \in \F_{\bm{\mu^*}}(\mathbf{x})} \frac{1}{m} \sum_{i=1}^m \sigma_i \loss(y_i,f(x_i)) \geq \frac{t}{4}
\right\}
+ {\Pr}_{\mathbf{z},\bm{\sigma}} 
\left\{
\sup_{f \in \F_{\bm{\mu^*},\eta}(\mathbf{x})} \frac{1}{m} \sum_{i=1}^m \sigma_i \loss(y_i,f(x_i)) \geq \frac{t}{4}
\right\} .
\end{align*}
\end{Proof}

Since the good random subclass $\F_{\bm{\mu^*}}(\mathbf{x})$ is just the all-but-$0$-good random subclass $\F_{\bm{\mu^*},0}(\mathbf{x})$, it is sufficient to bound the second term of the last line above for arbitrary $\eta \in [m]$. For fixed $\mathbf{z}$, the randomness of the subclass is annihilated and the above supremum over $\F_{\bm{\mu^*},\eta}(\mathbf{x})$ is a conditional Rademacher average. Bounding this conditional Rademacher average will call for a few results on the \emph{Gaussian} average of a related function class. 

First, note that for any $D \in \Dspace$, the dictionary $D$ can be factorized as $D = U S$, where all $U \in \Uspace \subset \real^{d \times k}$ satisfy the isometry property $U^T U = I$, and $S$ lives in a space $\Sspace := \left(\ball{}{k}\right)^k$ of lower-dimensional dictionaries \citep{maurer2010k}. Consider a particular choice of $S \in \Sspace$, linear hypothesis $w \in \Wspace$, and $m$-sample $\mathbf{x}$. The subclass of interest will be those functions corresponding to $U \in \Uspace$ such that the encoder $\varphi_{U S}$ is $s$-sparse on $\mathbf{x}$. It turns out that the Gaussian average of this subclass is well-behaved.

Recall that $\bm{\gamma} = (\gamma_1, \ldots \gamma_m)$ where the $\gamma_i$ are iid standard normals. 
\begin{Theorem}[Gaussian Average for Fixed $S$ and $w$]
\label{thm:gaussian-average-fixed-S-w}
Let $S \in \Sspace$, $s \in [k]$, and $\mathbf{x}$ be a fixed $m$-sample. Denote by $\Uspace_{\mathbf{x}}$ the particular subclass of $\Uspace$ defined as:
\begin{align*}
\Uspace_{\mathbf{x}} := \bigl\{ U \in \Uspace: \sparsephi{s, U S, \mathbf{x}} \bigr\} .
\end{align*}
Then
\begin{align}
\E_{\bm{\gamma}} \sup_{U \in \Uspace_{\mathbf{x}}} \frac{2}{m} \sum_{i=1}^m \gamma_i \langle w, \varphi_{U S}(x_i) \rangle 
& \leq \frac{4 r k \sqrt{2 s}}{\mu_{2s}(S) \sqrt{m}}. \label{eqn:SC-Uspace-Gaussian-bound}
\end{align}
\end{Theorem}

The proof of this result uses the following lemma that shows how the difference between the feature maps $\varphi_{U S}$ and $\varphi_{U' S}$ can be  characterized by the difference between $U$ and $U'$. Define the \emph{$s$-restricted $2$-norm} of $S$ as \glsadd{srestricted2norm} $\| S \|_{2,s} := \sup_{\{t \in \real^n: \|t\|=1, |\support(t)| \leq s\}} \| S t \|_2$. 
\begin{Lemma}[Difference Bound for Isometries]
\label{lemma:sc-phi-difference-bound-WRT-U}

Let $U,U' \in \Uspace$ be isometries as above, $S \in \Sspace$, and $x \in \ball{}{d}$. If $\|\varphi_{U S}(x)\|_0 \leq s$ and $\|\varphi_{U' S}(x)\|_0 \leq s$, then
\begin{align*}
\| \varphi_{U S}(x) - \varphi_{U' S}(x) \|_2 
\leq 
\frac{2 \| S \|_{2,2s} }{\mu_{2s}(S)} \| ({U'}^T - U^T) x \|_2.
\end{align*}
\end{Lemma}
\begin{Proof-sketch} 
The proof uses a perturbation analysis of solutions to linearly constrained positive definite quadratic programs \citep{daniel1973stability}, exploiting the sparsity of the optimal solutions to have dependence only on $\|S\|_{2,2s}$ and $\mu_{2s}(S)$ rather than $\|S\|_2$ and $\mu_{k}(S)$. 
\end{Proof-sketch}

\begin{Proof}[of Theorem \ref{thm:gaussian-average-fixed-S-w}] 
Define a Gaussian process $\Omega$, indexed by $U$, by $\Omega_U := \sum_{i=1}^m \gamma_i \langle w, \varphi_{U S}(x_i) \rangle$.  Our goal is to apply Slepian's Lemma (Lemma \ref{lemma:slepian}) to bound the expectation of the supremum of $\Omega$, which depends on $\varphi_{U S}$, by the expectation of the supremum of a Gaussian process $\Gamma$ which depends only on $U$. 
\begin{align}
\E_{\bm{\gamma}} \, (\Omega_{U} - \Omega_{U'})^2
&= \E_{\bm{\gamma}} \left( \sum_{i=1}^m \gamma_i \left\langle w, \varphi_{U S}(x_i) \right\rangle - \sum_{i=1}^m \gamma_i \left\langle w, \varphi_{U' S}(x_i) \right\rangle \right)^2 \nonumber \\
&= \sum_{i=1}^m \bigl( \langle w, \varphi_{U S}(x_i) - \varphi_{U' S}(x_i) \rangle \bigr)^2 \nonumber \\
&\leq r^2 \sum_{i=1}^m \| \varphi_{U S}(x_i) - \varphi_{U' S}(x_i) \|^2 \label{eqn:midway-slepian}
\end{align}

Applying the result from Lemma \ref{lemma:sc-phi-difference-bound-WRT-U}, we have 
\begin{align*}
\E_{\bm{\gamma}} \, (\Omega_U - \Omega_{U'})^2
&\leq r^2 \sum_{i=1}^m \| \varphi_{U S}(x_i) - \varphi_{U' S}(x_i) \|^2 \\
&\leq \left( \frac{2 r \|S\|_{2,2s}}{\mu_{2s}(S)} \right)^2 \sum_{i=1}^m \left\| ({U'}^T - U^T) x_i \right\|_2^2 \\
&= \left( \frac{2 r \|S\|_{2,2s}}{\mu_{2s}(S)} \right)^2 \sum_{i=1}^m \sum_{j=1}^k (\langle U' e_j, x_i \rangle - \langle U e_j, x_i \rangle)^2 \\
&= \left( \frac{2 r \|S\|_{2,2s}}{\mu_{2s}(S)} \right)^2
\E_{\bm{\gamma}} 
\left(
\Biggl( \sum_{i=1}^m \sum_{j=1}^k \gamma_{i j} \langle U' e_j, x_i \rangle \Biggr)
-
\Biggl( \sum_{i=1}^m \sum_{j=1}^k \gamma_{i j} \langle U e_j, x_i \rangle \Biggr)
\right)^2 \\
&= \E_{\bm{\gamma}} \, (\Gamma_U - \Gamma_{U'})^2
\end{align*}
\begin{flalign*}
\qquad \text{for } && \Gamma_U := \frac{2 r \|S\|_{2,2s}}{\mu_{2s}(S)} \sum_{i=1}^m \sum_{j=1}^k \gamma_{i j} \langle U e_j, x_i \rangle. &&
\end{flalign*}
By Slepian's Lemma (Lemma \ref{lemma:slepian}), $\E_{\bm{\gamma}} \sup_U \Omega_U \leq \E_{\bm{\gamma}} \sup_U \Gamma_U$. It remains to bound $E_\gamma \sup_U \Gamma_U$:
\begin{alignat*}{2}
\frac{\mu_{2s}(S)}{2 r \|S\|_{2,2s}} \E_{\bm{\gamma}} \sup_U \Gamma_U 
&=&& \E_{\bm{\gamma}} \sup_U \sum_{i=1}^m \sum_{j=1}^k \gamma_{i j} \langle U e_j, x_i \rangle  \\
&=&& \E_{\bm{\gamma}} \sup_U \sum_{j=1}^k \langle U e_j, \sum_{i=1}^m \gamma_{i j} x_i \rangle  \\
&\leq&& \E_{\bm{\gamma}} \sup_U \sum_{j=1}^k \| U e_j \| \| \sum_{i=1}^m \gamma_{i j} x_i \|  \\
&=&& k \E_{\bm{\gamma}} \, \| \sum_{i=1}^m \gamma_{i 1} x_i \|  \\
&\leq&& k \sqrt{ \E_{\bm{\gamma}} \, \| \sum_{i=1}^m \gamma_{i 1} x_i \|^2 }  \\
&=&& k \sqrt{ \E_{\bm{\gamma}} \left\langle \sum_{i=1}^m \gamma_{i 1} x_i, \sum_{i=1}^m \gamma_{i 1} x_i \right\rangle }  
= k \sqrt{ \sum_{i=1}^m \| x_i \|^2 }  
\leq k \sqrt{m}. 
\end{alignat*}
Hence,
\begin{align*}
\E_{\bm{\gamma}} \sup_{U \in \Uspace} \frac{2}{m} \sum_{i=1}^m \gamma_i \langle w, \varphi_{U S}(x_i) \rangle 
\,\,\,\, \leq \,\,\,\, \frac{4 r \|S\|_{2,2s} k}{\mu_{2s}(S) \sqrt{m}} 
\,\,\,\, \leq \,\,\,\, \frac{4 r k \sqrt{2 s}}{\mu_{2s}(S) \sqrt{m}}, 
\end{align*}
where we used the fact that $\|S\|_{2,2s} \leq \sqrt{2 s}$ (see Lemma \ref{lemma:restricted-operator-norm} in Appendix \ref{sec:infinite-proofs} for a proof). 
\end{Proof}

We present the \emph{pi\`ece de r\'esistance} of this section: 
\begin{Theorem}[Rademacher Average of Mostly Good Random Subclasses]
\label{thm:rademacher-mostly-good}
\begin{align*}
{\Pr}_{\mathbf{z},\bm{\sigma}}
\left\{
\sup_{f \in \F_{\bm{\mu^*},\eta}(\mathbf{x})} \frac{1}{m} \sum_{i=1}^m \sigma_i \loss(y_i,f(x_i)) \geq \frac{t}{4}
\right\} 
\leq 
{m \choose \eta} \left( \frac{8 (r/2)^{1/(k+1)}}{\varepsilon} \right)^{(k+1) k} 
\exp(-m t_3^2 / (2 b^2)) ,
\end{align*}
\begin{flalign*}
\text{for} && 
t_3 := 
  \frac{t}{4} 
  - \frac{L \varepsilon}{\lambda} \left( \frac{3 r \sqrt{s}}{\mu^*_s} + 1 \right)  
  - \frac{2 L \sqrt{\pi} r k \sqrt{s}}{\mu^*_{2s} \sqrt{m}} 
  - \frac{2 b \eta}{m} . &&
\end{flalign*}
\end{Theorem}

\begin{Proof}
As before, each dictionary $D \in \Dspace$ will be factorized as $D = U S$ for $U$ an isometry in $\Uspace$ and $S \in \Sspace = \left(\ball{}{k}\right)^k$. 
Let $\SspaceEps$ be a minimum-cardinality proper $\varepsilon$-cover (in operator norm) of $\{S \in \Sspace: \mu_s(S) \geq \mu^*_s, \mu_{2s}(S) \geq \mu^*_{2s}\}$, the set of suitably incoherent elements of $\Sspace$. 

Recall that the goal is to control the Rademacher complexity of $\F_{\bm{\mu^*},\eta}(\mathbf{x})$. Our strategy will be to control this complexity by controlling the complexity of each subclass from a partition of $\F_{\bm{\mu^*},\eta}(\mathbf{x})$. 
For an arbitrary $f = (D,w) \in \F$, let an index $i$ be \emph{good} if and only if $\smargingt{D, x_i, \tau}$, and let an index be \emph{bad} if and only if it is not good. 
Consider a fixed $m$-sample $\mathbf{z}$ and the occurrence of a set of $m - \eta$ good indices\footnote{Each of the remaining indices can be either good or bad.}. There are $N := {m \choose \eta}$ ways to choose this set of indices. 
We can partition $\F_{\bm{\mu^*},\eta}(\mathbf{x})$ into $N$ subclasses $\F_{\bm{\mu^*},\eta}^1(\mathbf{x}), \ldots, \F_{\bm{\mu^*},\eta}^N(\mathbf{x})$ 
such that for all functions in a given subclass, a particular set of $m - \eta$ indices is guaranteed to be good. To be precise, we can choose distinct good index sets $\Gamma_1, \ldots, \Gamma_{N}$, each of cardinality $m - \eta$, such that 
for each $\Gamma_j$, if $i \in \Gamma_j$ then all $f = (D,w)$ in $\F_{\bm{\mu^*},\eta}^j$ satisfy $\smargingt{D, x_i, \tau}$. 

Since the $\F_{\bm{\mu^*},\eta}^j(\mathbf{x})$ form a partition, we can control the complexity of $\F_{\bm{\mu^*},\eta}(\mathbf{x})$ via:
\begin{align*}
\sup_{f \in \F_{\bm{\mu^*},\eta}(\mathbf{x})} \sum_{i=1}^m \sigma_i \loss(y_i,f(x_i))
= \max_{j \in [N]} \sup_{f \in \F_{\bm{\mu^*},\eta}^j(\mathbf{x})} \sum_{i=1}^m \sigma_i \loss(y_i,f(x_i)) .
\end{align*}

To gain a handle on the complexity of each subclass $\F_{\bm{\mu^*},\eta}^j(\mathbf{x})$, we will approximate the subclasses as follows. 
For each $j \in [N]$, define an $\varepsilon$-neighborhood of $\F_{\bm{\mu^*},\eta}^j(\mathbf{x})$ as
\begin{align*}
\bar{\F}_{\mu^*,\eta}^j(\mathbf{x}) := 
\left\{ 
\begin{array}{llll}
f = (U S', w'): 
& \|S - S'\| \leq \varepsilon, & \|w - w'\| \leq \varepsilon, \\
& S \in \Sspace, & w \in \Wspace, & (U S, w) \in \F_{\bm{\mu^*},\eta}^j(\mathbf{x}) 
\end{array}
\right\} ;
\end{align*}
note that the $\varepsilon$ neighborhood is taken with respect to $S$ and $w$ but not $U$. Also, let $\WspaceEps$ be a minimum-cardinality $\varepsilon$-cover of $\Wspace$ and define an infinite-cardinality epsilon-cover of $\F$: 
\begin{align*}
\F_\varepsilon := \left\{ f = (U S', w') \in \F: U \in \Uspace, S' \in \SspaceEps, w' \in \WspaceEps \right\} .
\end{align*}

Finally, taking the intersection of $\bar{\F}_{\mu^*,\eta}^j(\mathbf{x})$ with $\F_\varepsilon$ yields the $\F_{\bm{\mu^*},\eta}^j(\mathbf{x})$-approximating subclass, a disjoint union of subclasses equal to
\begin{align*}
\bigcup_{S' \in \SspaceEps, w' \in \WspaceEps} \F_{\bm{\mu^*},\eta}^{j,S',w'}(\mathbf{x})
\end{align*}
for
\begin{align*}
\F_{\bm{\mu^*},\eta}^{j,S',w'}(\mathbf{x}) := \F_{\bm{\mu^*},\eta}^j(\mathbf{x}) \cap \left\{ f \in \F: f = (U S', w'): U \in \Uspace \right\} .
\end{align*}

To show that this disjoint union is a good approximator for $\F_{\bm{\mu^*},\eta}^j(\mathbf{x})$, for each $j \in [N]$ and arbitrary $\bm{\sigma} \in \{-1,1\}^m$ we compare
\begin{align*}
\sup_{f \in \F_{\bm{\mu^*},\eta}^j(\mathbf{x})} \frac{1}{m} \sum_{i=1}^m \sigma_i \loss(y_i,f(x_i)) 
\qquad \text{and} \qquad 
\max_{S' \in \SspaceEps, w' \in \WspaceEps} \sup_{f \in \F_{\bm{\mu^*},\eta}^{j,S',w'}(\mathbf{x})} \sum_{i=1}^m \sigma_i \loss(y_i,f(x_i)) .
\end{align*}
Without loss of generality, choose $j=1$ and take $\Gamma_1 = [m - \eta]$. 
If $f$ is in $\F_{\bm{\mu^*},\eta}^1(\mathbf{x})$, it follows that there exists an $f'$ in the disjoint union $\bigcup_{S' \in \SspaceEps, w' \in \WspaceEps} \F_{\bm{\mu^*},\eta}^{1,S',w'}(\mathbf{x})$ such that
\begin{align*}
\frac{1}{m} & \sum_{i=1}^m \sigma_i \left| \loss(y_i, \langle w, \varphi_D(x_i) \rangle) - \loss(y_i, \langle w', \varphi_{D'}(x_i) \rangle) \right| \\
&\leq \frac{L}{m} \left( \sum_{i=1}^{m-\eta} \sigma_i \left| \langle w, \varphi_D(x_i) \rangle - \langle w', \varphi_{D'}(x_i) \rangle \right| \right)
+ \frac{1}{m} \sum_{i=m - \eta + 1}^m \sigma_i \left| \loss(y_i, \langle w, \varphi_D(x_i) \rangle) - \loss(y_i, \langle w', \varphi_{D'}(x_i) \rangle) \right| \\
&\leq \frac{L \varepsilon}{\lambda} \left( \frac{3 r \sqrt{s}}{\mu^*_s} + 1 \right) 
      + \frac{b \eta}{m} ,
\end{align*}
where the last line is due to the Sparse Coding Stability Theorem (Theorem \ref{thm:new-sparse-coding-stability}). 

Therefore, for any $\bm{\sigma} \in \{-1,1\}^m$ it holds that
\begin{align*}
&\sup_{f \in \F_{\bm{\mu^*},\eta}(\mathbf{x})} \frac{1}{m} \sum_{i=1}^m \sigma_i \loss(y_i,f(x_i)) \\
&\leq \max_{j \in [N]} \max_{S' \in \SspaceEps, w' \in \WspaceEps} \sup_{f \in \F_{\bm{\mu^*},\eta}^{j,S',w'}(\mathbf{x})} \frac{1}{m} \sum_{i=1}^m \sigma_i \loss(y_i,f(x_i)) 
      + \frac{L \varepsilon}{\lambda} \left( \frac{3 r \sqrt{s}}{\mu^*_s} + 1 \right) 
      + \frac{b \eta}{m} .
\end{align*}
Thus, the approximation error from using the disjoint union is small (it is $O(\frac{1}{m})$ if $\varepsilon = \frac{1}{m}$).

It remains to control the complexity of the approximating subclass. From the above, for fixed $\mathbf{z}$:
\begin{align*}
&{\Pr}_{\bm{\sigma}}
\left\{
\sup_{f \in \F_{\bm{\mu^*},\eta}(\mathbf{x})} \frac{1}{m} \sum_{i=1}^m \sigma_i \loss(y_i,f(x_i)) \geq \frac{t}{4}
\right\} \\
&\leq {\Pr}_{\bm{\sigma}}
\left\{
\max_{\substack{j \in [N] \\S' \in \SspaceEps, w' \in \WspaceEps}} 
\sup_{f \in \F_{\bm{\mu^*},\eta}^{j,S',w'}(\mathbf{x})} \frac{1}{m} \sum_{i=1}^m \sigma_i \loss(y_i,f(x_i)) 
\geq \frac{t}{4} 
     - \frac{L \varepsilon}{\lambda} \left( \frac{3 r \sqrt{s}}{\mu^*_s} + 1 \right)  
     - \frac{b \eta}{m} 
\right\} \\
&\leq 
N \left( \frac{8 (r/2)^{1/(k+1)}}{\varepsilon} \right)^{(k+1) k} \cdot \\
& \quad \max_{\substack{j \in [N] \\S' \in \SspaceEps, w' \in \WspaceEps}} 
{\Pr}_{\bm{\sigma}}
\left\{
\sup_{f \in \F_{\bm{\mu^*},\eta}^{j,S',w'}(\mathbf{x})} \frac{1}{m} \sum_{i=1}^m \sigma_i \loss(y_i,f(x_i)) 
\geq \frac{t}{4} 
     - \frac{L \varepsilon}{\lambda} \left( \frac{3 r \sqrt{s}}{\mu^*_s} + 1 \right) 
     - \frac{b \eta}{m} 
\right\} .
\end{align*}

Now, from McDiarmid's inequality (Theorem \ref{thm:mcdiarmid}), for any fixed $j \in [N]$, $S' \in \SspaceEps$ and $w' \in \WspaceEps$,
\begin{align*}
{\Pr}_{\bm{\sigma}} \left\{ 
\sup_{f \in \F_{\bm{\mu^*},\eta}^{j,S',w'}(\mathbf{x})} \frac{1}{m} \sum_{i=1}^m  \sigma_i \loss(y_i,f(x_i)) 
> \E_{\bm{\sigma}} \sup_{f \in \F_{\bm{\mu^*},\eta}^{j,S',w'}(\mathbf{x})} \frac{1}{m} \sum_{i=1}^m \sigma_i \loss(y_i,f(x_i)) + t_1 
\right\}
\end{align*}
is at most $\exp(-m t_1^2 / (2 b^2))$.

To make the above useful, let us get a handle on the Rademacher complexity term
\begin{align*}
\E_{\bm{\sigma}} \sup_{f \in \F_{\bm{\mu^*},\eta}^{j,S',w'}(\mathbf{x})} \frac{1}{m} \sum_{i=1}^m \sigma_i \loss(y_i,f(x_i)) .
\end{align*}
Without loss of generality, again take $j = 1$ and $\Gamma_1 = [m - \eta]$. Then
\begin{align*}
&\E_{\bm{\sigma}} 
\sup_{f \in \F_{\bm{\mu^*},\eta}^{j,S',w'}(\mathbf{x})} \frac{1}{m} \sum_{i=1}^m \sigma_i \loss(y_i,f(x_i)) \\
&\leq \E_{\sigma_1, \ldots, \sigma_{m-\eta}} 
\left\{ 
\sup_{f \in \F_{\bm{\mu^*},\eta}^{j,S',w'}(\mathbf{x})} \frac{1}{m} \sum_{i=1}^{m-\eta} \sigma_i \loss(y_i,f(x_i)) 
\right\} \\
& \quad + \E_{\sigma_{m-\eta+1}, \ldots, \sigma_m} 
\left\{ 
\sup_{f \in \F_{\bm{\mu^*},\eta}^{j,S',w'}(\mathbf{x})} \frac{1}{m} \sum_{i=m - \eta + 1}^{m} \sigma_i \loss(y_i,f(x_i)) 
\right\} 
\\
&\leq \E_{\sigma_1, \ldots, \sigma_{m-\eta}} \left\{ \sup_{f \in \F_{\bm{\mu^*},\eta}^{j,S',w'}(\mathbf{x})} \frac{1}{m} \sum_{i=1}^{m-\eta} \sigma_i \loss(y_i,f(x_i)) \right\}
+ \frac{b \eta}{m} . 
\end{align*}

Now, Theorem \ref{thm:gaussian-average-fixed-S-w}, the Rademacher Loss Comparison Lemma (Lemma \ref{lemma:rademacher-loss}), and the Rademacher-Gaussian Average Comparison Lemma (Lemma \ref{lemma:rademacher-gaussian-bound}) imply that
\begin{align*}
\E_{\bm{\sigma}} \sup_{\{U \in \U: \sparsephi{s, U S,\mathbf{x}}\}} 
\frac{1}{m} \sum_{i=1}^{m-\eta} \sigma_i \loss \bigl( y_i,\langle w, \varphi_{U S}(x_i) \rangle \bigr) 
&\leq \frac{\sqrt{m-\eta}}{m} \frac{2 L \sqrt{\pi} r k \sqrt{s}}{\mu_{2s}(S)} \\
&\leq \frac{2 L \sqrt{\pi} r k \sqrt{s}}{\mu_{2s}(S) \sqrt{m}} ,
\end{align*}
and hence
\[
{\Pr}_{\bm{\sigma}} \left\{ 
\sup_{f \in \F_{\bm{\mu^*},\eta}^{j,S',w'}(\mathbf{x})} \frac{1}{m} \sum_{i=1}^m  \sigma_i \loss(f(x_i)) 
> \frac{2 L \sqrt{\pi} r k \sqrt{s}}{\mu^*_{2s} \sqrt{m}} + \frac{b \eta}{m} + t_1 
\right\} 
\leq \exp(-m t_1^2 / (2 b^2)) .
\]

Combining this bound with the fact that the bound is independent of the draw of $\mathbf{z}$ and applying Proposition \ref{prop:hypothesis-covering-cardinality} (with $d$ set to $k$) to extend the bound over all choices of $j$, $S'$, and $w'$ yields the final result.
\end{Proof}

For the case of $\eta = 0$, let
\begin{align*}
t_2 := 
  \frac{t}{4} 
  - \frac{L \varepsilon}{\lambda} \left( \frac{3 r \sqrt{s}}{\mu^*_s} + 1 \right)  
  - \frac{2 L \sqrt{\pi} r k \sqrt{s}}{\mu^*_{2s} \sqrt{m}} .
\end{align*}

It is now possible to prove the generalization error bound for the infinite-dimensional setting.

\begin{Proof}[of Theorem \ref{thm:final-learning-bound-infinite}] 
Since $\F_{\mu_*}(\mathbf{x})$ is equivalent to $\F_{\bm{\mu^*},0}(\mathbf{x})$, Lemma \ref{lemma:symmetrization-random-signs} and Theorem \ref{thm:rademacher-mostly-good} imply that
\begin{align*}
&{\Pr}_{\mathbf{z \, z'}} \left\{
\begin{array}{l}
\exists f \in \F_{\bm{\mu^*}} ,\,\, 
\smargingt{D, \mathbf{x}, \tau} 
\booland 
\left( \exists \, \mathbf{\tilde{x}} \subseteq_\eta \mathbf{x'} \,\, \smargingt{D, \mathbf{\tilde{x}}, \tau} \right) \\
\qquad \booland 
\left( (\ProbZp - \ProbZ) \lossof{f} \geq \frac{t}{2} \right)
\end{array}
\right\} \\
&\leq \left( \frac{8 (r/2)^{1/(k+1)}}{\varepsilon} \right)^{(k+1) k}
\left( 
\exp(-m t_2^2 / (2 b^2)) 
+ {m \choose \eta} \exp(-m t_3^2 / (2 b^2)) \right) \\
&\leq 2 {m \choose \eta} \left( \frac{8 (r/2)^{1/(k+1)}}{\varepsilon} \right)^{(k+1) k} 
\exp(-m t_3^2 / (2 b^2)) ,
\end{align*}
and consequently the full probability \eqref{eqn:full-probability} in
Proposition \ref{prop:symmetrization-ghost-infinite} can be upper bounded (using $\eta = \log \frac{1}{\delta}$) as:
\begin{align*}
&{\Pr}_{\mathbf{z \, x''}} \left\{
\begin{array}{l}
\fhatz \in \F_{\bm{\mu^*}} ,\,\,
\smargingt{\Dhatz, (\mathbf{x} \cup \mathbf{x''}), \tau} 
\booland
\left( (\Prob - \ProbZ) \lossof{\fhatz} \geq t \right) 
\end{array}
\right\} \\
&\leq 4 {m \choose \log \frac{1}{\delta}} \left( \frac{8 (r/2)^{1/(k+1)}}{\varepsilon} \right)^{(k+1) k} \exp(-m t_3^2 / (2 b^2)) + 2 \delta . 
\end{align*}

After some elementary manipulations and choosing $\varepsilon = \frac{1}{m}$, we nearly have the final learning bound. Let $\mu^*_s, \mu^*_{2s} > 0$, $s \in [k]$, and $m \geq \frac{43}{\tau^2 \lambda}$ be fixed \emph{a priori}. 
With probability at least $1 - \delta$ over a labeled $m$-sample $\mathbf{z} \sim \Prob^m$ and a second, unlabeled $m$-sample $\mathbf{x''} \sim \Pi^m$, if an algorithm learns hypothesis $\fhatz = (\Dhatz, \whatz)$ from $\mathbf{z}$ such that $\mu_{2 s}(\Dhatz) \geq \mu^*_{2s}$, $\mu_s(\Dhatz) \geq \mu^*_s$, $\sparsephi{s, \Dhatz, \mathbf{x} \cup \mathbf{x''}}$, and $\smargingt{\Dhatz, \mathbf{x} \cup \mathbf{x''}, \tau}$ all hold, then the generalization error $(\Prob - \ProbZ) \lossof{\fhatz}$ is bounded by:
\begin{align*}
& \frac{8 L \sqrt{\pi} r k \sqrt{s}}{\mu^*_{2s} \sqrt{m}}  
+ b \sqrt{\frac{8 \left( (k+1) k \log(8m) + k \log \frac{r}{2} + (\log m + 1) \log \frac{4}{\delta} + \log 2 \right)}{m}} \\
& \quad + \frac{1}{m} \left( \frac{4 L}{\lambda} \left( \frac{3 r \sqrt{s}}{\mu^*_s} + 1 \right) 
+ 8 b \log \frac{4}{\delta} \right) .
\end{align*}
Making this bound made adaptive to the incoherences, sparsity level, and margin on $\mathbf{z}$ and $\mathbf{x''}$ yields the following final result. 
With probability at least $1 - \delta$ over a labeled $m$-sample $\mathbf{z} \sim \Prob^m$ and a second, unlabeled sample $\mathbf{x''} \sim \Pi^m$, if an algorithm learns hypothesis $\fhatz = (\Dhatz, \whatz)$ such that $\varphi_{\Dhatz}$ is $s$-sparse on $(\mathbf{x} \cup \mathbf{x''})$, $\mu_{2s}(\fhatz) > 0$, and
\begin{align*}
m \geq \frac{43}{\smargin^2(\Dhatz, \mathbf{x} \cup \mathbf{x''}) \cdot \lambda}, 
\end{align*}
then the generalization error $(\Prob - \ProbZ) \lossof{\fhatz}$ is 
bounded by:
\begin{align*}
\frac{16 L \sqrt{\pi} r k \sqrt{s}}{\mu_{2s}(\fhatz) \sqrt{m}} 
&+ b \sqrt{\frac{8 \left( (k^2+k) \log(8m) + k \log \frac{r}{2} 
                          + (\log m + 1) \log \frac{7 \alpha k}{\delta} + \log 2
                   \right)}{m}}\\
&+ \frac{1}{m} \left( \frac{4 L}{\lambda} \left( \frac{6 r \sqrt{s}}{\mu_s(\fhatz)} + 1 \right)
                      + 8 b \log \frac{7 \alpha k}{\delta} \right) ,
\end{align*}
for $\alpha = \left(\log_2\bigl(\frac{4}{\mu_s(\fhatz)}\bigr) \log_2\bigl(\frac{4}{\mu_{2s}(\fhatz)}\bigr) \right)^2$.
\end{Proof}

\begin{figure}[!t]
\includegraphics[width=80mm]{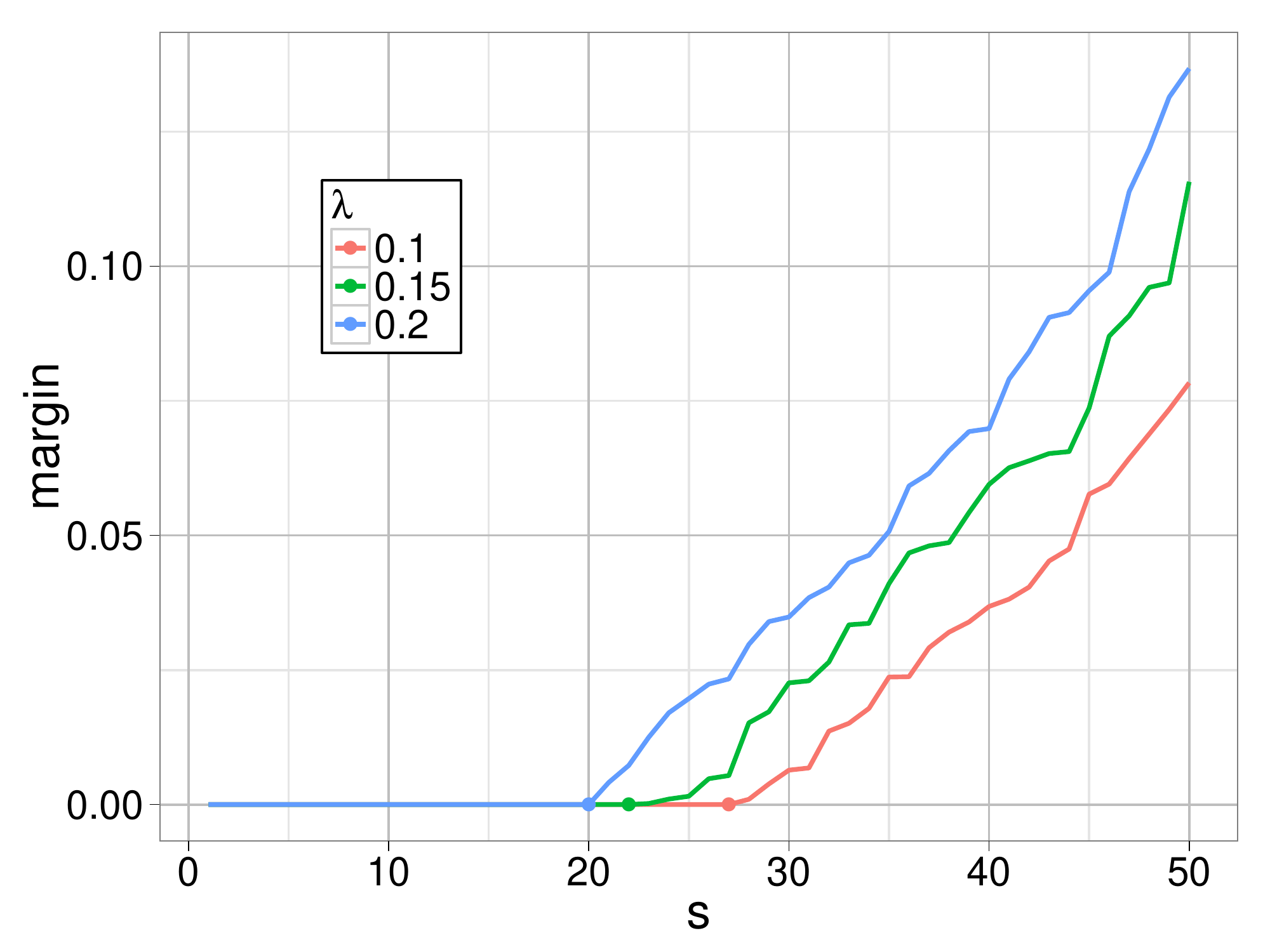}
\includegraphics[width=80mm]{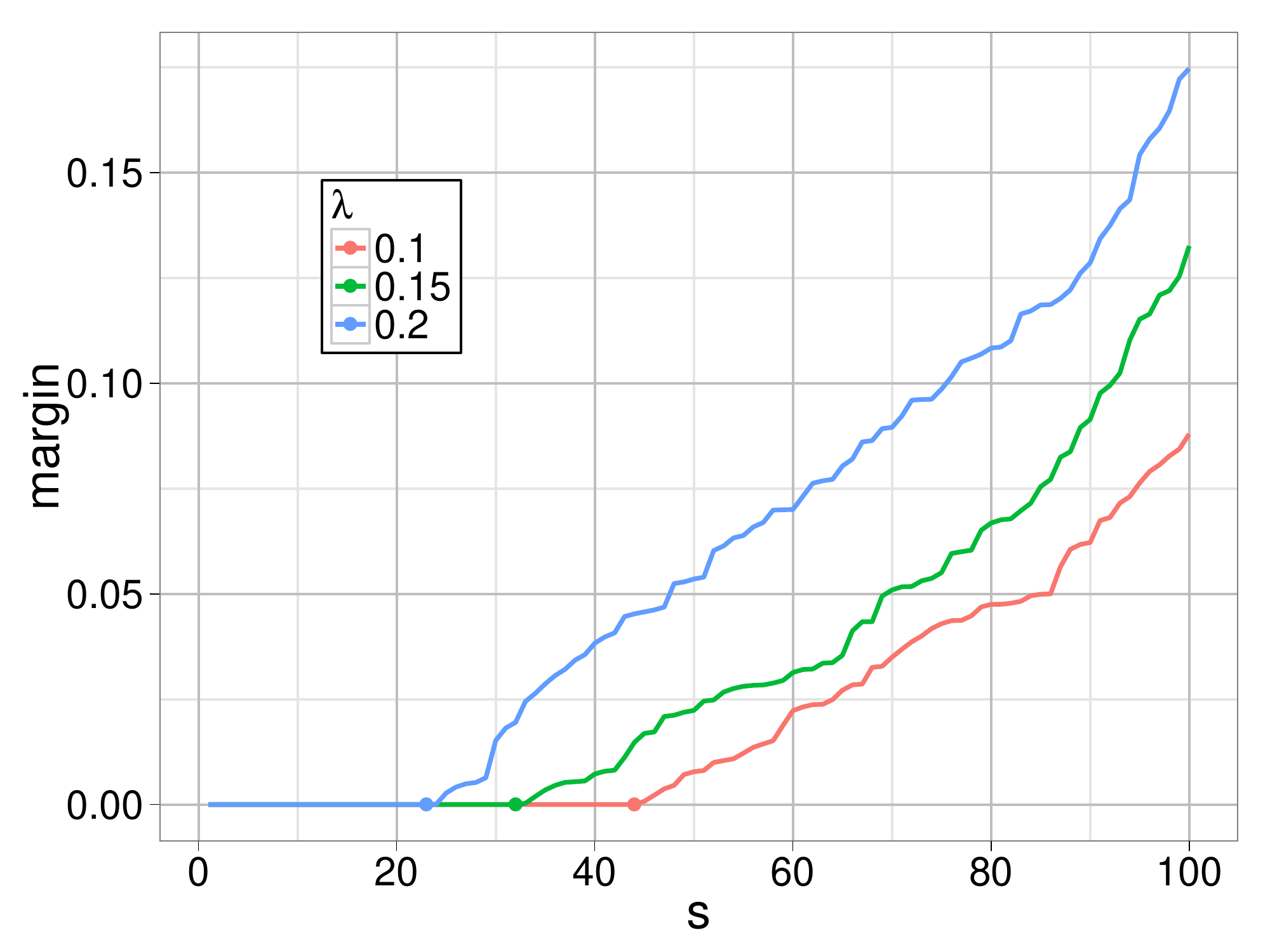}
\includegraphics[width=80mm]{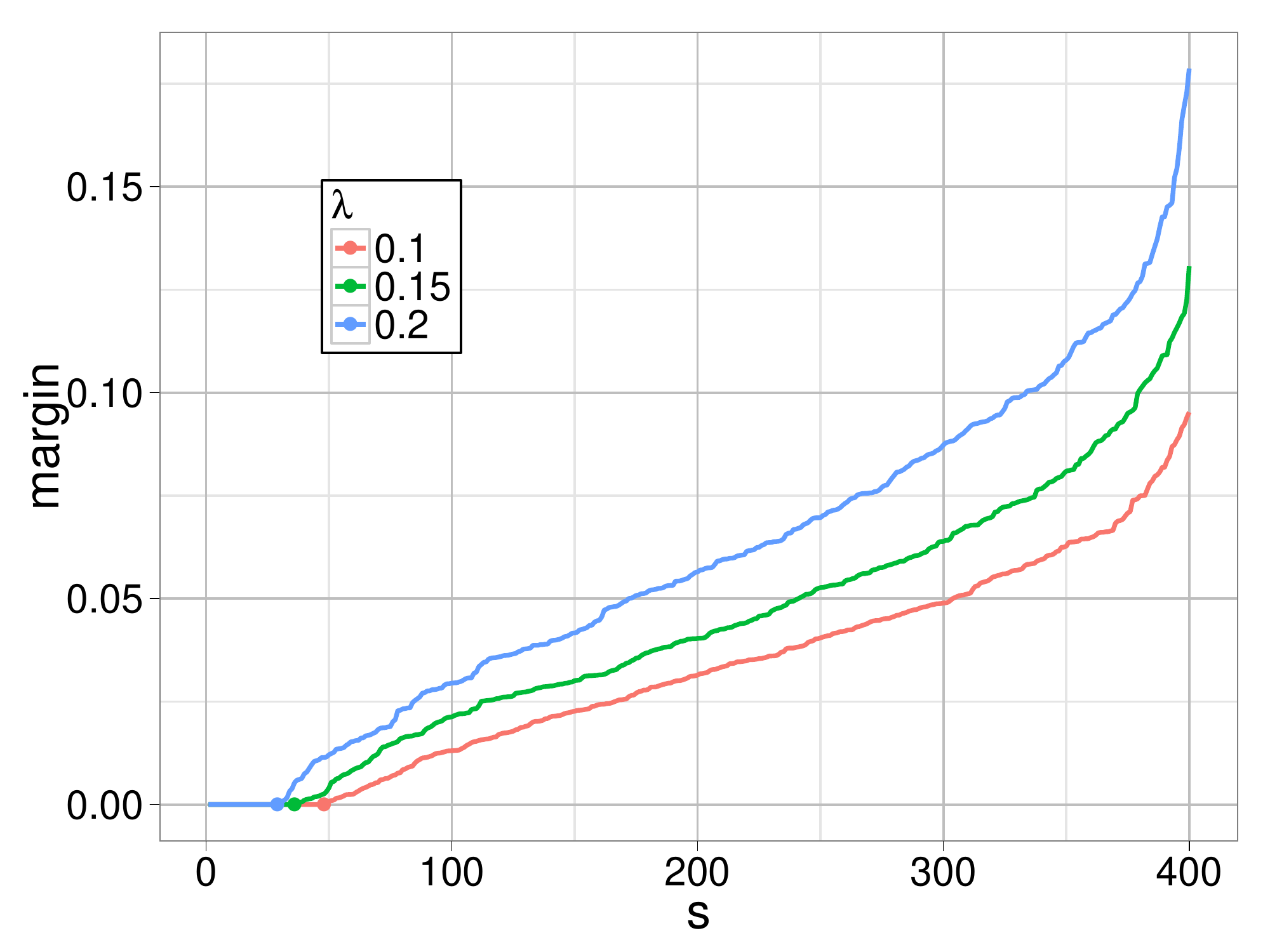}
\includegraphics[width=80mm]{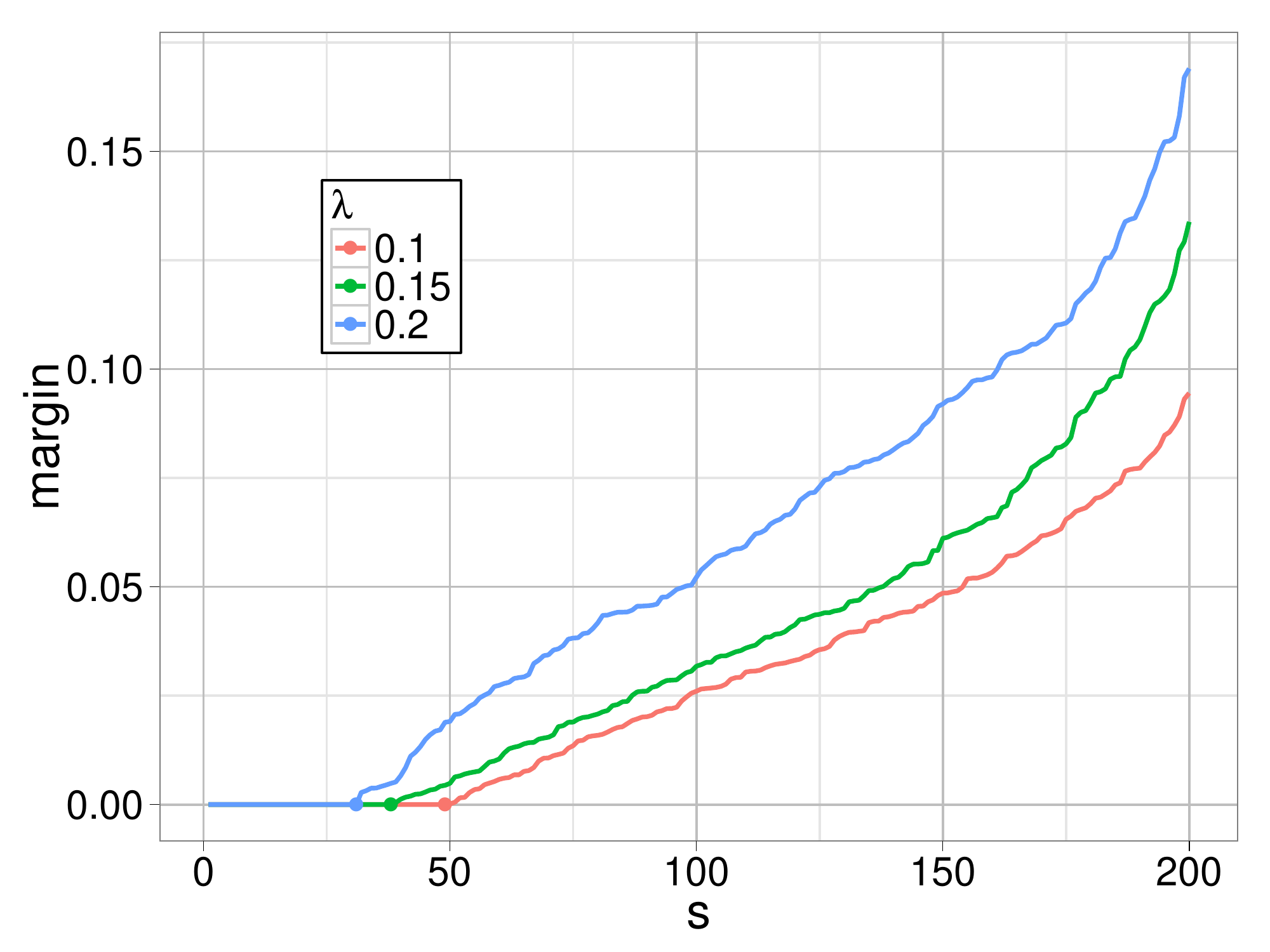}
\caption{\label{fig:usps} The $s$-margin for predictive sparse coding trained on the USPS training set, digit 4 versus all, for three settings of $\lambda$. Clockwise from top left: 50 atoms, 100 atoms, 200 atoms, and 400 atoms. The sparsity level (maximum number of non-zeros per code, taken across all codes of the training points) is indicated by the dots.}
\end{figure}

\section{An empirical study of the $s$-margin}
\label{sec:empirical}

\begin{figure}[!t]
\includegraphics[width=80mm]{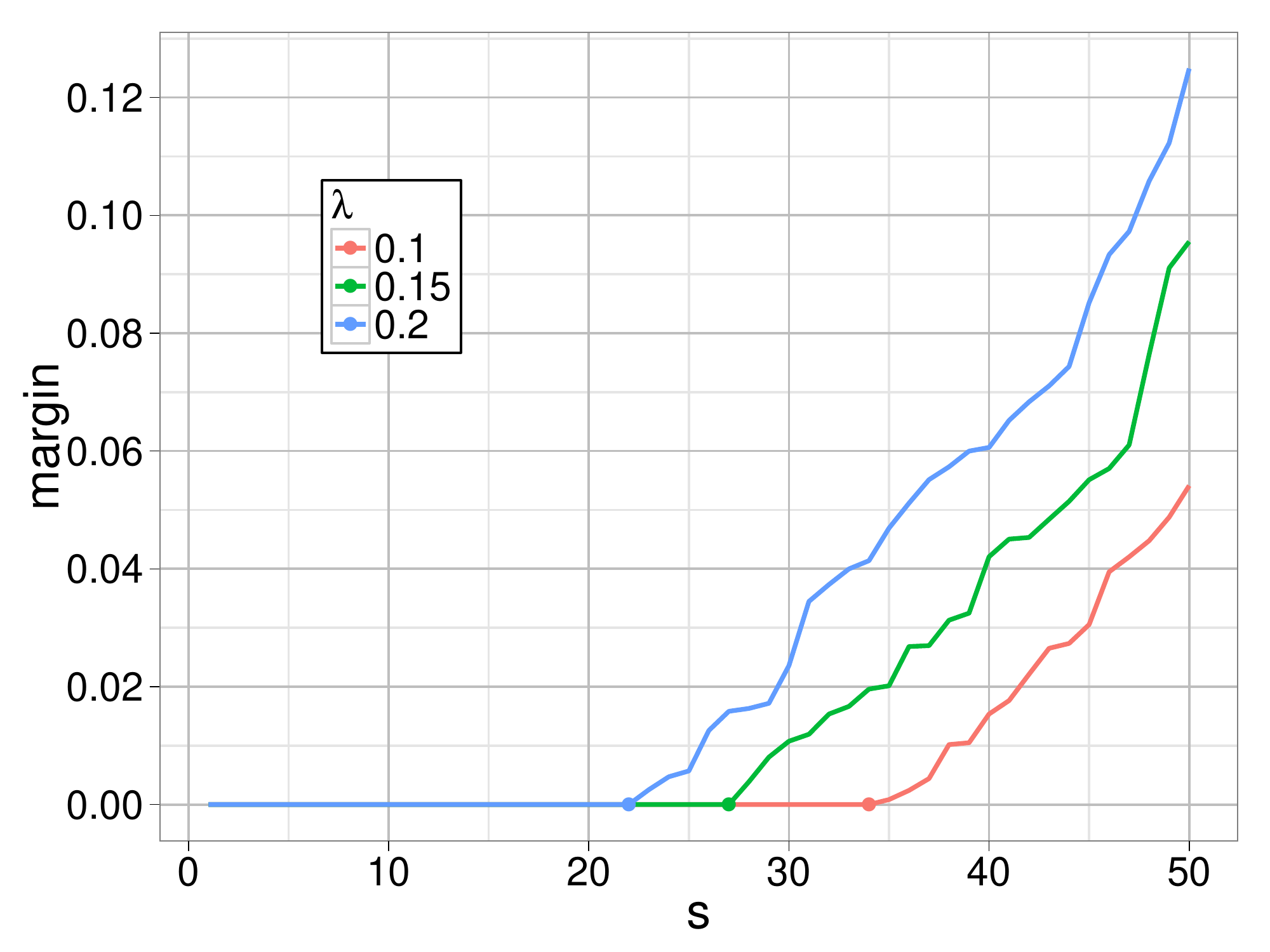}
\includegraphics[width=80mm]{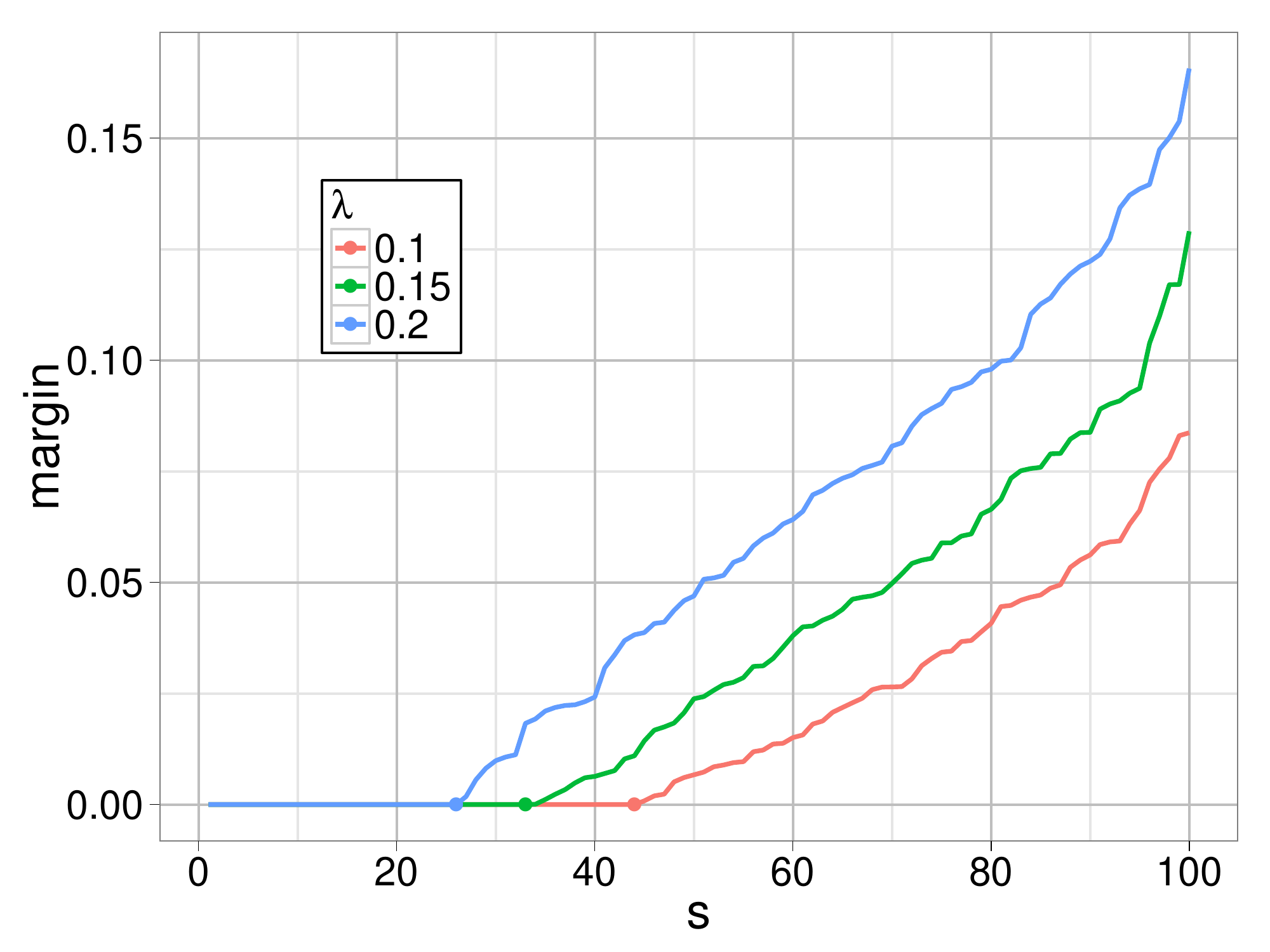}
\includegraphics[width=80mm]{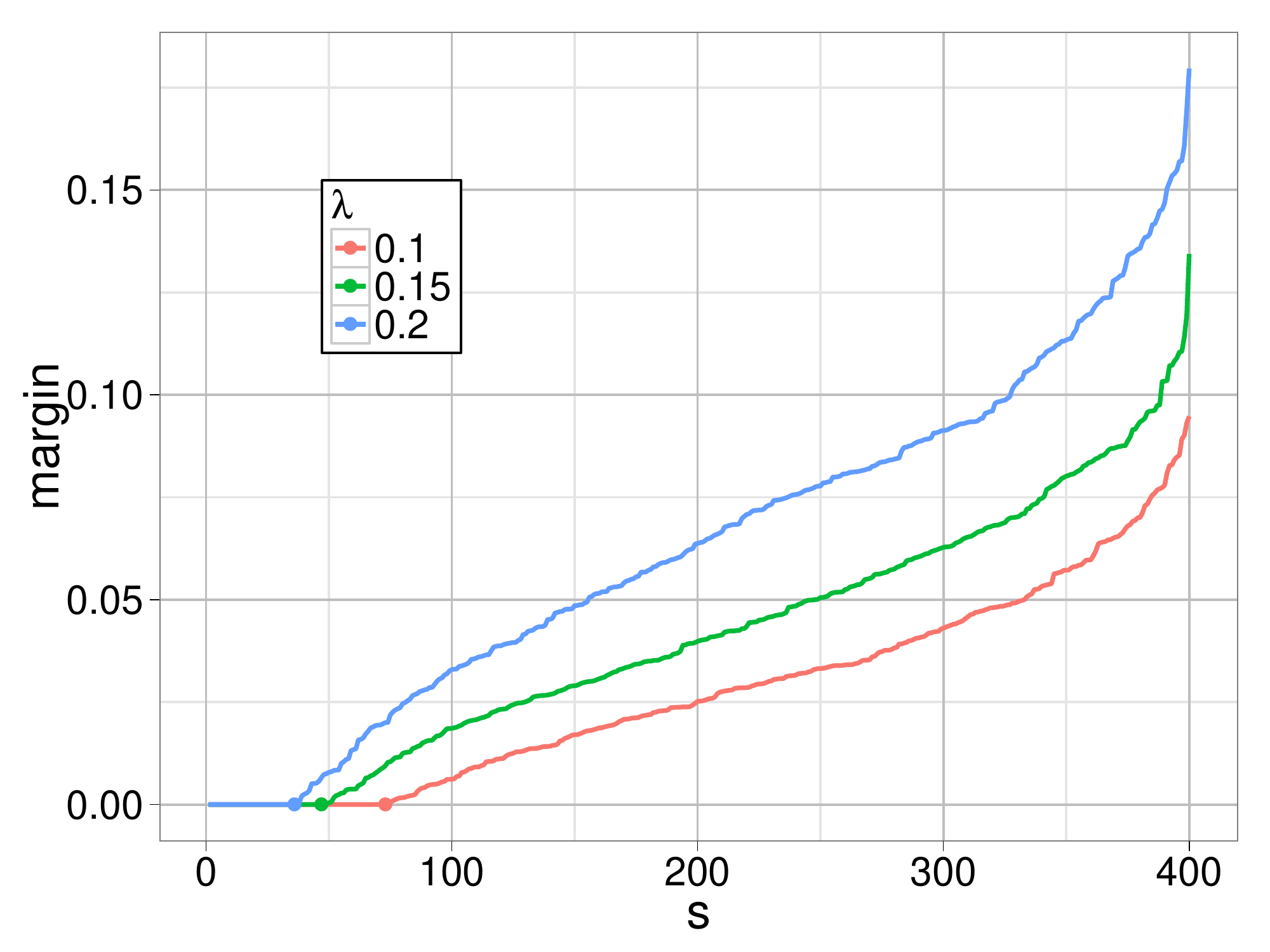}
\includegraphics[width=80mm]{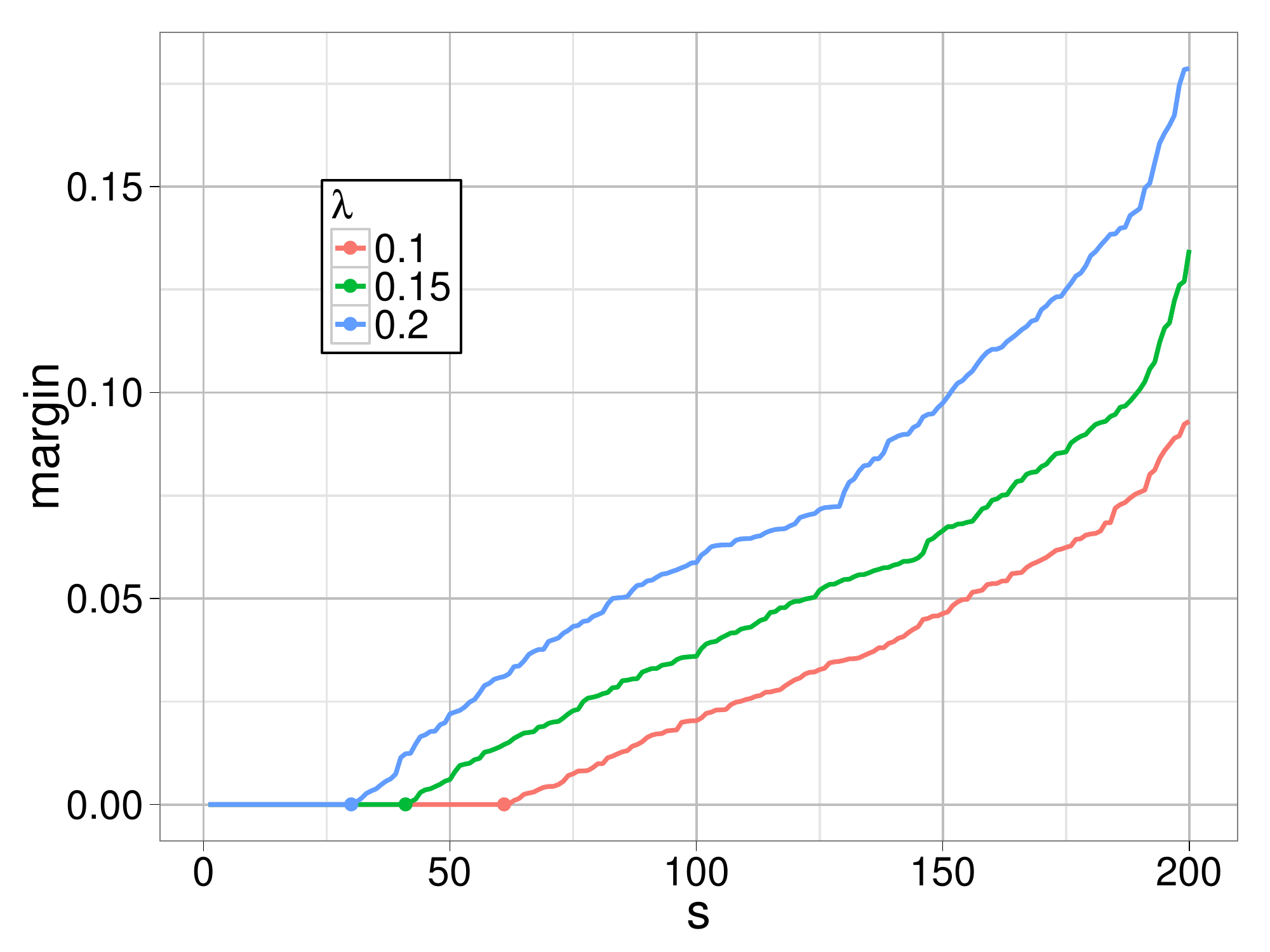}
\caption{\label{fig:mnist} The $s$-margin for predictive sparse coding trained on the MNIST training set, digit 4 versus all, for three settings of $\lambda$. Clockwise from top left: 50 atoms, 100 atoms, 200 atoms, and 400 atoms. The sparsity level (maximum number of non-zeros per code, taken across all codes of the training points) is indicated by the dots.} 
\end{figure}

We now establish some empirical evidence that the $s$-margin is well away from zero even when $s$ is only slightly larger than the observed sparsity level. We performed experiments on two separate digit classification tasks, from the USPS dataset and the MNIST dataset \cite{lecun1998gradient}. In both cases, we employed the single binary classification task of the digit 4 versus all the other digits, and for both datasets all the training data was used. Predictive sparse coding was trained as per the stochastic gradient descent approach of \cite{mairal2012task}.

The results for USPS and MNIST are shown in Figures \ref{fig:usps} and \ref{fig:mnist} respectively. Each data point (an image) was normalized to unit norm. In all plots, it is apparent that when the minimum sparsity level is $s$ (indicated by the colored dots on the x-axis of the plots), there is a non-trivial $(s+\rho)$-margin for $\rho$ a small positive integer. Using the $2s$-margin when $s$-sparsity holds may ensure that there is a moderate margin for only a constant factor increase to $s$.

\section{Discussion and open problems}
\label{sec:discussion}

We have shown the first generalization error bounds for predictive sparse coding. The learning bounds in Theorems \ref{thm:final-learning-bound-overcomplete} and \ref{thm:final-learning-bound-infinite} are intimately related to the stability of the sparse encoder, and consequently the bounds depend on properties that depend both on the learned dictionary and the training sample. 
Using the techniques of this work, in the infinite-dimensional setting it is unclear whether one can achieve the encoder stability guarantees without measuring properties of the encoder on an independent, unlabeled sample. It is an important open problem whether there is a generalization error bound for the infinite-dimensional setting which does not rely on the second sample. Additionally, the PRP condition in the Sparse Coding Stability Theorem (Theorem \ref{thm:new-sparse-coding-stability}) appears to be much stronger than what should be required. We conjecture that the PRP should actually be $O(\varepsilon)$ rather than $O(\sqrt{\varepsilon})$. If this conjecture turns out to be true, then the number of samples required before Theorems \ref{thm:final-learning-bound-overcomplete} and \ref{thm:final-learning-bound-infinite} kick in would be greatly reduced, as would be the size of many of the constants in the results. 

While this work establishes upper bounds on the generalization error for predictive sparse coding, two things remain unclear. How close are these bounds to the optimal ones? Also, what lower bounds can be established in each of the settings? If the conditions on which these bounds rely are of fundamental importance, then the presented data-dependent bounds provide motivation for an algorithm to prefer dictionaries for which small subdictionaries are well-conditioned and to additionally encourage large coding margin on the training sample.


\newpage

\appendix
\section{Proof of Sparse Coding Stability Theorem}
\label{sec:proof-sparse-coding-stability-theorem}

\begin{figure}[h]
\fbox{
\includegraphics[width=\linewidth, clip=true, trim = 0mm 0mm 0mm 25mm]{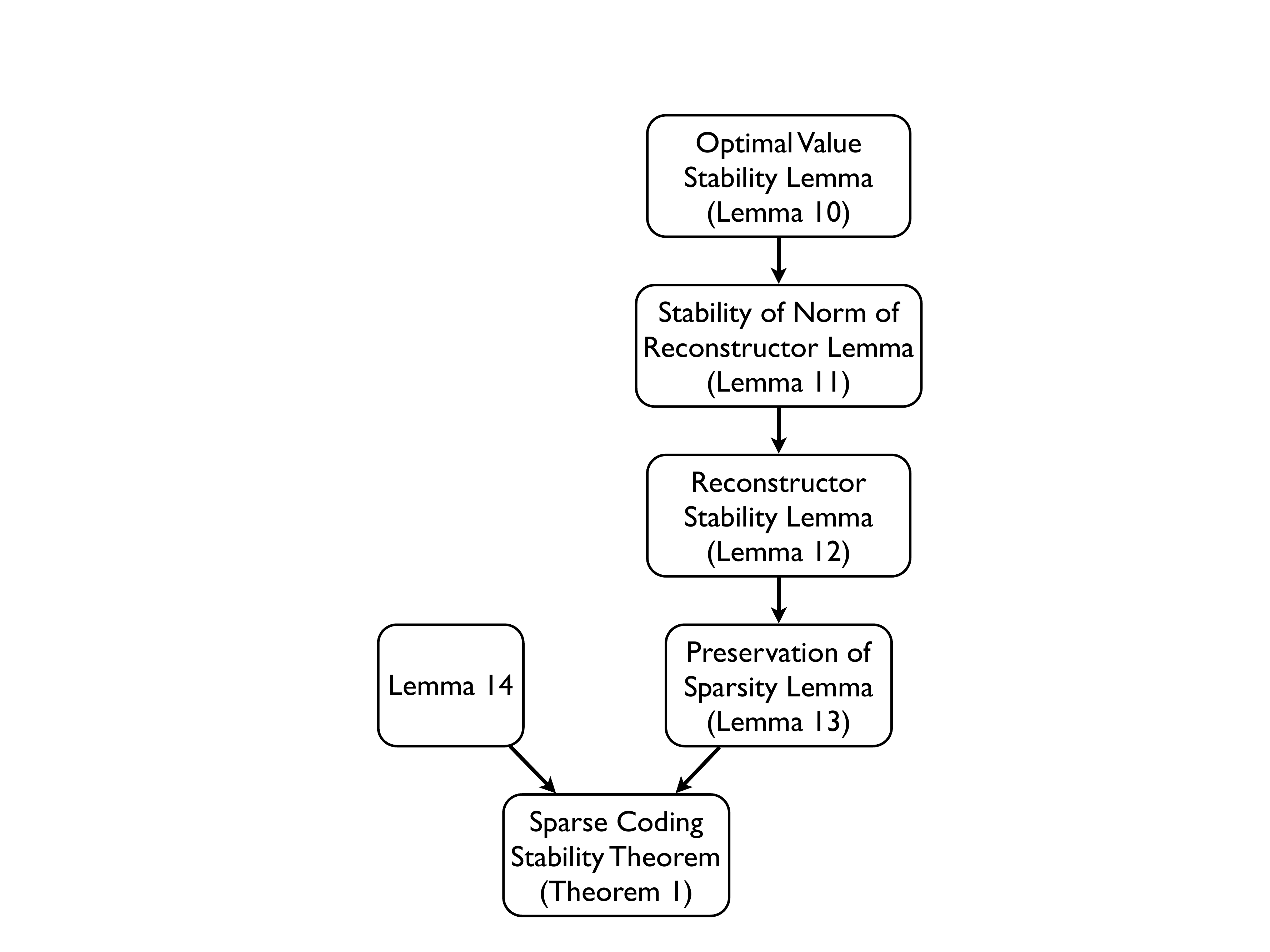}
}
\caption{
\label{fig:sparse-proof-flowchart}
Proof flowchart for the Sparse Coding Stability Theorem (Theorem \ref{thm:new-sparse-coding-stability}).
}
\end{figure}

The flow of this section is as follows. We first establish some preliminary notation and summarize important conditions. Several lemmas are then presented to support a key sparsity lemma. This sparsity lemma establishes that the solution to the perturbed problem is sparse provided the perturbation is not too large. Finally, the sparsity of this new solution is exploited to bound the difference of the new solution from the old solution. This flow is embodied by the proof flowchart in Figure \ref{fig:sparse-proof-flowchart}.

\subsection{Notation and assumptions}
Let $\alpha$ and $\tilde{\alpha}$ respectively denote the solutions to the LASSO problems:
\begin{align*}
\alpha = \argmin_z \frac{1}{2} \|x - D z\|_2^2 + \lambda \|z\|_1
&&
\tilde{\alpha} = \argmin_z \frac{1}{2} \|x - \tilde{D} z\|_2^2 + \lambda \|z\|_1 .
\end{align*}

First, let's review the optimality conditions for the LASSO \cite[conditions L1 and L2]{asif2010lasso}:
\begin{align*}
\langle D_j, x - D \alpha \rangle &= \sign(\alpha_j) \lambda \quad \text{if } \alpha_j \neq 0, \\
\left| \langle D_j, x - D \alpha \rangle \right| &< \lambda \quad \text{otherwise.}
\end{align*}

Note that the above optimality conditions imply that if $\alpha_j \neq 0$ then
\begin{align*}
\left| \langle D_j, x - D \alpha \rangle \right| = \lambda .
\end{align*}

\subsection*{Assumptions}
The statement of the Sparse Coding Stability Theorem (Theorem \ref{thm:new-sparse-coding-stability}) makes the following assumptions:
\paragraph{(A1) - Closeness}
$D$ and $\tilde{D}$ are close, as measured by operator norm:
\begin{align*}
\|\tilde{D} - D\|_2 \leq \varepsilon .
\end{align*}

\paragraph{(A2) - Incoherence}
There is a $\mu > 0$ such that, for all $J \subseteq [k]$ satisfying $|J| = s$:
\begin{align*}
\sigma_{\min}(D_J) \geq \mu .
\end{align*}

\paragraph{(A3) - Sparsity with margin}
For some fixed $\tau > 0$, there is a $\mathcal{I} \subseteq [k]$ with $|\mathcal{I}| = k - s$ such that for all $i \in \mathcal{I}$:
\begin{align*}
\left| \langle D_i, x - D \alpha \rangle \right| < \lambda - \tau .
\end{align*}
Consequently, all $i \in \mathcal{I}$ satisfy $\alpha_i = 0$.

\subsection{Useful observations}

Let $v_D^*$ be the optimal value of the LASSO for dictionary $D$:
\begin{align*}
v_D^* &= \min_z \frac{1}{2} \|x - D z\|_2^2 + \lambda \|z\|_1 \\
&= \frac{1}{2} \|x - D \alpha\|_2^2 + \lambda \|\alpha\|_1
\end{align*}
Likewise, let
\begin{align*}
v_{\tilde{D}}^* &= \frac{1}{2} \|x - \tilde{D} \tilde{\alpha}\|_2^2 + \lambda \|\tilde{\alpha}\|_1
\end{align*}

The first observation is that the values of the optimal solutions are close:
\begin{Lemma}[Optimal Value Stability]
\label{lemma:optival-values-close}
If $\|D - \tilde{D}\|_2 \leq \varepsilon$, then 
\begin{align*}
\left| v_D^* - v_{\tilde{D}}^* \right| \leq \frac{\varepsilon}{\lambda} .
\end{align*}
\end{Lemma}
\begin{Proof}
The proof is simple:
\begin{align*}
v_{\tilde{D}}^* 
&\leq \frac{1}{2} \|x - \tilde{D} \alpha\|_2^2 + \lambda \|\alpha\|_1 \\
&= \frac{1}{2} \|x - D \alpha + (D - \tilde{D}) \alpha\|_2^2 + \lambda \|\alpha\|_1 \\
&\leq 
\frac{1}{2} \left( 
\|x - D \alpha\|_2^2 
+ \|x - D \alpha\|_2 \|(D - \tilde{D}) \alpha\|_2 
+ \|(D - \tilde{D}) \alpha\|_2^2
\right)
+ \lambda \|\alpha\|_1  \\
&\leq \frac{1}{2} \|x - D \alpha\|_2^2 + \lambda \|\alpha\|_1
+ \frac{1}{2} \left(
\frac{\varepsilon}{\lambda}
+ \left( \frac{\varepsilon}{\lambda} \right)^2
\right) \\
&\leq v_D^* + \frac{\varepsilon}{\lambda}
\end{align*}
for $\frac{\varepsilon}{\lambda} \leq 1$. A symmetric argument shows that $v_D^* \leq v_{\tilde{D}}^* + \frac{\varepsilon}{\lambda}$. 
\end{Proof}

The second observation shows that the norms of the optimal reconstructors are close. 
\begin{Lemma}[Stability of Norm of Reconstructor]
\label{lemma:stability-norm-reconstructor}
If $\|D - \tilde{D}\|_2 \leq \varepsilon$, then
\begin{align*}
\left| \|D \alpha\|_2^2 -\|\tilde{D} \tilde{\alpha}\|_2^2 \right| 
\leq \frac{2 \varepsilon}{\lambda} .
\end{align*}
\end{Lemma}
Showing this is more involved than the previous observation. 
\begin{Proof}
First, we claim (and show) that
\begin{align}
(x - D \alpha)^T D \alpha = \lambda \|\alpha\|_1 .
\label{eqn:osborne}
\end{align}
The proof of the claim comes directly from \citet[circa (2.8)]{osborne2000lasso}
To see \eqref{eqn:osborne}, consider the equivalent dual problem for the LASSO (for appropriate choice of $t$):
\begin{align*}
\begin{aligned}
& \underset{z}{\text{minimize}}
& & \frac{1}{2} \|x - D z\|_2^2 \\
& \text{subject to}
& & \|z\|_1 \leq t .
\end{aligned}
\end{align*}

The Lagrangian is
\begin{align*}
\mathcal{L}(z, \lambda) = \frac{1}{2} \|x - D z\|_2^2  - \lambda (t - \|z\|_1) ,
\end{align*}
and the subgradient with respect to $z$ is
\begin{align*}
\partial_z \mathcal{L}(z, \lambda) = -D^T (x - D z) + \lambda v ,
\end{align*}
where $v_j = 1$ if $z_j > 0$, $v_j = -1$ if $z_j < 0$, and $v_j \in [-1,1]$ if $z_j = 0$. From the definition of $v$, it follows that
\begin{align*}
v^T z = \|z\|_1 .
\end{align*}

At an optimal point $\alpha$, $\partial_z \mathcal{L}(\alpha, \lambda) = 0$, and hence
\begin{align*}
D^T (x - D \alpha) &= \lambda v \\
&\Updownarrow \\
(x - D \alpha)^T D  &= \lambda v^T \\
&\Downarrow \\
(x - D \alpha)^T D \alpha  &= \lambda v^T \alpha \\
&\Updownarrow \\
(x - D \alpha)^T D \alpha  &= \lambda \|\alpha\|_1 ,
\end{align*}
as claimed. 

Now, we use the fact that the values of the optimal solutions are close (Lemma \ref{lemma:optival-values-close}):
\begin{align*}
\left| v_D^* - v_{\tilde{D}}^* \right| \leq \frac{\varepsilon}{\lambda} .
\end{align*}

But $v_D^*$ is just
\begin{align*}
\frac{1}{2} \langle x - D \alpha, x - D \alpha \rangle + \lambda \|\alpha\|_1 
&= \frac{1}{2} \langle x - D \alpha, x - D \alpha \rangle + \langle x - D \alpha, D \alpha \rangle \\
&= \frac{1}{2} \langle x, x - D \alpha \rangle - \frac{1}{2} \langle x - D \alpha, D \alpha \rangle + \langle x - D \alpha, D \alpha \rangle \\
&= \frac{1}{2} \left( \langle x, x - D \alpha \rangle + \langle x - D \alpha, D \alpha \rangle \right) \\
&= \frac{1}{2} \langle x + D \alpha, x - D \alpha \rangle \\
&= \frac{1}{2} \left( \|x\|_2^2 - \|D \alpha\|_2^2 \right) .
\end{align*}

Consequently,
\begin{align*}
\left|
  \frac{1}{2} \left( \|x\|_2^2 - \|D \alpha\|_2^2 \right) 
  - \frac{1}{2} \left( \|x\|_2^2 - \|\tilde{D} \tilde{\alpha}\|_2^2 \right) 
\right| 
\leq \frac{\varepsilon}{\lambda}
\end{align*}
and hence
\begin{align*}
\left| \|D \alpha\|_2^2 -\|\tilde{D} \tilde{\alpha}\|_2^2 \right| 
\leq \frac{2 \varepsilon}{\lambda} .
\end{align*}
\end{Proof}

Finally, we prove stability of the optimal reconstructor. Rather than showing that $\|D \alpha - \tilde{D} \tilde{\alpha}\|_2^2$ is $O(\varepsilon)$, it will be more convenient for later purposes to prove the following roughly equivalent result.
\begin{Lemma}[Reconstructor Stability]
\label{lemma:reconstructor-stability-proof}
If $\|D - \tilde{D}\|_2 \leq \varepsilon$, then 
\begin{align*}
\|D \alpha - D \tilde{\alpha}\|_2^2 \leq \frac{26 \varepsilon}{\lambda} .
\end{align*}
\end{Lemma}
\begin{Proof}
Let $\alpha' := \frac{1}{2} (\alpha + \tilde{\alpha})$. From the optimality of $\alpha$, it follows that $v_D(\alpha) \leq v_D(\alpha')$, or more explicitly:
\begin{align}
\frac{1}{2} \|x - D \alpha\|_2^2 + \lambda \|\alpha\|_1 
\leq \frac{1}{2} \|x - D \alpha'\|_2^2 + \lambda \|\alpha'\|_1 .
\label{eqn:lower-bound}
\end{align}

First, note that 
$\left| \|D \tilde{\alpha}\|_2^2 - \|\tilde{D} \tilde{\alpha}\|_2^2  \right| \leq \frac{5 \varepsilon}{\lambda}$, because
\begin{align*}
\left| \|D \tilde{\alpha}\|_2^2 - \|\tilde{D} \tilde{\alpha}\|_2^2  \right|
&\leq 2 \left| \langle D \tilde{\alpha}, (\tilde{D} - D) \alpha \rangle \right|
      + \|(\tilde{D} - D) \tilde{\alpha}\|_2^2 \\
&\leq 2 \|D \tilde{\alpha}\|_2 \|\tilde{D} - D\|_2 \|\tilde{\alpha}\|_2 + \left(\|\tilde{D} - D\|_2 \|\tilde{\alpha}\|_2 \right)^2 \\
&\leq 2 \left( 1 + \frac{\varepsilon}{\lambda} \right) \frac{\varepsilon}{\lambda} + \left( \frac{\varepsilon}{\lambda} \right)^2 \\
&\leq \frac{5 \varepsilon}{\lambda},
\end{align*}
assuming $\varepsilon \leq \lambda$. 
Combining this fact with Lemma \ref{lemma:stability-norm-reconstructor}, $\left| \|D \alpha\|_2^2 -\|\tilde{D} \tilde{\alpha}\|_2^2 \right| \leq \frac{2 \varepsilon}{\lambda}$, yields
\begin{align*}
\left| \|D \alpha\|_2^2 - \|D \tilde{\alpha}\|_2^2  \right| \leq \frac{7 \varepsilon}{\lambda} .
\end{align*}

By the convexity of the 1-norm, the RHS of \eqref{eqn:lower-bound} obeys:
\begin{align*}
&\frac{1}{2} \left\| x - D \left( \frac{\alpha + \tilde{\alpha}}{2} \right) \right\|_2^2 
  + \lambda \left\| \frac{\alpha + \tilde{\alpha}}{2} \right\|_1 \\
&\leq \frac{1}{2} \left\|x - \frac{1}{2} (D \alpha + D \tilde{\alpha}) \right\|_2^2
+ \frac{\lambda}{2} \|\alpha\|_1 + \frac{\lambda}{2} \|\tilde{\alpha}\|_1 \\
&= \frac{1}{2} \left( \|x\|_2^2 - 2 \langle x, \frac{1}{2} \left( D \alpha + D \tilde{\alpha} \right) \rangle 
                                + \frac{1}{4} \|D \alpha + D \tilde{\alpha}\|_2^2 \right) 
      + \frac{\lambda}{2} \|\alpha\|_1 + \frac{\lambda}{2} \|\tilde{\alpha}\|_1 \\
&= \frac{1}{2} \|x\|_2^2 - \frac{1}{2} \langle x, D \alpha \rangle - \frac{1}{2} \langle x, D \tilde{\alpha} \rangle
       + \frac{1}{8} \left( \|D \alpha\|_2^2 + \|D \tilde{\alpha}\|_2^2 
                                    + 2 \langle D \alpha, D \tilde{\alpha} \rangle \right)
      + \frac{\lambda}{2} \|\alpha\|_1 + \frac{\lambda}{2} \|\tilde{\alpha}\|_1 \\
&\leq \frac{1}{2} \|x\|_2^2 - \frac{1}{2} \langle x, D \alpha \rangle - \frac{1}{2} \langle x, D \tilde{\alpha} \rangle
       + \frac{1}{4} \|D \alpha\|_2^2 + \frac{1}{4} \langle D \alpha, D \tilde{\alpha} \rangle
      + \frac{\lambda}{2} \|\alpha\|_1 + \frac{\lambda}{2} \|\tilde{\alpha}\|_1 + \frac{7}{8} \frac{\varepsilon}{\lambda} \\
&\leq \frac{1}{2} \|x\|_2^2 - \frac{1}{2} \langle x, D \alpha \rangle - \frac{1}{2} \langle x, D \tilde{\alpha} \rangle
       + \frac{1}{4} \|D \alpha\|_2^2 + \frac{1}{4} \langle D \alpha, D \tilde{\alpha} \rangle
      + \frac{1}{2} \langle x - D \alpha, D \alpha \rangle + \frac{1}{2} \langle x - \tilde{D} \tilde{\alpha}, \tilde{D} \tilde{\alpha} \rangle + \frac{7}{8} \frac{\varepsilon}{\lambda} \\
&\leq \frac{1}{2} \|x\|_2^2 - \frac{1}{2} \langle x, D \alpha \rangle - \frac{1}{2} \langle x, D \tilde{\alpha} \rangle
      + \frac{1}{4} \|D \alpha\|_2^2 
      + \frac{1}{4} \langle D \alpha, D \tilde{\alpha} \rangle   
      + \frac{1}{2} \langle x, D \alpha \rangle - \frac{1}{2} \|D \alpha\|_2^2 
      + \frac{1}{2} \langle x, D \tilde{\alpha} \rangle - \frac{1}{2} \|D \alpha\|_2^2 \\
      & \quad + \left(\frac{7}{8} + \frac{3}{2} \right) \frac{\varepsilon}{\lambda} \\
\end{align*}
which simplifies to
\begin{align*}
&\frac{1}{2} \|x\|_2^2 - \frac{3}{4} \|D \alpha\|_2^2 
      - \frac{1}{2} \langle x, D \alpha \rangle 
      - \frac{1}{2} \langle x, D \tilde{\alpha} \rangle
      + \frac{1}{4} \langle D \alpha, D \tilde{\alpha} \rangle   
      + \frac{1}{2} \langle x, D \alpha \rangle
      + \frac{1}{2} \langle x, D \tilde{\alpha} \rangle 
      + \frac{19}{8} \frac{\varepsilon}{\lambda} \\
&= \frac{1}{2} \|x\|_2^2 - \frac{3}{4} \|D \alpha\|_2^2 
      + \frac{1}{4} \langle D \alpha, D \tilde{\alpha} \rangle 
      + \frac{19}{8} \frac{\varepsilon}{\lambda} .
\end{align*}

Now, taking the (expanded) LHS of \eqref{eqn:lower-bound} and the newly derived upper bound of the RHS of \eqref{eqn:lower-bound} yields the inequality:
\begin{align*}
&\frac{1}{2} \|x\|_2^2 - \langle x, D \alpha \rangle + \frac{1}{2} \|D \alpha\|_2^2 + \lambda \|\alpha\|_1 \\
&\leq
\frac{1}{2} \|x\|_2^2 - \frac{3}{4} \|D \alpha\|_2^2 
+ \frac{1}{4} \langle D \alpha, D \tilde{\alpha} \rangle 
+ \frac{19}{8} \frac{\varepsilon}{\lambda} .
\end{align*}
which implies that
\begin{align*}
&- \langle x, D \alpha \rangle + \frac{1}{2} \|D \alpha\|_2^2 + \lambda \|\alpha\|_1 \\
&\leq -\frac{3}{4} \|D \alpha\|_2^2 
      + \frac{1}{4} \langle D \alpha, D \tilde{\alpha} \rangle 
      + \frac{19}{8} \frac{\varepsilon}{\lambda} .
\end{align*}
Replacing $\lambda \|\alpha\|_1$ with $\langle x - D \alpha, D \alpha \rangle$ yields:
\begin{align*}
&-\langle x, D \alpha \rangle + \frac{1}{2} \|D \alpha\|_2^2 
   + \langle x, D \alpha \rangle - \|D \alpha\|_2^2 \\
&\leq -\frac{3}{4} \|D \alpha\|_2^2 
      + \frac{1}{4} \langle D \alpha, D \tilde{\alpha} \rangle 
      + \frac{19}{8} \frac{\varepsilon}{\lambda} ,
\end{align*}
implying that
\begin{align*}
 \frac{1}{4} \|D \alpha\|_2^2 
\leq \frac{1}{4} \langle D \alpha, D \tilde{\alpha} \rangle 
     + \frac{19}{8} \frac{\varepsilon}{\lambda} .
\end{align*}
Hence, 
\begin{align*}
\|D \alpha\|_2^2 
\leq \langle D \alpha, D \tilde{\alpha} \rangle
     + \frac{19}{2} \frac{\varepsilon}{\lambda} .
\end{align*}

Now, note that
\begin{align*}
\|D \alpha - D \tilde{\alpha}\|_2^2 
&= \|D \alpha\|_2^2 + \|D \tilde{\alpha}\|_2^2 - 2 \langle D \alpha, D \tilde{\alpha} \rangle \\
&\leq \|D \alpha\|_2^2 + \|D \tilde{\alpha}\|_2^2 - 2 \|D \alpha\|_2^2 + 19 \frac{\varepsilon}{\lambda} \\
&\leq \|D \alpha\|_2^2 + \|D \alpha\|_2^2 - 2 \|D \alpha\|_2^2 + 26 \frac{\varepsilon}{\lambda} \\
&= 26 \frac{\varepsilon}{\lambda} .
\end{align*}
\end{Proof}

\subsection{The sparsity lemma}

We now prove that the solution to the perturbed problem is sparse for sufficiently small $\varepsilon$. 

\begin{Lemma}[Preservation of Sparsity]
\label{lemma:sparsity-preservation}
Under Assumptions (A1)-(A3), if
\begin{align*}
\tau \geq \varepsilon \left( 1 + \frac{1}{\lambda} \right) + \sqrt{\frac{26 \varepsilon}{\lambda}} ,
\end{align*}
then $\tilde{\alpha}_i = 0$ for all $i \in \mathcal{I}$. 
\end{Lemma}
\begin{Proof}
Let $i \in \mathcal{I}$ be arbitrary. To prove that $\tilde{\alpha}_i = 0$, it is sufficient to show that
\begin{align*}
\left| \langle \tilde{D}_i, x - \tilde{D} \tilde{\alpha} \rangle \right| < \lambda ,
\end{align*}
since $\tilde{\alpha}_i$ is hence zero.

First, note that 
\begin{align*}
\left| \langle \tilde{D}_i, x - \tilde{D} \tilde{\alpha} \rangle \right| 
&= \left| \langle D_i + \tilde{D}_i - D_i, x - \tilde{D} \tilde{\alpha} \rangle \right| \\
&\leq \left| \langle D_i , x - \tilde{D} \tilde{\alpha} \rangle \right| + \| \tilde{D}_i - D_i \|_2 \|x - \tilde{D} \tilde{\alpha} \|_2 \\
&\leq \left| \langle D_i , x - \tilde{D} \tilde{\alpha} \rangle \right| + \varepsilon \qquad \qquad \qquad \qquad \qquad (\text{since } \|x\|_2 \leq 1)
\end{align*}
and 
\begin{align*}
\left| \langle D_i , x - \tilde{D} \tilde{\alpha} \rangle \right| 
&= \left| \langle D_i , x - (D + \tilde{D} - D) \tilde{\alpha} \rangle \right| \\
&\leq \left| \langle D_i , x - D \tilde{\alpha} \rangle \right| 
+ \left| \langle D_i, (\tilde{D} - D) \tilde{\alpha} \rangle \right| \\
&\leq \left| \langle D_i , x - D \tilde{\alpha} \rangle \right| 
+ \|D_i\|_2 \| \tilde{D} - D \|_2 \| \tilde{\alpha} \|_2 \\
&\leq \left| \langle D_i , x - D \tilde{\alpha} \rangle \right| 
+ \frac{\varepsilon}{\lambda} .
\end{align*}
Hence, 
\begin{align*}
\left| \langle \tilde{D}_i, x - \tilde{D} \alpha \rangle \right| 
\leq \left| \langle D_i, x - D \tilde{\alpha} \rangle \right| 
     + \varepsilon \left( 1 + \frac{1}{\lambda} \right) ,
\end{align*}
and so it is sufficient to show that
\begin{align*}
\left| \langle D_i, x - D \tilde{\alpha} \rangle \right| 
< \lambda - \varepsilon \left( 1 + \frac{1}{\lambda} \right) .
\end{align*}
Now,
\begin{align}
\left| \langle D_i, x - D \tilde{\alpha} \rangle \right|
&= \left| \langle D_i, x - D \tilde{\alpha} + D \alpha - D \alpha \rangle \right| \nonumber \\
&\leq \left| \langle D_i, x - D \alpha \rangle \right|
      + \left| \langle D_i, D \alpha - D \tilde{\alpha} \rangle \right| \nonumber \\
&< \lambda - \tau + \|D_i\|_2 \|D \alpha - D \tilde{\alpha}\|_2 \nonumber \\
&< \lambda - \tau + \sqrt{\frac{26 \varepsilon}{\lambda}} \label{eqn:reconstructor-stability},
\end{align}
where \eqref{eqn:reconstructor-stability} is due to Lemma \ref{lemma:reconstructor-stability-proof}. 
Consequently, it is sufficient if $\tau$ is chosen to satisfy
\begin{align*}
\lambda - \tau + \sqrt{\frac{26 \varepsilon}{\lambda}} 
\leq \lambda - \varepsilon \left( 1 + \frac{1}{\lambda} \right),
\end{align*}
yielding:
\begin{align*}
\tau \geq \varepsilon \left( 1 + \frac{1}{\lambda} \right) + \sqrt{\frac{26 \varepsilon}{\lambda}} .
\end{align*}
\end{Proof}

\subsection{Proof of Sparse Coding Stability Theorem}
\begin{Proof}[of Theorem \ref{thm:new-sparse-coding-stability}] 
Recall that $\varphi_D(x)$ is the unique optimal solution to the problem 
\begin{align*}
\min_{z \in \real^k} \frac{1}{2} \| x - D z \|_2^2 + \lambda_1 \| z \|_1 .
\end{align*}
If not for $\ell_1$ penalty, in standard form, the quadratic program is
\begin{align*}
\min_{z \in \real^k} z^T D^T D z - z^T (2 D x) + \lambda_1 \|x\|_1
\end{align*}
Similarly, let $\tilde{Q}(\cdot)$ be the objective using $\tilde{D}$ instead of $D$. Denoting $\bar{z} := \left( \begin{array}{l} z \\ z^+ \\ z^- \end{array} \right)$, an equivalent formulation is
\begin{align*}
& \underset{\bar{z} \in \real^{ 3 k}}{\text{minimize}}
& & Q(\bar{z}) := \frac{1}{2} \bar{z}^T 
\left(
\begin{array}{cc} 
D^T D & \mathbf{0}_{k \times 2k} \\
\mathbf{0}_{2k \times k} & \mathbf{0}_{2k \times 2k} \\
\end{array}
\right)
\bar{z} 
- \frac{1}{2} \bar{z}^T \left( \left( \begin{array}{c} 2 D^T \\ \mathbf{0}_{2k \times d} \end{array} \right) x \right) 
+ \lambda_1 (\mathbf{0}_{k}^T \mathbf{1}_{2k}^T) \bar{z} \\
& \text{subject to } 
& & z^+ \geq \mathbf{0}_k \qquad z^- \geq \mathbf{0}_k \qquad z - z^+ + z^- = \mathbf{0}_k.
\end{align*}
For optimal solutions $\bar{z}_* := \left( \begin{array}{l} z_* \\ z^+_* \\ z^-_* \end{array} \right) $ and $\bar{t}_* := \left( \begin{array}{c} t_* \\ t^+_* \\ t^-_* \end{array} \right)$ of $Q$ and $\tilde{Q}$ respectively, from \cite{daniel1973stability}, 
we have
\begin{align}
\label{eqn:Q-convexity-bound}
(\bar{u} - \bar{z}_*)^T \nabla Q(\bar{z}_*) \geq 0 \\
\label{eqn:Qprime-convexity-bound}
(\bar{u} - \bar{t}_*)^T \nabla \tilde{Q}(\bar{t}_*) \geq 0
\end{align}
for all $\bar{u} \in \real^{3 k}$. 
Setting $\bar{u}$ to $\bar{t}_*$ in \eqref{eqn:Q-convexity-bound} and $\bar{u}$ to $\bar{z}_*$ in \eqref{eqn:Qprime-convexity-bound} and adding \eqref{eqn:Qprime-convexity-bound} and \eqref{eqn:Q-convexity-bound} yields
\begin{align*}
(\bar{t}_* - \bar{z}_*)^T (\nabla Q(\bar{z}_*) - \nabla \tilde{Q}(\bar{t}_*)) \geq 0,
\end{align*}
which is equivalent to
\begin{align}
\label{eqn:grad-Q-bound}
(\bar{t}_* - \bar{z}_*)^T (\nabla \tilde{Q}(\bar{t}_*) - \nabla \tilde{Q}(\bar{z}_*)) \leq (\bar{t}_* - \bar{z}_*)^T (\nabla Q(\bar{z}_*) - \nabla \tilde{Q}(\bar{z}_*))
\end{align}

Here, 
\begin{align*}
\nabla Q(z) = \frac{1}{2} \left(
\begin{array}{cc} 
D^T D & \mathbf{0}_{k \times 2k} \\
\mathbf{0}_{2k \times k} & \mathbf{0}_{2k \times 2k}
\end{array}
\right)
z 
- \frac{1}{2} \left( \begin{array}{c} 2 D^T \\ \mathbf{0}_{2k \times d} \end{array} \right) x
+ \lambda_1 \left( \begin{array}{c} \mathbf{0}_k \\ \mathbf{1}_{2k} \end{array} \right) .
\end{align*}
After plugging in the expansions of $\nabla Q$ and $\nabla \tilde{Q}$ and incurring cancellations from the zeros, \eqref{eqn:grad-Q-bound} becomes
\begin{align}
(t_* - z_*)^T {\tilde{D}}^T {\tilde{D}} (t_* - z_*)
&\leq 
(t_* - z_*)^T \left( (D^T D - {\tilde{D}}^T \tilde{D}) z_* + 2 (\tilde{D} - D)^T x \right) \label{eqn:prelim-sc-covering-bound} \\
&\leq 
(t_* - z_*)^T (D^T D - {\tilde{D}}^T \tilde{D}) z_* 
+ 2 \|t_* - z_*\|_2 \|(\tilde{D} - D)^T x\|_2 \nonumber \\
&\leq 
(t_* - z_*)^T (D^T D - {\tilde{D}}^T \tilde{D}) z_* 
+ \|t_* - z_*\|_2 (2 \varepsilon) \nonumber \\
\end{align}
Let us gain a handle on the first term. Note that $\tilde{D} = D + E$ for some $E$ satisfying $\|E\|_2 \leq \varepsilon$. Hence,
\begin{align*}
&(t_* - z_*)^T (D^T D - {\tilde{D}}^T \tilde{D}) z_* \\
&= \left| (t_* - z_*)^T (E^TD + D^T E + E^T E) z_* \right| \\
&\leq   \left| (t_* - z_*)^T E^T D z_* \right|
      + \left| (t_* - z_*)^T D^T E z_* \right|
      + \left| (t_* - z_*)^T E^T E z_* \right| \\
&\leq   \| t_* - z_*\|_2 
        \Bigl(
          \|E\|_2 \|D z_*\|_2 
          + \left\| D (t_* - z_*) \right\|_2  \|E z_*\|_2
          + \|t_* - z_*\|_2 \|E\|_2^2 \|z_*\|_2
        \Bigr) \\
&\leq   \| t_* - z_*\|_2 
        \left(\frac{\varepsilon \sqrt{s}}{\lambda} 
             + \frac{\varepsilon \sqrt{s}}{\lambda} 
             + \frac{\varepsilon^2}{\lambda} \right) \\
&\leq   \| t_* - z_*\|_2 \frac{3 \varepsilon \sqrt{s}}{\lambda} ,
\end{align*}
where the last step follows because if $\|z_*\|_0 \leq s$, then Lemma \ref{lemma:restricted-operator-norm} in Appendix \ref{sec:infinite-proofs} implies that $\|D z_*\|_2 \leq \sqrt{s} \|z_*\|_2$ (and $\|z_*\|_2 \leq \|z_*\|_1 \leq \frac{1}{\lambda}$).

Now, observe from Lemma \ref{lemma:sparsity-preservation} that $\|t_* - z_*\|_0 \leq s$. Combining this result with the fact that $\tilde{D}$ has $s$-incoherence lower bounded by $\mu$ implies the desired result:
\begin{align*}
\|t_* - z_*\|_2 \leq \frac{3 \varepsilon \sqrt{s}}{\lambda \mu} .
\end{align*}
\end{Proof}

\section{Proof of Restricted Stability Theorem}
\begin{Proof}[of Theorem \ref{thm:old-sparse-coding-stability}] 
For convenience, define $\mathcal{A} := \support{\varphi_D(x)}$, let $\alpha$ be equal to $(\varphi_D(x))_{\mathcal{A}}$, and define the scaled sign vector $\zeta := \lambda \sign(\alpha)$. 
Our strategy will be to show that, for some $\Delta \in \real^s$, the optimal perturbed solution $\varphi_{\tilde{D}}(x)$ satisfies $(\varphi_{\tilde{D}}(x))_{\mathcal{A}} = \alpha + \Delta$ and $(\varphi_{\tilde{D}}(x))_{\mathcal{A}^c} = 0$, where $\mathcal{A}^c := [k] \setminus \mathcal{A}$. 

From the optimality conditions for the LASSO (e.g. see optimality conditions L1 and L2 of \cite{asif2010lasso}), it is sufficient to find $\Delta$ such that
\begin{align*}
\langle \tilde{D}_j, x - \tilde{D}_{\mathcal{A}} (\alpha + \Delta) \rangle &= \zeta_j \quad \text{if } j \in \mathcal{A} , \\
\left| \langle \tilde{D}_j, x - \tilde{D}_{\mathcal{A}} (\alpha + \Delta) \rangle \right| &< \lambda \quad \text{otherwise.}
\end{align*}

We proceed by setting up the linear system and characterizing the solution vector $\Delta$:
\[
\tilde{D}_\mathcal{A}^T (x - \tilde{D}_\mathcal{A} (\alpha + \Delta)) = \zeta
\quad \overset{\text{Solve for } \Delta}{\xrightarrow{\hspace*{1.5cm}}}  \quad
\Delta =  (\tilde{D}_\mathcal{A}^T \tilde{D}_\mathcal{A})^{-1} (\tilde{D}_\mathcal{A}^T (x - \tilde{D}_\mathcal{A} \alpha) - \zeta) .
\]

Since $\tilde{D} = D + E$ for $\|E\|_2 \leq \varepsilon$,
\begin{align*}
\tilde{D}_\mathcal{A}^T (x - \tilde{D}_\mathcal{A} \alpha) 
&= (D_\mathcal{A} + E_\mathcal{A})^T (x - (D_\mathcal{A} + E_\mathcal{A}) \alpha) \\
&= D_\mathcal{A}^T (x - D_\mathcal{A} \alpha) 
- D_\mathcal{A}^T E_\mathcal{A} \alpha 
+ E_\mathcal{A}^T (x - (D_\mathcal{A} + E_\mathcal{A}) \alpha) \\
&= \zeta
- D_\mathcal{A}^T E_\mathcal{A} \alpha 
+ E_\mathcal{A}^T (x - (D_\mathcal{A} + E_\mathcal{A}) \alpha) ,
\end{align*}

and so the solution for $\Delta$ can be reformulated as
\begin{align*}
\Delta =  (\tilde{D}_\mathcal{A}^T \tilde{D}_\mathcal{A})^{-1} (- D_\mathcal{A}^T E_\mathcal{A} \alpha + E_\mathcal{A}^T (x - (D_\mathcal{A} + E_\mathcal{A}) \alpha)).
\end{align*}

Now,
\begin{align*}
\|\Delta\|_2 
&\leq 
\|\tilde{D}_\mathcal{A}^T \tilde{D}_\mathcal{A})^{-1}\|_2 
(\|D_\mathcal{A}^T E_\mathcal{A} \alpha\|_2 + \|E_\mathcal{A}^T (x - (D_\mathcal{A} + E_\mathcal{A}) \alpha)\|_2) \\
&\leq \frac{1}{\mu} (\frac{\varepsilon \sqrt{s}}{\lambda} + \varepsilon) \\
&= \frac{\varepsilon}{\mu} \left( \frac{\sqrt{s}}{\lambda} + 1 \right) .
\end{align*}

For $y \in \real^s$, let $y_{k \times 1}$ be the extension to $\real^k$ satisfying $(y_{k \times 1})_{\mathcal{A}} = y$ and $(y_{k \times 1})_{\mathcal{A}^c} = 0$. 
For $(\alpha + \Delta)_{k \times 1}$ to be optimal for $\lasso(\lambda,\tilde{D},x)$, 
$(\alpha + \Delta)_{k \times 1}$ 
must satisfy the two optimality conditions and $\Delta$ must be small enough such that sign consistency holds between $\alpha$ and $(\alpha + \Delta)$ (i.e. $\sign(\alpha_j) = \sign(\alpha_j + \Delta_j)$ for all $j \in [s]$).

We first check the optimality conditions. The first optimality condition is equivalent to
\begin{align*}
\langle \tilde{D}_j, x - \tilde{D}_{\mathcal{A}} (\alpha + \Delta) \rangle = \lambda 
\quad \text{ for } j \in \mathcal{A};
\end{align*}
this condition is satisfied by construction. The second optimality condition is equivalent to
\begin{align*}
 \left| \langle \tilde{D}_j, x - \tilde{D}_{\mathcal{A}} (\alpha + \Delta) \rangle \right| < \lambda 
\quad \text{ for } j \notin \mathcal{A}.
\end{align*}
But for $j \notin \mathcal{A}$,
\begin{align*}
 \left| \langle \tilde{D}_j, x - \tilde{D}_{\mathcal{A}} (\alpha + \Delta) \rangle \right|
&= \left| \langle D_j + E_j, x - (D_{\mathcal{A}} + E_{\mathcal{A}}) (\alpha + \Delta) \rangle \right| \\
&= \left| \langle D_j, x - D_{\mathcal{A}} \alpha \rangle
- \langle D_j, D_{\mathcal{A}} \Delta \rangle
+ \langle E_j, x - (D_{\mathcal{A}} + E_{\mathcal{A}}) (\alpha + \Delta) \rangle 
- \langle D_j, E_{\mathcal{A}} (\alpha + \Delta) \rangle
\right|\\
&< \lambda 
- \tau 
+ \frac{\varepsilon \sqrt{s}}{\mu} \left( \frac{\sqrt{s}}{\lambda} + 1 \right) 
+ \varepsilon
+ \frac{\varepsilon}{\lambda} \\
&= \lambda 
- \tau 
+ \varepsilon \left( \frac{\frac{s}{\lambda} + \sqrt{s}}{\mu} + \frac{1}{\lambda} + 1 \right) ,
\end{align*}
and so this condition is satisfied provided that
\begin{align*}
\varepsilon \left( \frac{\frac{s}{\lambda} + \sqrt{s}}{\mu} + \frac{1}{\lambda} + 1 \right)
\leq \tau.
\end{align*}

Now, we check sign consistency. Clearly sign consistency holds over $\mathcal{A}^c$. It remains to check that it holds over $\mathcal{A}$. 
Observe that
\[
\|\Delta\|_\infty \leq \|\Delta\|_2 \leq \frac{\varepsilon}{\mu} \left( \frac{\sqrt{s}}{\lambda} + 1 \right).
\]
Hence, sign consistency holds provided that
\[
|\alpha_i| > \varepsilon \left( \frac{1}{\mu} \left( \frac{\sqrt{s}}{\lambda} + 1 \right) \right).
\]

All the above constraints are satisfied if $\tau$ satisfies
\begin{align*}
\varepsilon \left( \frac{\frac{s}{\lambda} + \sqrt{s}}{\mu} + \frac{1}{\lambda} + 1 \right)
\leq \tau .
\end{align*}
\end{Proof}

\section{Proof of Symmetrization by Ghost Sample Lemma}
\label{sec:proof-symmetrization-ghost-lemma}

\begin{Proof}[of Lemma \ref{lemma:symmetrization-ghost}] 
Replace $\F(\sigma_n)$ from the notation of \cite{mendelson2004importance} with $\F(\mathbf{z},\mathbf{x''})$. A modified one-sided version of \cite[Lemma 2.2]{mendelson2004importance} that uses the more favorable Chebyshev-Cantelli inequality implies that, for every $t > 0$:
\begin{align*}
\left( 1 - \frac{4 \sup_{f \in \F} \mathrm{Var}(\lossof{f})}{4 \sup_{f \in \F} \mathrm{Var}(\lossof{f}) + m t^2} \right) 
&{\Pr}_{\mathbf{z \, x''}} \left\{
\exists f \in \F(\mathbf{z},\mathbf{x''}) ,\,\, 
(\Prob - \ProbZ) \lossof{f} \geq t
\right\} \\
\leq
&{\Pr}_{\mathbf{z \, z' x''}} \left\{
\exists f \in \F(\mathbf{z},\mathbf{x''}) ,\,\, 
(\ProbZp - \ProbZ) \lossof{f} \geq \frac{t}{2}
\right\} .
\end{align*}
As the losses lie in $[0,b]$ by assumption, 
it follows that $\sup_{f \in \F} \mathrm{Var}(\lossof{f}) \leq \frac{b^2}{4}$. The lemma follows since the left hand factor of the LHS of the above inequality is at least $\frac{1}{2}$ whenever $m \geq \left(\frac{b}{t}\right)^2$. 
\end{Proof}

\section{Proofs for overcomplete setting}
\label{sec:overcomplete-proofs}

\begin{Proof}[of Theorem \ref{thm:final-learning-bound-overcomplete}] 
Proposition \ref{prop:symmetrization-ghost-overcomplete} and Lemmas \ref{lemma:good-ghost} and \ref{lemma:large-deviation-good-ghost} imply that
\begin{align*}
  {\Pr}_{\mathbf{z}} & \left\{ 
  \begin{array}{l}
    \exists f \in \F_\mu ,\,\, 
    \smargingt{D, \mathbf{x}, \iota} \\
    \qquad \booland
    \left( (\Prob - \ProbZ) \lossof{f} > t \right) 
  \end{array} 
  \right\} \\
  &\leq 2 \left( \left( \frac{8 (r/2)^{1/(d+1)}}{\varepsilon} \right)^{(d+1) k} \exp(-m \varpi^2 / ( 2 b^2)) + \delta \right) .
\end{align*}
Equivalently,
\begin{align*}
  {\Pr}_{\mathbf{z}} & \left\{ 
  \begin{array}{l}
    \exists f \in \F_\mu ,\,\, \smargingt{D, \mathbf{x}, \iota} \\
    \qquad \booland 
    \left( (\Prob - \ProbZ) \lossof{f} > 2 \left(\varpi + 2 L \beta + \frac{b \eta(m,d,k,\varepsilon,\delta)}{m} \right) \right) 
  \end{array}
  \right\} \\
  &\leq 2 \left( \left( \frac{8 (r/2)^{1/(d+1)}}{\varepsilon} \right)^{(d+1) k} \exp(-m \varpi^2 / ( 2 b^2)) + \delta \right) .
\end{align*}
Now, expand $\beta$ and $\eta$ and replace $\delta$ with $\delta / 4$:
\begin{align*}
{\Pr}_{\mathbf{z}} & \left\{ 
\begin{array}{l}
\exists f \in \F_\mu ,\,\, \smargingt{D, \mathbf{x}, \iota} \booland \\
\qquad 
(\Prob - \ProbZ) \lossof{f} 
> 2 \left(\varpi + 2 L \varepsilon \frac{1}{\lambda} \left( 1 + \frac{3 r \sqrt{s}}{\mu} \right)
+ \frac{b 
(
d k \log \frac{3096}{\smargin^2(D,\mathbf{x}) \cdot \lambda} + \log (2 m + 1) + \log \frac{4}{\delta})
}{m} \right) 
\end{array}
\right\} \\
&\leq 2 \left( \frac{8 (r/2)^{1/(d+1)}}{\varepsilon} \right)^{(d+1) k} \exp(-m \varpi^2 / ( 2 b^2)) + \frac{\delta}{2} .
\end{align*}
Choosing $\frac{\delta}{4} = \left( \frac{8 (r/2)^{1/(d+1)}}{\varepsilon} \right)^{(d+1) k} \exp(-m \varpi^2 / ( 2 b^2))$ yields
\begin{align*}
{\Pr}_{\mathbf{z}} & \left\{ 
\begin{array}{l} 
\exists f \in \F_\mu ,\,\, \smargingt{D, \mathbf{x}, \iota} \booland \\
\qquad (\Prob - \ProbZ) \lossof{f} 
> 2 \left(
\begin{array}{l}
\varpi 
+ 2 L \varepsilon \frac{1}{\lambda} \left( 1 + \frac{3 r \sqrt{s}}{\mu} \right)
 + \\ \frac{b 
(d k \log \frac{3096}{\smargin^2(D,\mathbf{x}) \cdot \lambda} + \log (2 m + 1) + 
(d+1) k \log \frac{\varepsilon}{8 (r / 2)^{1/(d+1)}} + \frac{m \varpi^2}{b^2})
}{m} 
\end{array}
\right)
\end{array}
\right\} \\
&\leq 4 \cdot \left( \frac{8 (r/2)^{1/(d+1)}}{\varepsilon} \right)^{(d+1) k} \exp(-m \varpi^2 / ( 2 b^2)) ,
\end{align*}
which is equivalent to
\begin{align*}
{\Pr}_{\mathbf{z}} & \left\{
\begin{array}{l}
\exists f \in \F_\mu ,\,\, \smargingt{D, \mathbf{x}, \iota} \booland \\
\qquad (\Prob - \ProbZ) \lossof{f} 
> 2 \left(
\begin{array}{l}
\varpi 
+ 2 L \varepsilon \frac{1}{\lambda} \left( 1 + \frac{3 r \sqrt{s}}{\mu} \right)
+ \\ \frac{b 
(
d k \log \frac{3096}{\smargin^2(D,\mathbf{x}) \cdot \lambda} 
- (d + 1) k \log \frac{8}{\varepsilon} + k \log \frac{2}{r} + \log (2 m + 1)  
+ \frac{m \varpi^2}{b^2}
)
}{m} 
\end{array}
\right) 
\end{array}
\right\} \\
&\leq 4 \cdot \left( \frac{8 (r/2)^{1/(d+1)}}{\varepsilon} \right)^{(d+1) k}
\exp(-m \varpi^2 / ( 2 b^2)) ,
\end{align*}

Let $\delta$ (a new variable, not related to the previous incarnation of $\delta$) be equal to the upper bound, and solve for $\varpi$, yielding:
\begin{align*}
\varpi = 
b \sqrt{\frac{2 ( (d+1) k \log \frac{8}{\varepsilon} + k \log \frac{r}{2} + \log \frac{4}{\delta})}{m}}
\end{align*}
and hence
\begin{align*}
{\Pr}_{\mathbf{z}} & \left\{
\begin{array}{l} 
\exists f = (D,w) \in \F_\mu ,\,\, \smargingt{D, \mathbf{x}, \iota} \booland \\
\qquad (\Prob - \ProbZ) \lossof{f} 
> 2 \left(
\begin{array}{l}
b \sqrt{\frac{2 ( (d+1) k \log \frac{8}{\varepsilon} + k \log \frac{r}{2} + \log \frac{4}{\delta})}{m}} 
+ 2 L \varepsilon \frac{1}{\lambda} \left( 1 + \frac{3 r \sqrt{s}}{\mu} \right)
+ \\ \frac{b 
(
d k \log \frac{3096}{\smargin^2(D,\mathbf{x}) \cdot \lambda} + \log (2 m + 1)  + \log \frac{4}{\delta}
)
}{m} 
\end{array}
\right)
\end{array}
 \right\} \\
&\leq \delta ,
\end{align*}

If we set $\varepsilon = \frac{1}{m}$, then provided that $m > \frac{387}{\smargin^2(D,\mathbf{x}) \cdot \lambda}$:
\begin{align*}
{\Pr}_{\mathbf{z}} & \left\{ 
\begin{array}{l}
\exists f \in \F_\mu ,\,\, \smargingt{D, \mathbf{x}, \iota} \booland \\
\qquad (\Prob - \ProbZ) \lossof{f} 
> 2 \left(
\begin{array}{l}
b \sqrt{\frac{2 ( (d+1) k \log (8 m) + k \log \frac{r}{2} + \log \frac{4}{\delta})}{m}} + \frac{2 L}{m} \left( \frac{1}{\lambda} (1 + \frac{3 r \sqrt{s}}{\mu}) \right) 
+ \\ \frac{b}{m} 
\left(
d k \log \frac{3096}{\smargin^2(D,\mathbf{x}) \cdot \lambda} + \log (2 m + 1) + \log \frac{4}{\delta}
\right)
\end{array}
\right)
\end{array}
\right\} \\
&\leq \delta .
\end{align*}

It remains to distribute a prior across the bounds for each choice of $s$ and $\mu$. To each choice of $s \in [k]$ assign prior probability $\frac{1}{k}$. To each choice of $i \in \nat \cup \{0\}$ for $2^{-i} \leq \mu$ assign prior probability $(i+1)^{-2}$. For a given choice of $s \in [k]$ and $2^{-i} \leq \mu$ we use
$\delta(s,i) := \frac{6}{\pi^2} \frac{1}{(i+1)^2} \frac{1}{k} \delta$ (since $\sum_{i=1}^\infty \frac{1}{i^2} = \frac{\pi^2}{6}$). 
Then, provided that 
\begin{align*}
m > \frac{387}{\smargin(D,\mathbf{x})^2 \lambda} ,
\end{align*}
the generalization error $(\Prob - \ProbZ) \lossof{f}$ is 
bounded by:
\begin{align*}
& \,\, 2 b \sqrt{\frac{2 \left( (d+1) k \log (8 m) + k \log \frac{r}{2} + \log \frac{2 \pi^2 \left( \log_2 \frac{4}{\mu_s(D)} \right)^2 k}{3 \delta} 
\right)}{m}} \\
& \,\, + \frac{2 b}{m} 
\left( d k \log \frac{3096}{\smargin^2(D,\mathbf{x}) \cdot \lambda} + \log (2 m + 1)  
+ \log \frac{2 \pi^2 \left( \log_2 \frac{4}{\mu_s(D)} \right)^2 k}{3 \delta}
\right) \\
& \,\, + \frac{4 L}{m} 
\left( \frac{1}{\lambda} (1 + \frac{6 r \sqrt{s}}{\mu_s(D)}) \right) .
\end{align*}

\end{Proof}

\section{Infinite-dimensional setting}
\label{sec:infinite-proofs}

\begin{Proof}[of Lemma \ref{lemma:unlikely-bad-ghost}] 
Recall that $\eta = \log \frac{1}{\delta}$. Suppose, as in the event being measured, that there is no subset of the ghost sample $\mathbf{x'}$ of size at least $\eta$ such that the $\tau$-level $s$-margin condition holds for the entire subset. Equivalently, there is a subset of at least $\eta$ points in the ghost sample $\mathbf{x'}$ that violate the $\tau$-level $s$-coding margin condition. From the permutation argument, if no point of $\mathbf{x''}$ violates $\smargingt{D, \cdot, \tau}$, then the probability that over $\eta = \log \frac{1}{\delta}$ points of $\mathbf{x'}$ will violate $\smargingt{D, \cdot, \tau}$ is at most $\delta$.
\end{Proof}

\begin{Proof}[of Lemma \ref{lemma:sc-phi-difference-bound-WRT-U}] 
By definition, $\varphi_{U S}(x) = \argmin_{z \in \real^k} \| x - U S v \|_2 + \lambda \| z \|_1$. Note that $\argmin_{z \in \real^k} \| x - U S z \|_2 = \argmin_{z \in \real^k} \| U^T x - U^T U S z \|_2 = \argmin_{z \in \real^k} \| U^T x - S z \|_2 $, where the first equality follows because any $x$ in the complement of the image of $U$ will be orthogonal to $U S z$, for any $z$; hence, it is sufficient to approximate the projection of $x$ onto the range of $U$. 
Thus, $\varphi_{U S}(x) = \argmin_{z \in \real^k} \|U^T x - S z\|_2^2 + \lambda \|z\|_1$. 
It will be useful to apply a well-known reformulation of this minimization problem as a quadratic program with linear constraints. Denoting $\bar{z} := \bar{z} := (z^T {z^+}^T {z^-}^T)^T$, an equivalent formulation is
\begin{equation*}
\begin{aligned}
& \underset{\bar{z} \in \real^{ 3 k}}{\text{minimize}}
& & Q_U(\bar{z}) := \bar{z}^T 
\left(
\begin{array}{cc} 
S^T S & \mathbf{0}_{k \times 2k} \\
\mathbf{0}_{2k \times k} & \mathbf{0}_{2k \times 2k}
\end{array}
\right)
\bar{z} 
- \bar{z}^T \left( \left( \begin{array}{c} 2 S^T U^T \\ \mathbf{0}_{2k \times d} \end{array} \right) x \right) 
+ \lambda (\mathbf{0}_k^T \, \mathbf{1}_{2k}^T) \bar{z} \\
& \text{subject to } 
& & z^+ \geq \mathbf{0}_k \qquad z^- \geq \mathbf{0}_k \qquad z - z^+ + z^- = \mathbf{0}_k,
\end{aligned}
\end{equation*}

For optimal solutions $\bar{z}_* := \left( \begin{array}{c} z_* \\ z^+_* \\ z^-_* \end{array} \right) $ and $\bar{t}_* := \left( \begin{array}{c} t_* \\ t^+_* \\ t^-_* \end{array} \right)$ of $Q_U$ and $Q_{U'}$ respectively, from \cite{daniel1973stability}, we have
\begin{align}
\label{eqn:Q-convexity-bound2}
(\bar{u} - \bar{z}_*)^T \nabla Q_U(\bar{z}_*) \geq 0 \\
\label{eqn:Qprime-convexity-bound2}
(\bar{u} - \bar{t}_*)^T \nabla Q_{U'}(\bar{t}_*) \geq 0
\end{align}
for all $\bar{u} \in \real^{3 k}$. 
Setting $\bar{u}$ to $\bar{t}_*$ in \eqref{eqn:Q-convexity-bound2} and $\bar{u}$ to $\bar{z}_*$ in \eqref{eqn:Qprime-convexity-bound2} and adding \eqref{eqn:Q-convexity-bound2} and \eqref{eqn:Qprime-convexity-bound2} yields
\begin{align*}
(\bar{t}_* - \bar{z}_*)^T (\nabla Q_U(\bar{z}_*) - \nabla Q_{U'}(\bar{t}_*)) \geq 0,
\end{align*}
which is equivalent to
\begin{align}
\label{eqn:grad-Q-bound2}
(\bar{t}_* - \bar{z}_*)^T (\nabla Q_{U'}(\bar{t}_*) - \nabla Q_{U'}(\bar{z}_*)) \leq (\bar{t}_* - \bar{z}_*)^T (\nabla Q_U(\bar{z}_*) - \nabla Q_{U'}(\bar{z}_*)) .
\end{align}

Here, $\nabla Q_U(z) = \left(
\begin{array}{cc} 
S^T S & \mathbf{0}_{k \times 2k} \\
\mathbf{0}_{2k \times k} & \mathbf{0}_{2k \times 2k}
\end{array}
\right)
z 
- \left( \begin{array}{c} 2 S^T U^T \\ \mathbf{0}_{2k \times d} \end{array} \right) x
+ \lambda \left( \begin{array}{c} \mathbf{0}_k \\ \mathbf{1}_{2k} \end{array} \right)$.
After plugging in the expansions of $\nabla Q_U$ and $\nabla Q_{U'}$ and incurring cancellations from the zeros, \eqref{eqn:grad-Q-bound2} becomes
\begin{multline*}
(t_* - z_*)^T (S^T S t_* - 2 S^T {U'}^T x - S^T S z_* + 2 S^T {U'}^T x) \\
\leq (t_* - z_*)^T (S^T S z_* - 2 S^T U^T x - S^T S z_* + 2 S^T {U'}^T x ) ,
\end{multline*}
which reduces to
\begin{align*}
(t_* - z_*)^T S^T S (t_* - z_*) &\leq 2 (t_* - z_*)^T S^T ({U'}^T - U^T) x .
\end{align*}

Since both $t_*$ and $z_*$ are $s$-sparse, wherever we typically would consider the operator norm $\|S\|_2 := \sup_{\|t\|=1} \|S t\|_2$, we instead need only consider the $2s$-restricted operator norm $\|S\|_{2,2s}$. 

Note that $(t_* - z_*)^T S^T S (t_* - z_*) \geq \mu_{2s}(S) \| t_* - z_* \|_2^2$, which implies that
\begin{align*}
\|t_* - z_* \|_2^2 \leq \frac{2}{\mu_{2s}(S)} \| t_* - z_* \| \| S \|_{2,2s} \| ({U'}^T - U^T) x \| 
\end{align*}
and hence
\begin{align*}
\|t_* - z_*\|_2 \leq \frac{2 \| S \|_{2,2s} }{\mu_{2s}(S)} \| ({U'}^T - U^T) x \|_2. 
\end{align*}
\end{Proof}

\begin{Lemma}
\label{lemma:restricted-operator-norm}
If $S \in \left(\ball{}{k}\right)^k$, then $\|S\|_{2,s} \leq \sqrt{s}$. 
\end{Lemma}
\begin{Proof}
Define $S_\Lambda$ as the submatrix of $S$ that selects the columns indexed by $\Lambda$. Similarly, for $t \in \real^k$ define the coordinate projection $t_\Lambda$ of $t$.
\begin{align*}
&\sup_{\{t: \|t\|=1, |\support(t)| \leq s \}} \| S t \|_2 \\
&= \min_{\{\Lambda \subseteq [k]: |\Lambda| \leq s\}} \sup_{\{t: \|t\| = 1, \support(t) \subseteq \Lambda\}} \| S_\Lambda t_\Lambda \|_2 \\
&= \min_{\{\Lambda \subseteq [k]: |\Lambda| \leq s\}} \sup_{\{t: \|t\| = 1, \support(t) \subseteq \Lambda\}} \left\| \sum_{\omega \in \Lambda} t_\omega S_\omega  \right\|_2 \\
&\leq \min_{\{\Lambda \subseteq [k]: |\Lambda| \leq s\}} \sup_{\{t: \|t\| = 1, \support(t) \subseteq \Lambda\}} \sum_{\omega \in \Lambda} | t_\omega | \| S_\omega \|_2 \\
&\leq \min_{\{\Lambda \subseteq [k]: |\Lambda| \leq s\}} \sup_{\{t: \|t\| = 1, \support(t) \subseteq \Lambda\}} \sum_{\omega \in \Lambda} | t_\omega | \\
&\leq \min_{\{\Lambda \subseteq [k]: |\Lambda| \leq s\}} \sup_{\{t: \|t\| = 1, \support(t) \subseteq \Lambda\}} \| t_\Lambda \|_1 \\
&\leq \min_{\{\Lambda \subseteq [k]: |\Lambda| \leq s\}} \sup_{\{t: \|t\| = 1, \support(t) \subseteq \Lambda\}} \sqrt{s} \| t_\Lambda \|_2 \\
&= \sqrt{s} .
\end{align*}
\end{Proof}

\section{Covering numbers}
\label{sec:covering-numbers-appendix}

For a Banach space $E$ of dimension $d$, the $\varepsilon$-covering numbers of the radius $r$ ball of $E$ are bounded as 
$\mathcal{N}(r B_E, \varepsilon) \leq (4 r / \varepsilon)^d$ 
\citep[Chapter I, Proposition 5]{cucker2002mathematical}. 

For spaces of dictionaries obeying some deterministic property, such as 
\begin{align*}
\Dspace_\mu = \{D \in \Dspace: \mu_s(D) \geq \mu\} ,
\end{align*}
one must be careful to use a \emph{proper} $\varepsilon$-cover so that the representative elements of the cover also obey the desired property; a proper cover is more restricted than a cover in that a proper cover must be a subset of the set being covered, rather than simply being a subset of the ambient Banach space. That is, if $A$ is a proper cover of a subset $T$ of a Banach space $E$, then $A \subseteq T$. For a cover, we need only $A \subseteq E$. The following bound relates proper covering numbers to covering numbers (a simple proof can be found in \citealt[Lemma 2.1]{vidyasagar2002learning}):
If $E$ is a Banach space and $T \subseteq E$ is a bounded subset, then 
\begin{align*}
\mathcal{N}(E, \varepsilon, T) \leq \mathcal{N}_{\mathrm{proper}}(E, \varepsilon / 2, T) .
\end{align*}

Let $d,k \in \nat$. 
Define $E_\mu := \{E \in \left(\ball{}{d}\right)^k : \mu_s(D) \geq \mu\}$ 
and $\Wspace := \ball{r}{d}$. 
The following bounds derive directly from the above.
\begin{Proposition}
\label{prop:proper-covering-cardinality}
The proper $\varepsilon$-covering number of $E_\mu$ is bounded by 
$\left( 8 / \varepsilon \right)^{d k}$.
\end{Proposition}

\begin{Proposition}
\label{prop:hypothesis-covering-cardinality}
The product of the proper $\varepsilon$-covering number of $E_\mu$ and the $\varepsilon$-covering number of $\Wspace$ is bounded by
\[
\left( \frac{8 (r/2)^{1/(d+1)}}{\varepsilon} \right)^{(d+1) k} .
\]
\end{Proposition}

\vskip 0.2in
\bibliography{predictive_sparse_coding}


\newpage

\section{Glossary}
\label{sec:glossary}

\renewcommand*{\glossaryname}{}

\printglossary[style=long3colheaderborder]

\end{document}